\documentclass[twoside,11pt]{article}
\usepackage{amssymb,amsmath,color,graphicx}
\usepackage{url,times}

\usepackage{hyperref}
 
\usepackage{amssymb,amsmath,color,times}

\newcommand{\BlackBox}{\rule{1.5ex}{1.5ex}}  
\newenvironment{proof}{\par\noindent{\bf Proof\ }}{\hfill\BlackBox\\[2mm]}
 
\newtheorem{theorem}{Theorem}
\newtheorem{lemma}[theorem]{Lemma} 
\newtheorem{proposition}[theorem]{Proposition}

 \long\def\acks#1{\vskip 0.3in\noindent{\large\bf Acknowledgments}\vskip 0.2in
\noindent #1}

 \RequirePackage{natbib}
\RequirePackage{graphicx}
\bibpunct{(}{)}{;}{a}{,}{,}

\def \C_1_2{ C_{1/2} }
\def \Kc{  {\widetilde{K}} }
\def \f{  {\mathbf{f}} }

\def \hf{  \hat{f} }
\def \W{ {\mathbf{W} } }
\def \hW { {\hat{W} } }

  \def \bS { \boldsymbol\Sigma }  
  \def \bmu { \boldsymbol\mu }  
   \def \bC { \boldsymbol C } 
   \def \hS { \widehat{\Sigma} }

\def \bzeta{ {\boldsymbol \zeta} }
\def \E{{\mathbb E}}
\def \P{{\mathbb P}}
\def \anc{ {\rm A} }
\def \hull{ {\rm hull} }

\def \sou{ {\rm sources} }
\def \des{ {\rm D} }
\def \num{ {\rm num} }

\def \sinks{ {\rm sinks} }
\def \depth{ {\rm depth} }
\def \X { \mathcal{X}}
\def \F { \mathcal{F}}
\def \lmin{ \lambda_{\min}}
\def \lmax{ \lambda_{\max}}

\def \X{  \mathcal{X}}
\def \F{  \mathcal{F}}
\def \Y{ \mathcal{Y}}

\def \lmin{ \lambda_{\min} } 
\def \lmax{ \lambda_{\max} }

\def \f{ \mathbf{f}}
\def \h{ \mathbf{h}}

\def \g{ \mathbf{g}}

\def \h{ \mathbf{h}  }

\def \lmaxQ{ \lmax(Q) }

\newcommand{\BEAS}{\begin{eqnarray*}}
\newcommand{\EEAS}{\end{eqnarray*}}
\newcommand{\BEA}{\begin{eqnarray}}
\newcommand{\EEA}{\end{eqnarray}}
\newcommand{\BEQ}{\begin{equation}}
\newcommand{\EEQ}{\end{equation}}
\newcommand{\BIT}{\begin{itemize}}
\newcommand{\EIT}{\end{itemize}}
\newcommand{\BNUM}{\begin{enumerate}}
\newcommand{\ENUM}{\end{enumerate}}
\newcommand{\BA}{\begin{array}}
\newcommand{\EA}{\end{array}}

\newcommand{\Diag}{\mathop{\rm Diag}}

\newcommand{\tr}{\mathop{ \rm tr}}

\newcommand{\idm}{{ \rm I } }
\newcommand{\rb}{\mathbb{R}}

\newcommand{\mysec}[1]{Section~\ref{sec:#1}}
\newcommand{\eq}[1]{Eq.~(\ref{eq:#1})}
\newcommand{\myfig}[1]{Figure~\ref{fig:#1}}

\RequirePackage{natbib}

\title{High-Dimensional Non-Linear Variable Selection \\ through Hierarchical Kernel Learning}

\author{ \Large{Francis Bach} \\[.5cm]
INRIA - WILLOW Project-Team \\
Laboratoire d'Informatique de l'Ecole Normale Sup\'erieure \\
(CNRS/ENS/INRIA UMR 8548) \\
23, avenue d'Italie,
75214 Paris, France \\
\texttt{francis.bach@inria.fr}
       }

\begin{document}
\maketitle

\begin{abstract}
We consider the problem of high-dimensional non-linear variable selection for supervised learning.  Our approach is based on performing linear selection among exponentially many  appropriately defined  positive definite kernels that characterize non-linear interactions between the original variables. To select efficiently from these many kernels, we use the natural hierarchical structure of the problem to extend the multiple kernel learning framework to kernels that can 
 be embedded in a directed acyclic graph; we show that it is then possible to perform kernel selection through a graph-adapted sparsity-inducing norm,  in polynomial time in the number of selected kernels. Moreover, we study the consistency of variable selection  in high-dimensional settings, showing that under certain assumptions, our regularization framework allows a number of irrelevant variables which is  exponential in the number of observations. Our simulations on synthetic datasets and datasets from the UCI repository show state-of-the-art predictive performance for non-linear regression problems. \end{abstract}

 \section{Introduction}
High-dimensional problems represent a recent and important topic in machine learning, statistics and signal processing. In such settings, some notion of sparsity is a fruitful way of avoiding overfitting, for example through variable or feature selection. This has led to many  algorithmic and theoretical advances.  In particular, regularization by sparsity-inducing norms such as the $\ell_1$-norm has  attracted a lot of interest in recent years. While early work has focused on efficient algorithms to solve the convex optimization problems, recent research has looked at the model selection properties and predictive performance of such methods, in the linear case~\citep{Zhaoyu,yuanlin,zou,martin,tsyb,zhangL1} or within constrained non-linear settings such as the multiple kernel learning framework~\citep{gert,srebro,grouplasso,kolt,ying} or generalized additive models~\citep{spam,cosso}.
 
However, most of the recent work dealt with \emph{linear high-dimensional} variable selection, while the focus of much of the earlier work in machine learning and statistics was  on \emph{non-linear low-dimensional} problems: indeed, in the last two decades,  kernel methods have been a prolific  theoretical and algorithmic machine learning framework. By using appropriate regularization by Hilbertian norms, representer theorems enable to consider large and potentially infinite-dimensional feature spaces while working within an implicit feature space no larger than the number of observations. This has led to numerous works on kernel design adapted to specific data types and generic kernel-based algorithms for many learning tasks \citep[see, e.g.,][]{smola-book,Cristianini2004}. However, while non-linearity is required in many domains such as computer vision or bioinformatics, most theoretical results related to non-parametric methods do not scale well with input dimensions. In this paper, our goal is to bridge the gap between linear and non-linear methods, by tackling \emph{high-dimensional non-linear} problems.

The task of non-linear variable section is a hard problem with few approaches that have both good theoretical and algorithmic properties, in particular in high-dimensional settings. Among classical  methods, some are implicitly or explicitly based on sparsity and model selection, such as boosting~\citep{adaboost}, multivariate additive regression splines~\citep{mars}, 
 decision trees~\citep{cart84}, random forests~\citep{randomforests}, Cosso~\citep{cosso} or Gaussian process based methods~\citep[see, e.g.,][]{GP}, while some others do not rely on sparsity, such as nearest neighbors or kernel methods~\citep[see, e.g.,][]{ptpr,Cristianini2004}.

First attempts were made to combine non-linearity and sparsity-inducing norms by considering \emph{generalized additive models}, where the predictor function is assumed to be a sparse linear combination of non-linear functions of each variable~\citep{skm,grouplasso,spam}. However, as shown in \mysec{universal}, higher orders of interactions are needed for universal consistency, i.e., to adapt to the potential high complexity of the interactions between the relevant variables; we need to potentially allow $2^p$ of them for $p$ variables (for all possible subsets of the $p$ variables). Theoretical results suggest that with appropriate assumptions,  sparse methods such as greedy methods and methods based on the $\ell_1$-norm would be able to deal correctly with $2^p$ features if $p$ is of the order of the number of observations $n$~\citep{martin,cs2,zhang_greedy}. However, in presence of more than a few dozen variables, in order to deal with that many features, or even to simply enumerate those, a certain form of factorization or recursivity is needed.
In this paper, we propose to use a hierarchical structure based on directed acyclic graphs, which is natural in our context of non-linear variable selection.
 
We consider a positive definite kernel that can be expressed as a large sum of positive definite \emph{basis} or \emph{local kernels}. This exactly corresponds to the situation where a large feature space is the concatenation of  smaller feature spaces, and we aim to do selection among these many kernels (or equivalently feature spaces), which may be done through  multiple kernel learning~\citep{skm}. One major difficulty however is that the number of these smaller kernels is usually exponential in the dimension of the input space and applying multiple kernel learning directly to this decomposition would be intractable. As shown in \mysec{decompositions}, for non-linear variable selection, we consider a sum of kernels which are indexed by the set of subsets of all considered variables, or more generally by $\{0,\dots,q\}^{p}$, for $q\geqslant 1$.
 
In order to perform selection efficiently, we make the extra assumption that these small kernels can be embedded in a \emph{directed acyclic graph} (DAG). Following~\citet{cap}, we consider in \mysec{mkl} a specific combination of $\ell_2$-norms that is adapted to the DAG, and that will restrict the authorized sparsity patterns to certain configurations; in our specific kernel-based framework, we are able to use the DAG to design an optimization algorithm which has polynomial complexity in the number of selected kernels (\mysec{optimization}).  In simulations (\mysec{simulations}), we focus on  \emph{directed grids}, where our framework allows to perform non-linear variable selection.
We provide some experimental validation of our novel regularization framework; in particular, we compare it to the regular $\ell_2$-regularization, greedy forward selection and non-kernel-based methods, and shows that it is always competitive and often leads to better performance, both on synthetic examples, and standard regression datasets from the UCI repository.

Finally, we extend in \mysec{consistency} some of the known consistency results of the Lasso and multiple kernel learning~\citep{Zhaoyu,grouplasso}, and give a partial answer to the model selection capabilities of our regularization framework by giving necessary and sufficient conditions for model consistency. In particular, we show that our framework is adapted to estimating consistently only the \emph{hull}  of the relevant variables. Hence, by restricting the statistical power of our method, we gain computational efficiency. Moreover, we show that we can obtain scalings between the number of variables and the number of observations which are similar to the linear case~\citep{martin,cs2,Zhaoyu,yuanlin,zou,martin,tsyb,zhangL1}: indeed, we show that our regularization framework may achieve non-linear variable selection consistency even with a number of variables $p$ which is exponential in the number of observations $n$. Since we deal with $2^p$ kernels, we achieve consistency with a number of kernels which is \emph{doubly} exponential in $n$. Moreover, for general directed acyclic graphs, we show that the total number of vertices may grow unbounded as long as the maximal out-degree (number of children) in the DAG is less than exponential in the number of observations.

This paper extends previous work~\citep{hkl}, by providing more background on multiple kernel learning, detailing all proofs, providing new consistency results in high dimension, and comparing our non-linear predictors with non-kernel-based methods.

\paragraph{Notation.}
Throughout the paper we consider Hilbertian norms $\| f\|$ for elements $f$ of Hilbert spaces, where the specific Hilbert space can always be inferred from the context (unless otherwise stated).
For rectangular matrices $A$, we denote
by $\| A \|_{\rm op}$ its largest singular value. We   denote by $\lmaxQ$ and $\lambda_{\min}(Q)$ the largest and smallest eigenvalue of a symmetric matrix $Q$. These are naturally extended to compact self-adjoint operators~\citep{brezis80analyse,conway}.

 Moreover, given a vector $v$ in the product space $ \F_1 \times \cdots \times \F_p$ and a subset $I$ of $\{1,\dots,p\}$, $v_I$ denotes the vector in $ (\F_i)_{i \in I}$ of elements of $v$ indexed by $I$. Similarly, for a matrix $A$ defined with $p \times p$ blocks adapted to $\mathcal{F}_1,\dots,\F_p$, $A_{IJ}$  denotes the submatrix of  $A$ composed of blocks of $A$ whose rows are in $I$ and columns are in $J$. Moreover, $|J|$ denotes the cardinal of the set $J$ and $|\mathcal{F}|$ denotes the dimension of the Hilbert space $\F$.
 We denote by $1_n$   the $n$-dimensional vector of ones. We denote by $(a)_+ = \max \{0,a\}$ the positive part of a real number $a$.  
 Besides, given matrices $A_1,\dots,A_n$, and a subset $I$ of $\{1,\dots,n\}$, $\Diag(A)_I$ denotes the block-diagonal matrix composed of the blocks indexed by 
$I$.
 Finally, we let denote $\P$ and $\E$ general probability measures and expectations.

\section{Review of Multiple Kernel Learning}
\label{sec:mkl}

We consider the problem a predicting a \emph{response} $Y \in \rb$ from a variable
 $X \in \X$, where $\X$ may be any set of inputs, referred to as the \emph{input space}. In this section, we review the multiple kernel learning framework our paper relies on.
 
 \subsection{Loss Functions}
 \label{sec:loss}

 We assume that we are given $n$ observations of the couple $(X,Y)$, i.e., $(x_{i},y_i) \in \X \times \Y$ for $i =1,\dots,n$.  
 We define the \emph{empirical risk} of a function $f$ from $\mathcal{X}$ to 
$\rb$ as 
$$\frac{1}{n} \sum_{i=1}^n \ell(y_i, f(x_{i})),$$
where $\ell: \rb \times \rb \mapsto \rb^+$ is a \emph{loss function}. We only assume that $\ell$ is     convex with respect to the second parameter (but not necessarily differentiable).

\begin{table}
\begin{center}
\begin{tabular}{|l|l|l|}
\hline
& Loss $\varphi_i(u_i)$ & Fenchel conjugate $\psi_i(\beta_i) $ \\
\hline
Least-squares regression & $\frac{1}{2}(y_i -u_i)^2 $& $\frac{1}{2} \beta_i^2 + \beta_i y_i $ \\
\hline
1-norm support  & $ (|y_i -u_i| -\varepsilon)_+ $ &$  \beta_i y_i + | \beta_i| \varepsilon $ if $|\beta| \leqslant  1$ \\
  vector regression   (SVR) & &  $+\infty$ otherwise \\
\hline
 2-norm support&  $\frac{1}{2}(|y_i -u_i|-\varepsilon)^2_+ $ &  $\frac{1}{2} 
\beta_i^2+ \beta_i y_i + | \beta_i| \varepsilon $  \\
  vector regression  (SVR) & &    \\
 \hline
H\"uber regression &  $\frac{1}{2}(y_i - u_i)^2$ if $|y_i \!-\! u_i| \leqslant \varepsilon\!\!$ &  $ \frac{1}{2} \beta_i^2 + \beta_i y_i$ if $|\beta_i| \leqslant \varepsilon$\\
 & $\varepsilon
 |y_i - u_i| - \frac{\varepsilon^2}{2}$ otherwise &   $+\infty$ otherwise \\
 \hline
 Logistic regression & $ \log(1+\exp(-y_i u_i)) $  & 
 $  (1\!+\!\beta_i y_i) \log(1\!+\!\beta_i y_i)\! - \!\beta_i y_i \log(-\beta_i y_i)\!\!$ \\
  & &   if $\beta_i y_i \in [-1,0]$, $+ \infty$ otherwise \\
  \hline
  1-norm support  & $\max(0,1-y_i u_i)$ & $ y_i \beta_i$
if $\beta_i y_i \in [-1,0]$ \\
vector machine (SVM) & & $+ \infty$ otherwise \\\hline
  2-norm support  & $\frac{1}{2}\max(0,1-y_i u_i)^2$ & $\frac{1}{2} \beta_i^2 + \beta_i y_i$
if $\beta_i y_i  \leqslant 0$ \\
vector machine (SVM) & & $+ \infty$ otherwise \\
\hline
\end{tabular}

\vspace*{-.5cm}

\end{center}
\caption{Loss functions with corresponding Fenchel conjugates, for regression (first three losses, $y_i \in \rb$) and binary classification (last three losses, $y_i \in \{-1,1\}$.}
\label{tab:loss}
\end{table}

\begin{figure}
\begin{center}
\includegraphics[scale=.51]{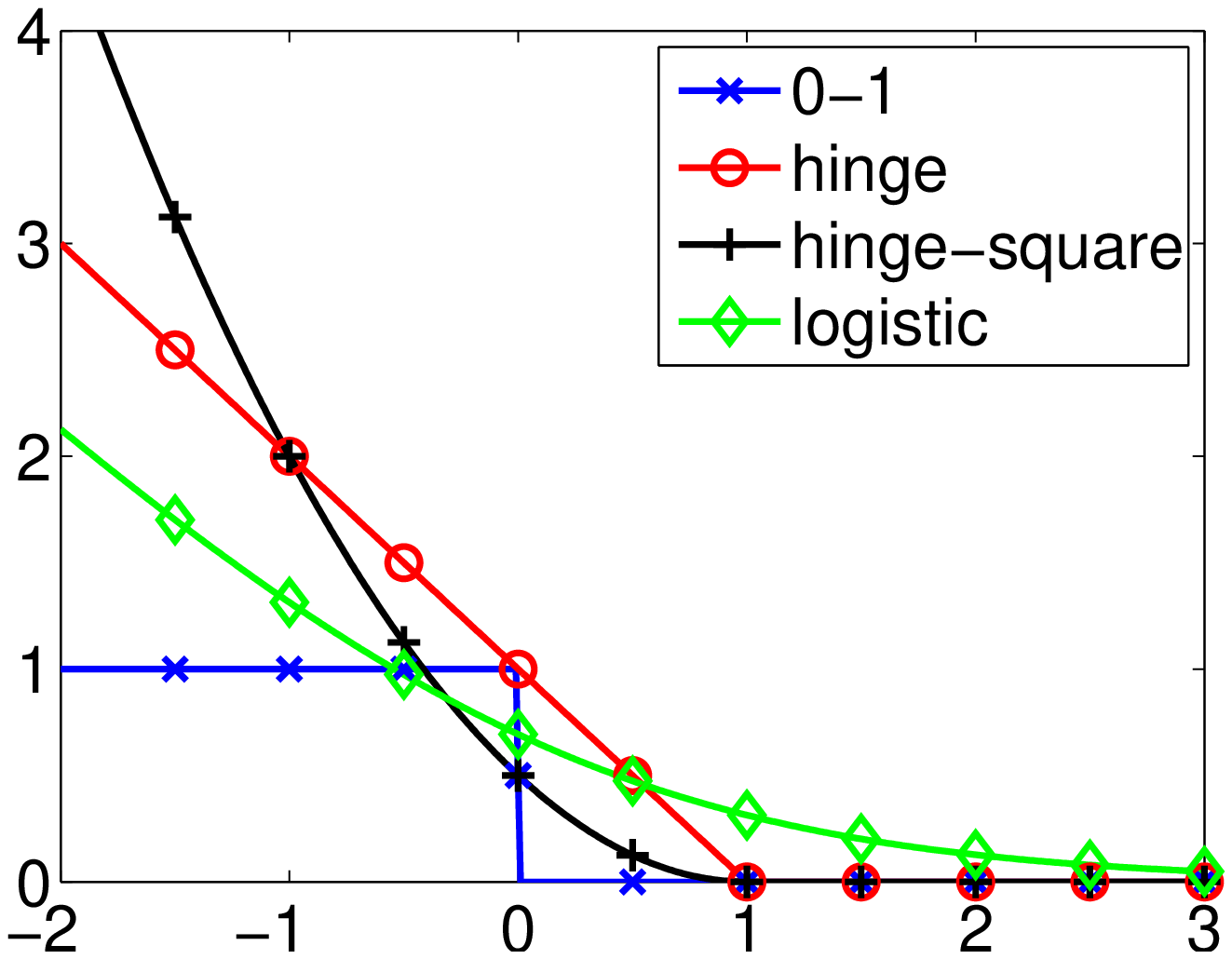} \hspace*{.5cm}
\includegraphics[scale=.51]{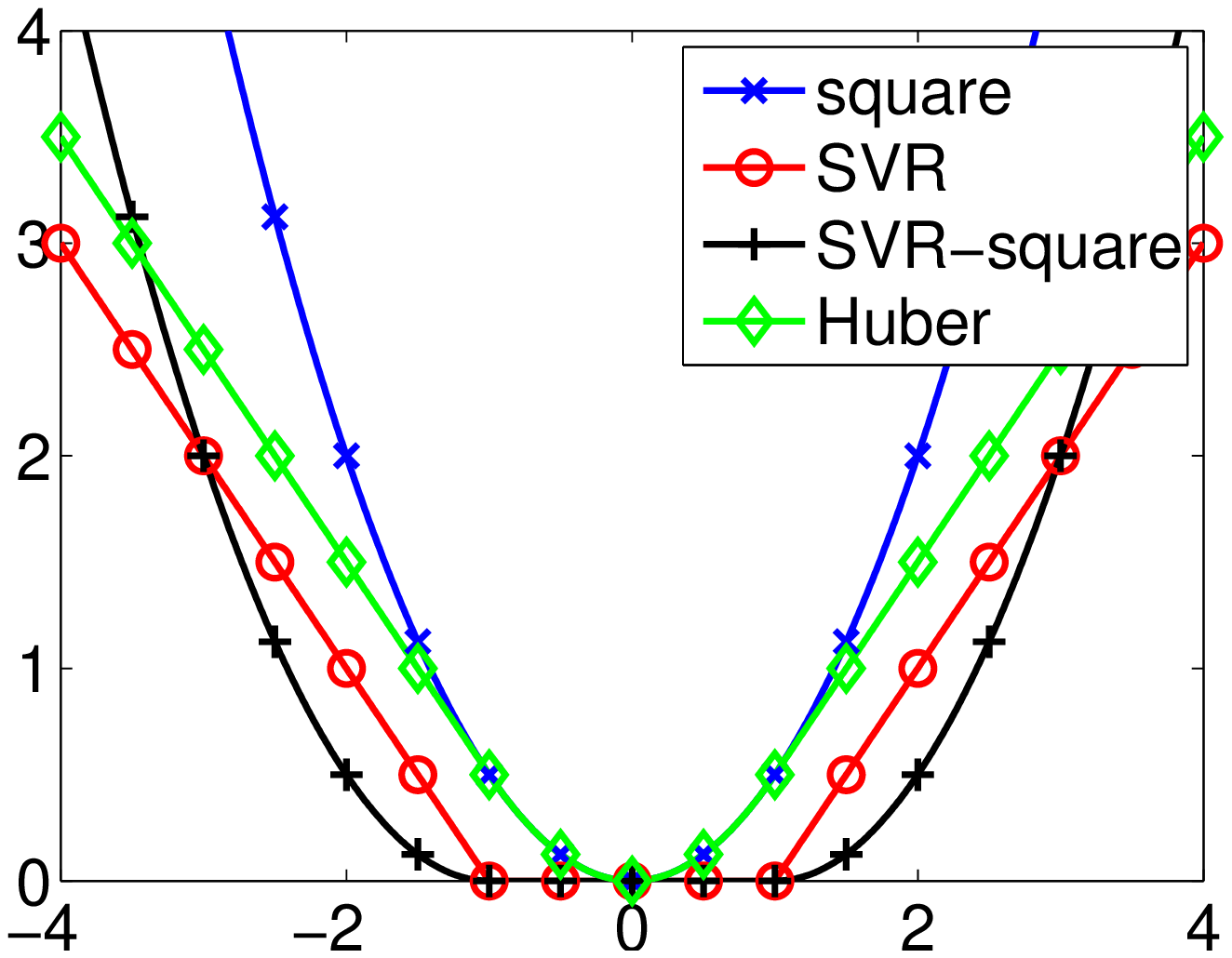}

\vspace*{-.2cm}

\caption{(Left) Losses for binary classification (plotted with $y_i=1$). (Right)   Losses for regression (plotted with $y_i=0$).}
\label{fig:loss}
\end{center}
\end{figure}

Following~\citet{bach_thibaux} and \citet{sonnenburg}, in order to derive optimality conditions for all losses, we need to introduce Fenchel conjugates (see examples in Table~\ref{tab:loss} and \myfig{loss}).
Let $\psi_i:\rb \mapsto \rb$, be the Fenchel conjugate~\citep{boyd} of the convex function $\varphi_i : u_i \mapsto \ell(y_i,u_i )$, defined
 as 
 $$\psi_i(\beta_i) = \max_{u_i \in \rb} \ u_i \beta_i  - \varphi_i(u_i) 
  = \max_{u_i \in \rb} \ u_i \beta_i  - \ell(y_i,u_i) 
 . $$
  The function
 $\psi_i$ is  always convex and, because we have assumed that $\varphi_i$ is convex, we can represent $\varphi_i$ as the Fenchel conjugate of $\psi_i$, i.e., for all $u_i \in \rb$,
 $$
 \ell(y_i,u_i) = \varphi_i(u_i) = \max_{\beta_i \in \rb} \  u_i \beta_i - \psi_i(\beta_i).
 $$
Moreover, in order to include an unregularized constant term, we will need to be able to solve with respect to $b\in \rb$ the following  optimization problem:
\BEQ
\label{eq:intercept}
\min_{b \in \rb} \frac{1}{n} \sum_{i=1}^n\varphi_i(u_i + b).
\EEQ
For $u \in \rb^n$, we let denote by $b^\ast(u)$ any solution of \eq{intercept}. It can either be obtained in closed form (least-squares regression), using Newton-Raphson (logistic regression), or by ordering the values $u_i \in \rb$, $i=1,\dots,n$ (all other piecewise quadratic losses).  In \mysec{optimization}, we study in details losses for which the Fenchel conjugate $\psi_i$ is strictly convex, such as for logistic regression, 2-norm SVM, 2-norm SVR and least-squares regression.

 \subsection{Single Kernel Learning Problem}
In this section, we assume that we are given a positive definite kernel $k(x,x')$ on $\X$. We can then define a reproducing kernel Hilbert space (RKHS) as the completion of the linear span of functions 
 $x \mapsto k(x,x')$ for $x' \in \X$~\citep{rkhs}. We can define the \emph{feature map} $\Phi: \X \mapsto \F$ such that for all $x \in \X$, $f(x) = \langle f,\Phi(x) \rangle$ and for all $x,x' \in \X$, $\Phi(x)(x') = k(x,x')$; we denote by $\|f\|$ the norm of the function $f \in \F$.
  We consider the single kernel learning problem:
 \BEQ
 \label{eq:primal-single}
 \min_{   f\in \F, \ b \in \rb}   \ 
\frac{1}{n} \sum_{i=1}^n \ell\left(y_i,f(x_i) + b \right)
+ \frac{\lambda }{2}
 \|f\|^2.
\EEQ
The following proposition gives its dual, providing a convex instance of the representer theorem~\citep[see, e.g.][and proof in Appendix~\ref{app:single}]{Cristianini2004,smola-book}:
 \begin{proposition}[Dual problem for single kernel learning problem]
 \label{prop:single}
The dual of the optimization problem  in \eq{primal-single} is 
\BEQ
\label{eq:dual-single}
\max_{\alpha \in \rb^n , \ 1_n^\top \alpha = 0}
- \frac{1}{n} \sum_{i=1}^n \psi_i(-n\lambda \alpha_i)  
 - \frac{\lambda}{2}  \alpha^\top K \alpha,
\EEQ
where $K \in \rb^{n \times n}$ is the kernel matrix defined as $K_{ij} = k(x_i,x_j)$. The unique primal solution $f$ can be found from an optimal $\alpha$ as 
$f =   \sum_{i=1}^n \alpha_i \Phi(x_i)$, and $b=b^\ast(K \alpha)$.
\end{proposition}
Note that if the Fenchel conjugate is strictly convex or if the kernel matrix is invertible, then the dual solution $\alpha$ is also unique. In \eq{dual-single}, the kernel matrix $K$ may be replaced by its \emph{centered} version
$$\Kc =  \Big( \idm - \frac{1}{n} 1_n 1_n^\top \Big) K  \Big( \idm - \frac{1}{n} 1_n 1_n^\top \Big),$$
 defined as the kernel matrix of the centered observed features~\citep[see, e.g.][]{Cristianini2004,smola-book}. Indeed, we have $\alpha^\top \Kc \alpha = \alpha^\top K \alpha$ in \eq{dual-single}; however, in the definition of $b=b^\ast(K \alpha)$, $K$ cannot be replaced by $\Kc$.

Finally, the duality gap obtained from a vector $\alpha \in \rb^n$ such that $1_n^\top \alpha = 0$, and the associated primal candidates from Proposition~\ref{prop:single} is equal to
\BEQ
 { \rm gap }_{\rm kernel} \left( K  , \alpha \right)
 =  
 \frac{1}{n} \sum_{i=1}^n  \varphi_i\left[  (K \alpha)_i + b^\ast(K \alpha) \right]  
 +  {\lambda} \alpha^\top \Kc \alpha
+ \frac{1}{n} \sum_{i=1}^n \psi_i(-n\lambda \alpha_i) .
 \EEQ

 \subsection{Sparse Learning with Multiple Kernels}
 
 We now assume that we are given $p$ different reproducing kernel Hilbert spaces $\F_j$ on $\X$, associated with positive definite kernels $k_j: \X  \times \X \to \rb$, $j=1,\dots,p$, and associated feature maps $\Phi_j:\mathcal{X} \to \mathcal{F}_j$. We consider generalized additive models~\citep{hastie_GAM}, i.e., predictors parameterized by $f = (f_1,\dots,f_p) \in \F = \F_1 \times \dots \times \F_p$ of the form
$$f(x)+b = \sum_{j=1}^p f_j(x) + b = \sum_{j=1}^p \langle f_j , \Phi_j(x) \rangle+ b,$$
where each $f_j \in \mathcal{F}_j$ and $b \in \rb$ is a constant term. We let denote $\|f\|$ the Hilbertian norm of~$f \in \F_1 \times \cdots \times \F_p$, defined as $\|f\|^2 = \sum_{j=1}^p \| f_j \|^2$.

We consider regularizing by the sum of the Hilbertian norms,
$\sum_{j=1}^p \| f_j \|$ (which is not itself a Hilbertian norm), with the intuition that this norm will push some of the functions $f_j$ towards zero, and thus provide data-dependent selection of the feature spaces $\F_j$, $j=1,\dots,p$, and hence selection of the kernels $k_j$, $j=1,\dots,p$. We thus consider the following optimization problem:
\BEQ
\label{eq:NPGL}
\min_{f_1 \in \F_1, \ \dots,f_p \in \F_p,  \ b \in \rb}  \ \frac{1}{n} \sum_{i=1}^n \ell  \bigg(y_i, \sum_{j=1}^p f_j(x_{i}) + b \bigg) +  \frac{\lambda}{2}  \bigg( \sum_{j=1}^p \| f_j \|`
\EEQ
Note that using the squared sum of norms does not change the regularization properties: for all solutions of the problem regularized by $ \sum_{j=1}^p \| f_j \|$, there corresponds a solution of the problem in \eq{NPGL} with a different regularization parameter, and vice-versa~\citep[see, e.g.,][Section 3.2]{borlew}. The previous formulation encompasses a variety of situations, depending on how we set up the input spaces $\X_1,\dots,\X_p$:
\BIT
\item \textbf{Regular $\ell_1$-norm and group $\ell_1$-norm regularization}: if each $\X_j$ is the space of real numbers, then we exactly get back penalization by the $\ell_1$-norm, and for the square loss, the Lasso~\citep{lasso}; if we consider finite dimensional vector spaces, we get back the block $\ell_1$-norm formulation and the group Lasso for the square loss~\citep{grouped}. Our general Hilbert space formulation can thus be seen as a  ``non-parametric group Lasso''.
\item \textbf{``Multiple input space, multiple feature spaces''}: In this section, we assume that we have a single input space $\X$ and multiple feature spaces $\F_1,\dots,\F_p$ defined on the same input space. We could also consider that we have $p$ different input spaces $\X_j$ and one feature space $\F_j$ per $\X_j$,
$j=1,\dots,p$, a situation common in generalized additive models.
We can go from the ``single input space, multiple feature spaces'' view to the  ``multiple input space/feature space pairs''
 view by considering $p$ identical copies
 $\X_1,\dots,\X_p$ or $\X$, while we can go in the other direction using projections from
 $\X = \X_1 \times \cdots \times \X_p$.
\EIT
The sparsity-inducing norm formulation defined in \eq{NPGL} can be seen from several points of views and this has led to interesting algorithmic and theoretical developments, which we review in the next sections. In this paper, we will build on the approach of \mysec{eta}, but all results could be derived through the approach presented in \mysec{conic} and \mysec{sdp}.

\subsection{Learning convex combinations of kernels}
\label{sec:eta}
\cite{pontil-jmlr} and \citet{simpleMKL} show that $$\bigg( \sum_{j=1}^p \| f_j \|  \bigg)^2
= \min_{ \zeta \in \rb_+^p, \ 1_p^\top \zeta  = 1} \sum_{j=1}^p \frac{\| f_j \| ^2}{\zeta_j}, $$
where the minimum is attained at $\zeta_j =  \| f_j \| / \sum_{k=1}^p \| f_k \|   $. This variational formulation of the squared sum of norms
allows to find an equivalent problem to \eq{NPGL}, namely:
\BEQ
\label{eq:NPGL-eta}
\min_{ \zeta \in \rb_+^p, \ 1_p^\top \zeta  = 1 }  \ \  
\min_{f_1 \in \F_1,  \dots,f_p \in \F_p, \ b \in \rb}  \ \frac{1}{n} \sum_{i=1}^n \ell \bigg(y_i, \sum_{j=1}^p f_j(x_{i}) + b \bigg) +   \frac{\lambda}{2} 
\sum_{j=1}^p \frac{\| f_j \|^2}{\zeta_j}.
\EEQ
Given $\zeta \in \rb^p_+$ such that $ 1_p^\top \zeta = 1$,  
using the change of variable $\tilde{f}_j = f_j \zeta_j^{-1/2}$ and $\tilde{\Phi}_j(x) = \zeta_j^{1/2} \Phi_j(x)$, 
$j=1,\dots,p$, the problem in \eq{NPGL-eta} is equivalent to:
$$
\min_{ \zeta \in \rb_+^p, \ 1_p^\top \zeta  = 1 }  \ \  
\min_{\tilde{f} \in \F, \ b \in \rb}\frac{1}{n} \sum_{i=1}^n \ell \big(y_i, \langle \tilde{f}, \tilde{\Phi}(x_{i}) \rangle+b \big) + \frac{\lambda}{2} 
\| \tilde{f} \|^2 
,$$ 
with respect to  $\tilde{f}$. Thus $\tilde{f}$ is the solution of the single kernel learning problem with kernel
$$k(\zeta)(x,x')
= \langle \tilde{\Phi}(x), \tilde{\Phi}(x')
\rangle = \sum_{j=1}^p \langle\zeta_j^{1/2} \Phi_j(x), \zeta_j^{1/2} \Phi_j(x') \rangle 
= \sum_{j=1}^p \zeta_j k_j(x,x').$$
 This shows that the non-parametric group Lasso formulation amounts in fact to learning implicitly a weighted combination of 
kernels~\citep{skm,simpleMKL}. Moreover, the optimal functions $f_j$ can then be computed as $f_j(\cdot) = \zeta_j \sum_{i=1}^n \alpha_i k_j(\cdot,x_{i})$, where the vector $\alpha \in \rb^n$ is \emph{common} to all feature spaces $\F_j$, $j=1,\dots,p$.

\subsection{Conic convex duality}
\label{sec:conic}
One can also consider the convex optimization problem in \eq{NPGL} and derive the convex dual using conic programming~\citep{socp,skm,grouplasso}:
\BEQ
\label{eq:dualMKL}
\max_{ \alpha \in \rb^n, \  1_n^\top \alpha =0 }
\bigg\{
-\frac{1}{n} \sum_{i=1}^n \psi_i( -  n \lambda \alpha_i) 
- \frac{\lambda }{2  } \max_{ j \in \{1,\dots,p\}  } \alpha^\top {\Kc}_j \alpha  \bigg\},
\EEQ
where $\Kc_j$ is the centered  {kernel matrix} associated with the $j$-th kernel. From the optimality conditions for second order cones, one can also get that there exists positive weights $\zeta$ that sum to one, such that $f_j(\cdot) = \zeta_j \sum_{i=1}^n \alpha_i k_j(\cdot,x_{i})$ \citep[see][for details]{skm}. Thus, both the kernel weights $\zeta$ and the solution $\alpha$ of the correspond learning problem can be derived from the solution of a single convex optimization problem based on second-order cones. Note that this formulation may be actually solved for small $n$ with  general-purpose toolboxes for second-order cone programming, although QCQP approaches may be used as well~\citep{genomic_fusion}.

\subsection{Kernel Learning with Semi-definite Programming}
\label{sec:sdp}
There is another way of seeing the same problem. Indeed, the dual problem in \eq{dualMKL} may be rewritten as follows:
\BEQ
\label{eq:dualMKL2}
\max_{ \alpha \in \rb^n, \  1_n^\top \alpha =0 }
\ \ \min_{\zeta \in \rb_+^p, \ 1_p^\top \zeta  =1 } 
 \bigg\{
-\frac{1}{n} \sum_{i=1}^n \psi_i( -  n \lambda \alpha_i) 
- \frac{\lambda}{2  }  \alpha^\top \bigg( \sum_{j=1}^p \zeta_j {\Kc}_j  \bigg)  \alpha  \bigg\},
\EEQ
and by convex duality~\citep{boyd,Rock70} as:
\BEQ
\label{eq:dualMKL3}
\min_{\zeta \in \rb_+^p, \ 1_p^\top \zeta  =1 } \ \ 
\max_{ \alpha \in \rb^n, \  1_n^\top \alpha =0 }
\bigg\{
-\frac{1}{n} \sum_{i=1}^n \psi_i( -  n \lambda \alpha_i) 
- \frac{\lambda}{2  }  \alpha^\top \bigg( \sum_{j=1}^p \zeta_j {\Kc}_j  \bigg) \alpha  \bigg\}.
\EEQ
If we   denote   $G(K) = \max_{ \alpha \in \rb^n, \  1_n^\top \alpha =0 } 
\left\{
-\frac{1}{n} \sum_{i=1}^n \psi_i(  -  n \alpha_i) 
- \frac{\lambda }{2  }  \alpha^\top \Kc \alpha  \right\},
$ the optimal value of the single kernel learning problem in \eq{primal-single} with loss $\ell$ and kernel matrix $K$ (and centered kernel matrix $\widetilde{K}$), then the multiple kernel learning problem is equivalent to minimizing $G(K)$ over convex combinations of the $p$ kernel matrices associated with all $p$ kernels, i.e., equivalent to minimizing $B(\zeta) = G(\sum_{j=1}^p \zeta_j K_j)$.

This function $G(K)$, introduced by several authors in slightly different contexts
 \citep{gert,pontil-jmlr,hyperkernels}, leads to  a more general kernel learning framework where one can learn more than simply convex combinations of kernels---in fact, any kernel matrix which is positive semi-definite.  In terms of theoretical analysis, results from general kernel classes may be brought to bear~\citep{gert,srebro,ying}; however, the special case of convex combination allows the sparsity interpretation and some additional theoretical analysis~\citep{grouplasso,kolt}.
The practical and theoretical advantages of allowing more general potentially non convex combinations (not necessarily with positive coefficients) of kernels is still an open problem and subject of ongoing work~\citep[see, e.g.,][and references therein]{Varma09}.

Note that  regularizing in \eq{NPGL} by the sum of squared norms $\sum_{j=1}^p \| f_j\|^2$ (instead of the squared sum of norms), is equivalent to considering the sum of kernels matrices, i.e., $K = \sum_{j=1}^p K_j$. Moreover, if all kernel matrices have rank one, then the kernel learning problem is equivalent to an $\ell_1$-norm problem, for which dedicated algorithms are usually much more efficient~\citep[see, e.g.,][]{lars,descent}.

\subsection{Algorithms}

The multiple facets of the multiple kernel learning problem have led to multiple algorithms. The first ones were based on the minimization of $B(\zeta) = G(\sum_{j=1}^p \zeta_j K_j)$ through general-purpose toolboxes for semidefinite programming~\citep{gert,hyperkernels}. While this allows to get a solution with high precision, it is not scalable to medium and large-scale problems. Later, approaches based on conic duality and smoothing were derived~\citep{skm,bach_thibaux}. They were based on existing efficient techniques for the support vector machine (SVM) or potentially other supervised learning problems, namely sequential minimal optimization~\citep{platt}. Although they are by design scalable, they require to recode existing learning algorithms and do not reuse pre-existing implementations. The latest formulations based on the direct minimization of a cost function that depends directly on $\zeta$ allow to reuse existing code~\citep{sonnenburg,simpleMKL} and may thus benefit from the intensive optimizations and tweaks already carried through. Finally, active set methods have been recently considered for finite groups~\citep{roth2,obozinski-joint}, an approach we extend to hierarchical kernel learning in \mysec{activeset}.

\section{Hierarchical Kernel Learning (HKL) }
\label{sec:hkl}

We now extend the multiple kernel learning framework to kernels which are indexed by vertices in a directed acyclic graph. We first describe examples of such graph-structured positive definite kernels from \mysec{graph} to \mysec{features}, and defined the graph-adapted norm in \mysec{graphreg}.

\subsection{Graph-Structured Positive Definite Kernels}
\label{sec:graph}

We assume that we are given a \emph{positive definite kernel} $k:\X \times \X \to \rb$, and that this kernel can be expressed as the sum, over an index set $V$, of basis kernels $k_v$, 
$v\in V$, i.e., for all $x,x' \in \X$:
$$k(x,x') = \sum_{v \in V} k_v(x,x').$$
 For each $v\in V$, we   denote by $\mathcal{F}_v$ and $\Phi_v$ the 
feature space and feature map of $k_v$, i.e., for all $x,x' \in \X$,
 $k_v(x,x') = \langle \Phi_v(x), \Phi_v(x') \rangle$.

Our sum assumption corresponds to a situation where the feature map $\Phi(x)$ and feature space~$\mathcal{F}$  for $k$ are the \emph{concatenations} of the feature maps $\Phi_v(x)$ and feature spaces $\F_v$ for each kernel $k_v$, i.e., 
$\F = \prod_{v \in V} \F_v$ and $\Phi(x) = ( \Phi_v(x) )_{v \in V}$. Thus, looking for a certain $f \in \F$ and a predictor function
$f(x) = \langle f, \Phi(x) \rangle$ is equivalent to looking jointly for
$f_v \in \F_v$, for all $v \in V$, and 
$$f(x) = \langle f , \Phi(x) \rangle  = \sum_{v \in V} 
 \langle f_v , \Phi_v(x) \rangle. $$

 As mentioned earlier, we make the assumption that the set $V$ can be embedded into a \emph{directed acyclic graph}\footnote{Throughout this paper, for simplicity, we use the same notation to refer to the graph and its set of vertices.}. Directed acyclic graphs (referred to as DAGs) allow to naturally define the 
notions of \emph{parents}, \emph{children}, \emph{descendants} and \emph{ancestors}~\citep{graph}.  
Given a node $w \in V$, we   denote by $\anc(w) \subset V$ the set of its ancestors,
and by $\des(w)\subset V$, the set of its descendants. We use the convention that any $w$ is a descendant and an ancestor of itself, i.e., $w \in \anc(w)$ and $w \in \des(w)$.
 Moreover, for $W \subset V$, we let denote $\sou(W)$ the set of \emph{sources} (or \emph{roots}) of the graph $V$ restricted to $W$, that is, nodes in  $W$ with no parents belonging to $W$.  
 
\begin{figure}
\begin{center}

\includegraphics[scale=.73]{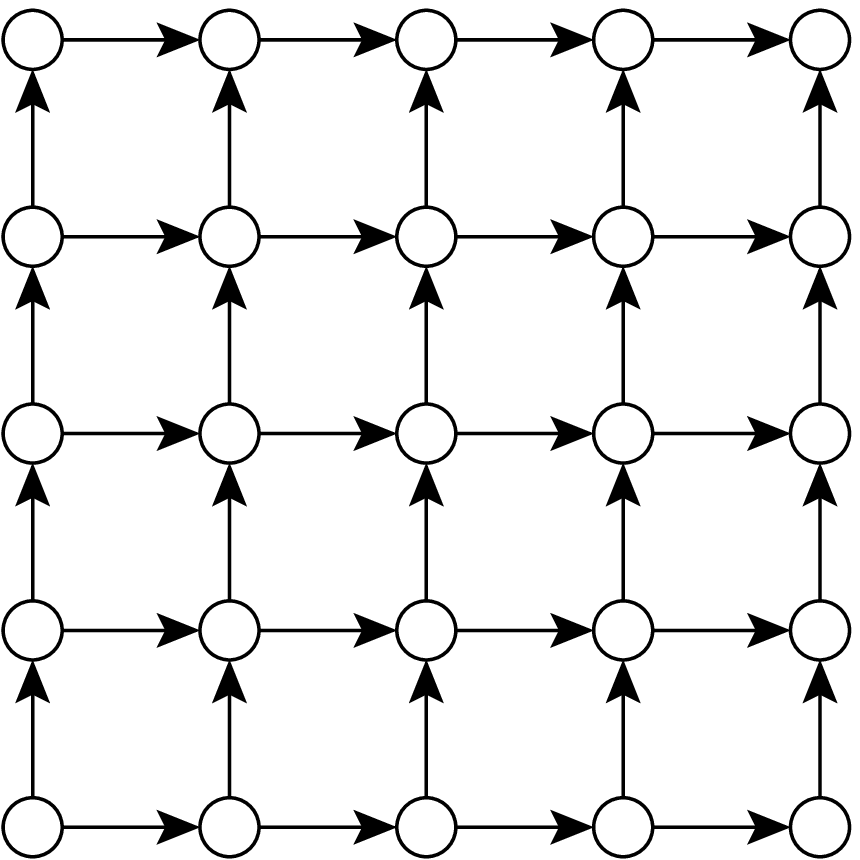} \hspace*{2cm}
\includegraphics[scale=.73]{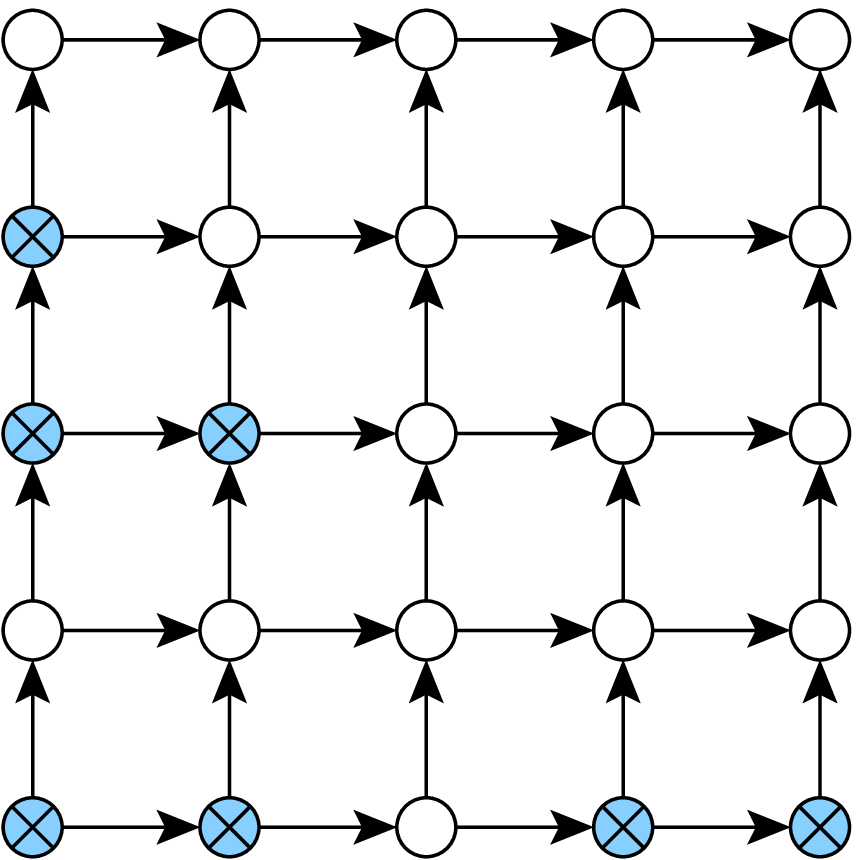} \\[1.5cm]
\includegraphics[scale=.73]{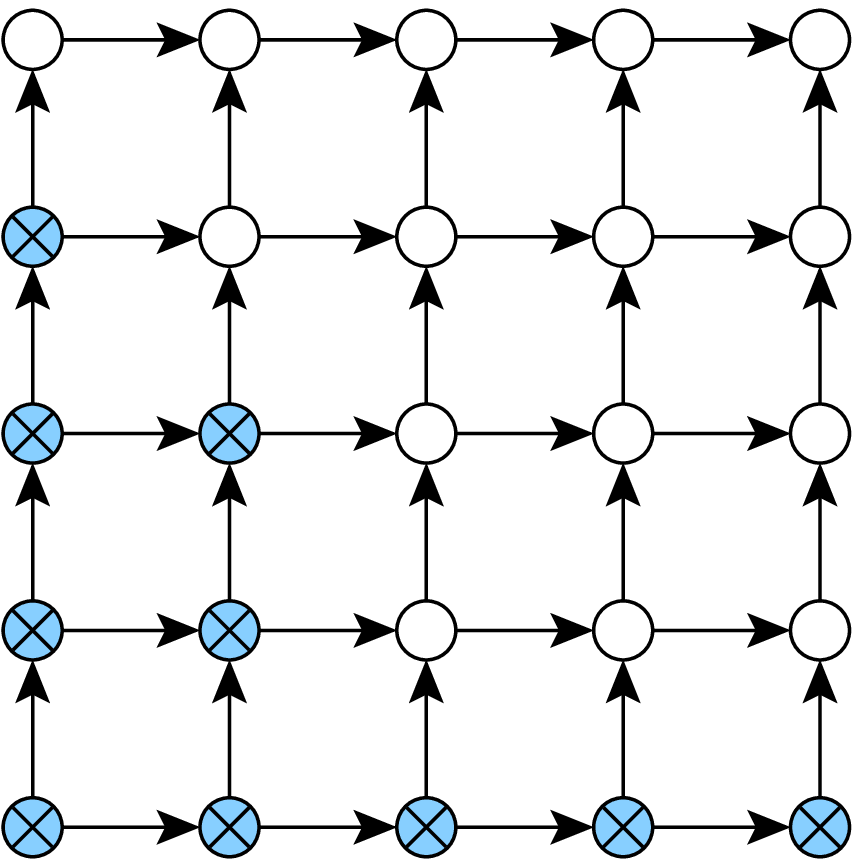}   \hspace*{2cm}
\includegraphics[scale=.73]{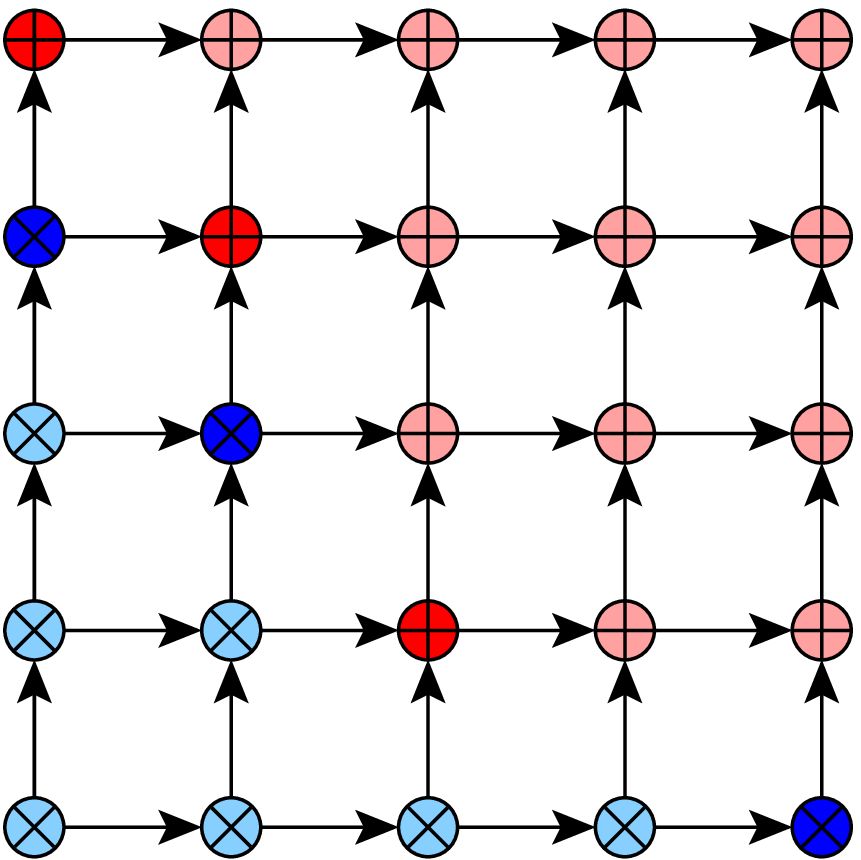}

 \end{center}

\caption{Examples of directed acyclic graphs (DAGs) and associated notions: (top left) 2D-grid (number of input variables $p=2$, maximal order in each dimension $q=4$);  (top right) example of sparsity pattern which is not equal to its hull ($\times$ in light blue) and (bottom left) its hull ($\times$ in light blue); (bottom right) dark blue points ($\times$) are extreme points of the set of all active points (blue $\times$); dark red points ($+$) are the sources of the complement of the hull (set of all red $+$). Best seen in color.
}
\label{fig:2d}
\end{figure}

 Moreover, given a subset of nodes $W \subset V$, we can define the \emph{hull} of $W$ as the union of all ancestors of $w \in W$, i.e.,
$$\hull(W) = \bigcup_{w \in W} A(w).$$
 Given a set $W$, we define the set of \emph{extreme points}  (or \emph{sinks}) of $W$ as the smallest subset $T \subset W $ such that $\hull(T) = \hull(W)$; it is always well defined, as 
 (see \myfig{2d} for   examples of these notions):
 $$\sinks(W) = \bigcap_{T \subset V, \
\hull(T) = \hull(W)} T.$$

\begin{figure}
\begin{center}

\hspace*{-.5cm}
\includegraphics[scale=.5]{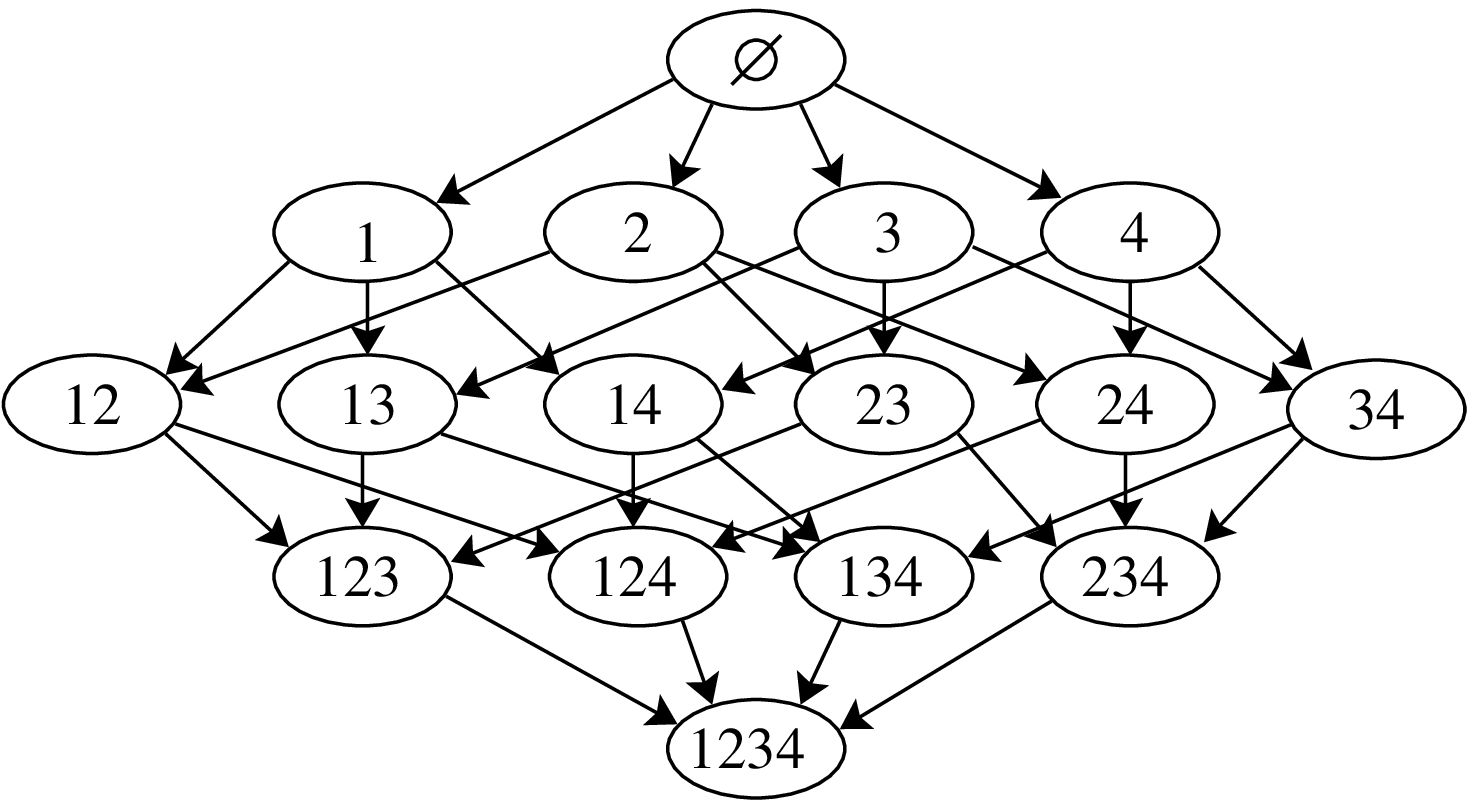} 
\hspace*{.25cm}
\includegraphics[scale=.5]{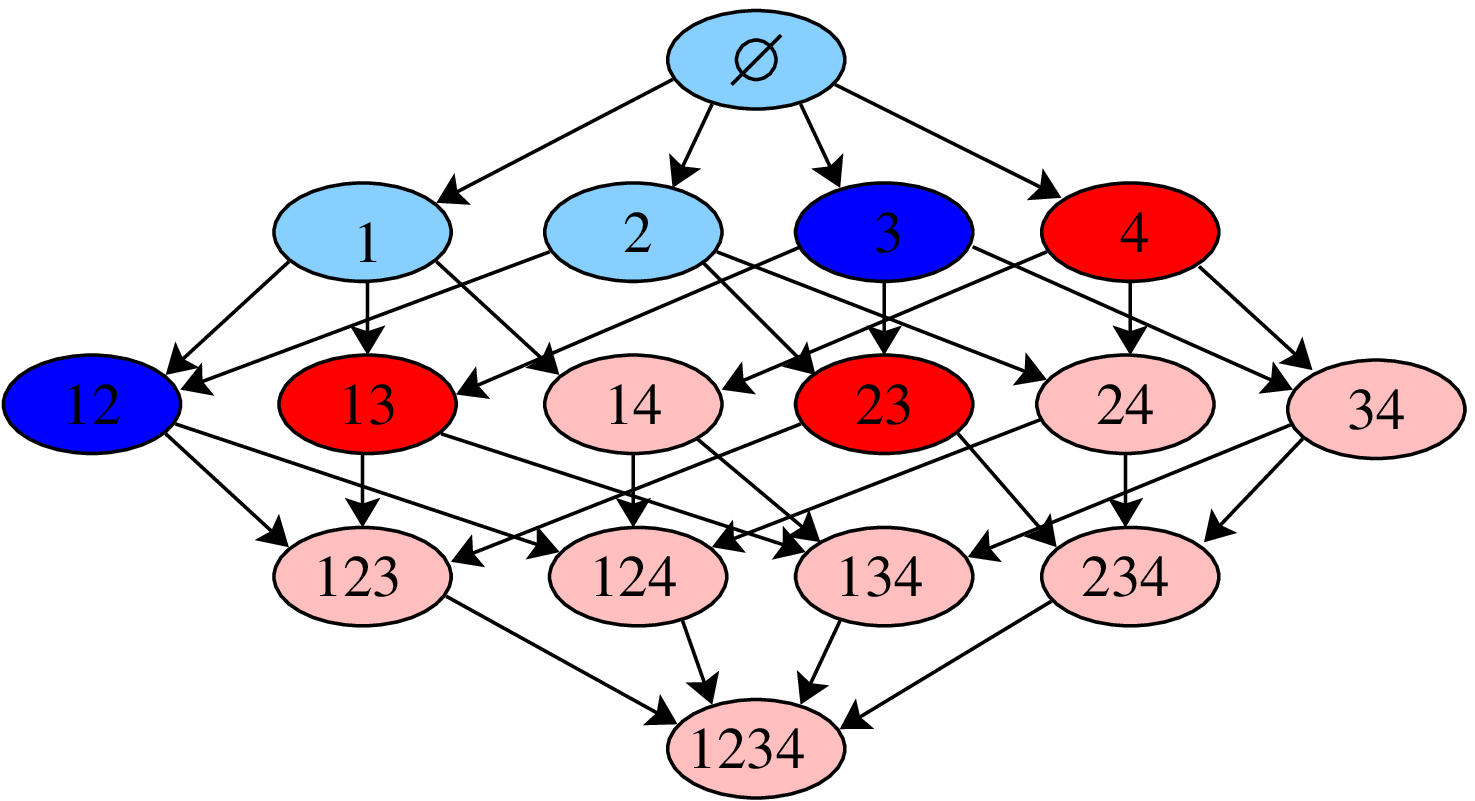} 
\hspace*{-.5cm}
 \end{center}

\vspace*{-.65cm}

\caption{Directed acyclic graph  of subsets of size 4: (left) DAG of subsets ($p=4$, $q=1$); (right)  example of sparsity pattern (light and dark blue), dark blue points are extreme points of the set of all active points; dark red points are the sources of the set of all red points. Best seen in color.
}
\label{fig:subsets}
\end{figure}

The goal of this paper is to perform kernel selection among the kernels $k_v$, $v \in V$. We essentially use the graph to limit the search to specific subsets of $V$. 
Namely, instead of considering all possible subsets of active (relevant) vertices, we 
will consider active sets of vertices which are equal to their hulls, i.e., subsets that contain the ancestors of all their elements, thus limiting the search space (see \mysec{graphreg}).

\subsection{Decomposition of Usual Kernels in Directed Grids}
\label{sec:decompositions}
\label{sec:kernels}
\begin{figure}
  \begin{center}
   \hspace*{-.2cm}
   \includegraphics[scale=.57]{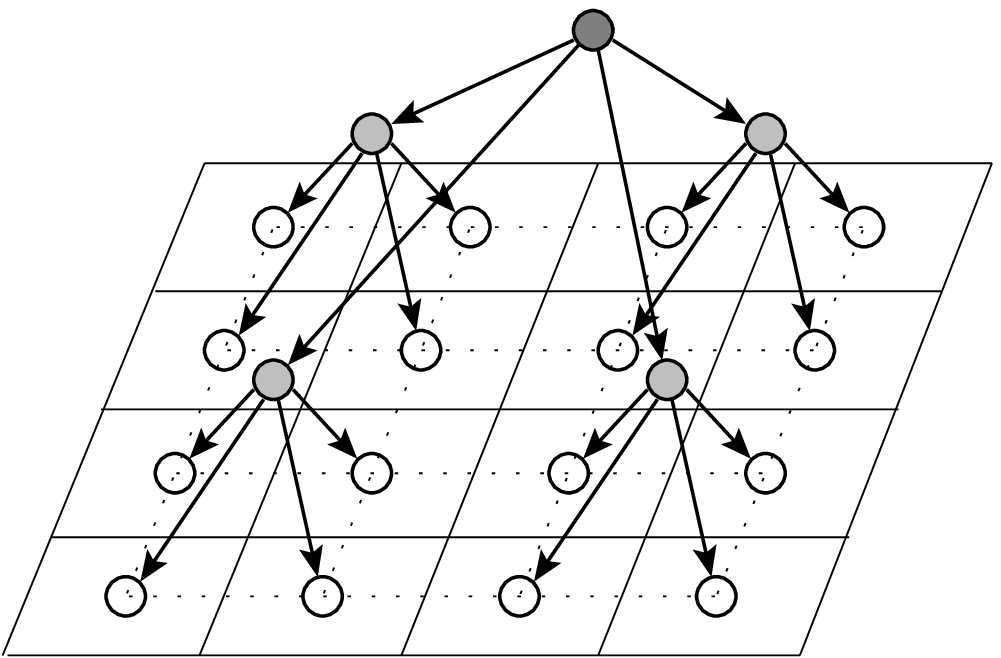} \hspace*{-.525cm}  \includegraphics[scale=.6]{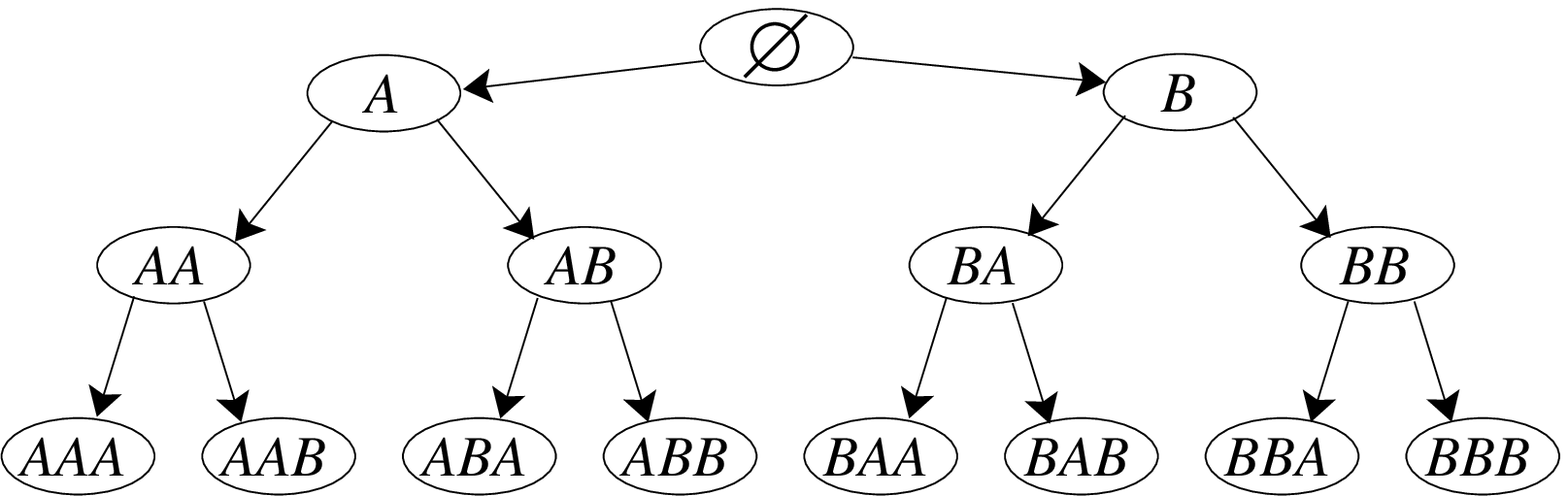} \hspace*{-.2cm}
       \end{center}

\vspace*{-.7cm}

  \caption{Additional examples of discrete structures. Left: pyramid over an image; a region is selected only after all larger regions that contains it are selected.  Right: set of substrings of size 3 from the alphabet $\{A,B\}$; in bioinformatics~\citep{kernel_methodsvert2004} and text processing~\citep{stringkernels}, occurence of certain potentially long strings is an important feature and considering the structure may help selecting among the many possible strings.}
  \label{fig:otherexamples}
  \end{figure}

In this paper, we primarily focus on kernels that can be expressed as ``products of sums'', and on the associated $p$-dimensional directed grids, while noting that our framework is applicable to many other kernels (see, e.g., \myfig{otherexamples}). Namely, we assume that the input space $\X$ factorizes into $p$ components $\X = \X_1 \times \cdots  \times \X_p$ and that
we are given $p$ sequences of length $q+1$ of kernels
$k_{ij}(x_i,x_i')$, $i \in \{1,\dots,p\}$, $j \in \{0,\dots,q\}$, such that (note the implicit different conventions for indices in $k_i$ and $k_{ij}$):
\BEQ
\label{eq:decompose}
k(x,x') = \prod_{i=1}^p k_i(x_i,x_i')
= \prod_{i=1}^p  \bigg(\sum_{j=0}^q k_{ij}(x_i,x_i') \bigg)= \sum_{j_1,\dots,j_p=0}^q \ \prod_{i=1}^p k_{ij_i}(x_i,x_i')
.\EEQ
Note that in this section and the next section, $x_i$ refers to the $i$-th component of the tuple $x=(x_1,\dots,x_p)$ (while in the rest of the paper, $x_i$ is the $i$-th observation, which is itself a tuple).
 We thus have   a sum of $(q+1)^p$ kernels, that can be computed efficiently as a product of $p$ sums of $q+1$ kernels.
A natural DAG on $V = \{0,\dots,q\}^p$
is defined by connecting each $(j_1,\dots,j_p)$ respectively
to $(j_1\!+\!1,j_2,\dots,j_p)$, $\dots$,
$(j_1,\dots,j_{p-1}, j_p\! +\!1)$ as long as $j_1 < q, \dots, j_p <q$, respectively . As shown in \mysec{graphreg}, this DAG (which has a single source) will correspond to the constraint of selecting a given product of kernels only after all the subproducts are selected.
Those DAGs are especially suited to non-linear variable selection, in particular with the polynomial, Gaussian and spline kernels. In this context, products of kernels correspond to interactions between certain variables, and our DAG constraint implies that \emph{we select an interaction only after all sub-interactions were already selected}, a constraint that is similar to the one used in multivariate additive splines~\citep{mars}.

\paragraph{Polynomial kernels.}
 We consider $\X_i = \rb$, 
 $k_i(x_i,x_i') = ( 1 + x_i x_i')^q$ and for all $j \in \{0,\dots,q\}$,
 $k_{ij}(x_i,x_i') ={ q \choose j}  (x_i x_i')^{j}$;  the full kernel is then equal to 
 $$k(x,x') = \prod_{i=1}^p ( 1 + x_i x_i')^q =  \sum_{j_1,\dots,j_p=0}^q\  \prod_{i=1}^p{ q \choose j_i}  (x_i x_i')^{j_i} . $$ 
 Note that this is not exactly the usual polynomial kernel $(1+x^\top x')^q$ (whose
feature space is the space of multivariate polynomials of \emph{total} degree less than $q$), since our kernel considers  polynomials  of \emph{maximal} degree $q$.

\paragraph{Gaussian kernels (Gauss-Hermite decomposition).} 
 We also consider $\X_i = \rb$, and the Gaussian-RBF kernel $  e^{-b(x_i -x_i')^2}$ with $b>0$. The following decomposition is   
 the eigendecomposition of the non centered covariance operator corresponding to a normal distribution with variance $1/4a$~\citep[see, e.g.,][]{williams00effect,grouplasso}:
\BEQ
\label{eq:gaussian}
 e^{-b(x_i-x_i')^2}
\! = \!  \left( \!1\! -\! \frac{b^2}{A^2} \right)^{-1/2}\! \sum_{j=0}^{\infty} \! \frac{(b/A)^j}{2^j j!}  e^{-\frac{b}{A}(a+c) x_i^2  }
H_j( \sqrt{2c} x_i) 
e^{-\frac{b}{A}(a+c) (x_i')^2  }
H_j( \sqrt{2c} x_i') ,
\EEQ 
where $c^2 = a^2 + 2ab$, $A = a+b+c$, and $H_j$ is the $j$-th Hermite polynomial~\citep{szego}. By appropriately
truncating the sum, i.e., by considering that the first $q$ basis kernels are obtained from the first $q$   Hermite polynomials, and the $(q+1)$-th kernel is summing over all other kernels, we obtain a decomposition of a uni-dimensional Gaussian kernel into $q+1$ components (the first $q$ of them are one-dimensional, the last one is infinite-dimensional, but can  be computed by differencing).
The decomposition ends up being close to a polynomial kernel of infinite degree, modulated by an exponential~\citep{Cristianini2004}. One may also use an \emph{adaptive} decomposition using kernel PCA~\citep[see, e.g.][]{Cristianini2004,smola-book}, which is equivalent to using the eigenvectors of the empirical covariance operator associated with the data (and not the population one associated with the Gaussian distribution with same variance). In prior work~\citep{hkl}, we tried both with no significant differences.

\paragraph{All-subset Gaussian kernels.} 
When $q=1$, the directed grid is isomorphic to the power set (i.e., the set of subsets, see \myfig{subsets}) with the DAG defined as the Hasse diagram of the partially ordered set of all subsets~\citep{cameron1994ctt}. In this setting, we can decompose the all-subset Gaussian kernel~\citep[see, e.g.,][]{Cristianini2004} as:
$$
\prod_{i=1}^p ( 1 +  \alpha e^{-b(x_i-x'_i)^2} )= 
 \sum_{ J \subset \{1,\dots,p\} } \prod_{i \in J} \alpha
e^{-b (x_i-x'_i)^2 } 
=  \sum_{ J \subset \{1,\dots,p\} } \alpha^{|J|}
e^{-b \| x_J - x_J' \|^2},
$$ and our framework will select the relevant subsets for the Gaussian kernels, with the DAG presented in \myfig{subsets}. A similar decomposition is considered by~\citet{cosso}, but only on a subset of the power set. Note that the DAG of subsets is different from the ``kernel graphs'' introduced for the same type of kernel by~\citet{Cristianini2004} for expliciting the computation of polynomial kernels and ANOVA kernels.

\paragraph{Kernels on structured data.}
Although we mainly focus on directed grids in this paper, many kernels on structured data can also be naturally decomposed through a hierarchy (see \myfig{otherexamples}), such as the pyramid match kernel and related kernels~\citep{grauman,cuturi}, string kernels or graph kernels~\citep[see, e.g.,][]{Cristianini2004}. The main advantage of using $\ell_1$-norms inside the feature space, is that the method will adapt the complexity to the problem, by only selecting the right order of complexity from  exponentially many features.

\subsection{Designing New Decomposed Kernels}
 
As shown in \mysec{consistency}, the problem is well-behaved numerically and statistically if there is not too much correlation between the various feature maps $\Phi_v$, $v \in V$. Thus, kernels such as the the all-subset Gaussian kernels may not be appropriate as each feature space contains the feature spaces of its ancestors\footnote{More precisely, this is true for the closures of these spaces of functions.}. Note that a strategy we could follow would be to remove some contributions of all ancestors by appropriate orthogonal projections. We now design specific kernels for which the feature space of each node is orthogonal to the feature spaces of its ancestors (for well-defined dot products).

\paragraph{Spline kernels.} 
In \eq{decompose}, we may chose, with $q=2$:
\BEAS
k_{i0}(x_i,x_i') & = &  1 \\
 k_{i1}(x_i,x_i') &  = &  x_i x_i' \\
 k_{i2}(x_i,x_i')  & = &  \min\{|x_i|,|x_i'|\}^2( 3 \max\{|x_i|,|x_i'|\} - \min \{|x_i|,|x_i'|\} ) /6
 ,\mbox{ if } x_i x'_i \geqslant 0 \\
 & = & 0, \mbox{ otherwise} ,
 \EEAS
 leading to tensor products of one-dimensional cubic spline kernels~\citep{wahba,gu}. This kernel has the advantage of (a) being parameter free and (b) explicitly starting with linear features and essentially provides a convexification of multivariate additive regression splines~\citep{mars}. Note that
it may be more efficient here to use natural splines in the estimation method~\citep{wahba} than using kernel matrices.
  
\paragraph{Hermite kernels.}
We can start from the following identity, valid for $\alpha<1$ and from which the decomposition of the Gaussian kernel in \eq{gaussian} may be obtained~\citep{szego}:
$$
\sum_{j=0}^\infty \frac{ \alpha^j }{ j! 2^j  }H_j(x_i) H_j(x_i') 
=  (1-\alpha^2)^{-1/2}
\exp\left(  \frac{ - 2  \alpha (x_i - x_i')^2}{1-\alpha^2}    + \frac{(x_i^2+(x_i')^2) \alpha}{1+\alpha} \right).
$$
We can then define a sequence of kernel which also starts with linear kernels:
\BEAS
k_{i0}(x_i,x_i') & = &  H_0(x) H_0(x') = 1 \\
 k_{ij}(x_i,x_i')  & = &  \frac{\alpha^j}{2^j j!} H_j(x) H_j(x') \mbox{ for } j \in \{1,\dots,q-1\} \\
  k_{iq}(x_i,x_i')  & = &\sum_{j=q}^\infty \frac{ \alpha^j }{ j! 2^j  }H_j(x_i) H_j(x_i') .
 \EEAS

Most kernels that we consider in this section (except the polynomial kernels) are universal kernels~\citep{universal1,universal2}, that is, on a compact set of $\rb^p$, their reproducing kernel Hilbert space is dense in $L^2(\rb^p)$. This is the basis for the universal consistency results in \mysec{universal}. Moreover, some kernels such as the spline and Hermite kernels explicitly include the linear kernels inside their decomposition: in this situation, the sparse decomposition will start with linear features.
In \mysec{universal}, we briefly study the universality of the kernel decompositions that we consider.

\subsection{Kernels or Features?}  
\label{sec:features}
In this paper, we emphasize the \emph{kernel view}, i.e., we assume we are given a positive definite kernel (and thus a feature space) and we explore it using $\ell_1$-norms. Alternatively, we could use the \emph{feature view}, i.e., we would assume that we have a large structured set of features that we try to select from; however, the techniques developed in this paper assume that (a) each feature might be infinite-dimensional and (b) that we can sum all the local kernels efficiently (see in particular \mysec{reduced}). Following the kernel view thus seems slightly more natural, but by no means necessary---see~\citet{jenatton} for a more general ``feature view'' of the problem.

In order to apply our optimization techniques in the feature view, as shown in \mysec{optimization}, we simply need a specific \emph{upper bound} on the kernel to be able to be computed efficiently.
More precisely, we need to be able to compute $\sum_{ w \in \des(t)}  
\left(\sum_{v \in \anc(w) \cap \des(t) } d_{v}  \right)^{-2} K_w$ for all $t \in V$, or
an upper bound thereof, for appropriate weights (see \mysec{reduced} for further details).

\subsection{Graph-Based Structured Regularization}

\label{sec:graphreg}
Given $f \in \prod_{v \in V} \mathcal{F}_v$, the natural Hilbertian norm
$\|f\|$ is defined through
$\|f\|^2 = \sum_{v \in V } \| f_v \|^2$. Penalizing with this norm is efficient because
summing all kernels $k_v$ is assumed feasible in polynomial time and we can bring to bear the usual kernel machinery; however, it does not lead to sparse solutions, where many $f_v$ will be exactly equal to zero, which we try to achieve in this paper.

We use the DAG to limit the set of active patterns to certain configurations, i.e., sets which are equal to their hulls, or equivalenty sets which contain all ancestors of their elements. If we were using a regularizer such as $\sum_{v\in V} \| f_v\|$ we would get sparse solutions, but the set of active kernels would be scattered throughout the graph and would not lead to optimization algorithms which are sub-linear in the number of vertices $|V|$.

All sets which are equal to their hull can be obtained by removing all the  descendants of certain vertices. Indeed, the hull of a set $I$
is characterized by the set of $v$, such that $\des(v) \subset I^c$, i.e., such that all descendants of $v$ are in the complement $I^c$ of $I$: $$\hull(I)= \{ v \in V, \  \des(v) \subset I^c\}^c . $$
Thus, if we try to estimate a set $I$ such that $\hull(I)=I$, we thus need to determine which $v \in V$ are such that
$\des(v) \subset I^c$. In our context, we   are hence looking at selecting  vertices $v \in V$ for which $f_{\des(v)} = (f_w)_{w \in \des(v)} = 0$.
We thus consider the following  structured block $\ell_1$-norm defined  on $\F = \F_1 \times \cdots \times \F_p$ as
\BEQ
\label{eq:omega}
\Omega(f) = \sum_{v \in V} d_v \| f_{\des(v)} \|
= \sum_{v \in V} d_v  \bigg( \sum_{w \in \des(v)} \| f_w\|^2  \bigg)^{1/2}
,
\EEQ
where $(d_v)_{v \in V}$ are strictly positive weights.
 We assume that for all vertices but the sources of the DAG, we have $d_v = \beta^{ \depth(v) }$ with $\beta>1$, where $\depth(v)$ is the depth of node $v$, i.e., the length of the smallest path to the sources. We denote by $d_r \in (0,1]$ the common weights to all sources. 
 Other weights could be considered, in particular, weights inside the blocks $\des(v)$~\citep[see, e.g.][]{jenatton}, or weights that lead to penalties closer to the Lasso (i.e., $\beta <1$), for which the   effect of the DAG would be weaker.
 Note that when the DAG has no edges, we get back the usual block $\ell_1$-norm with uniform weights $d_r$, and thus, the results presented in this paper (in particular the algorithm presented in \mysec{algorithm} and non-asymptotic analysis presented in \mysec{nonasymptotic}) can be applied to multiple kernel learning.

\begin{figure}
\begin{center}

\hspace*{-.5cm}
\includegraphics[scale=.5]{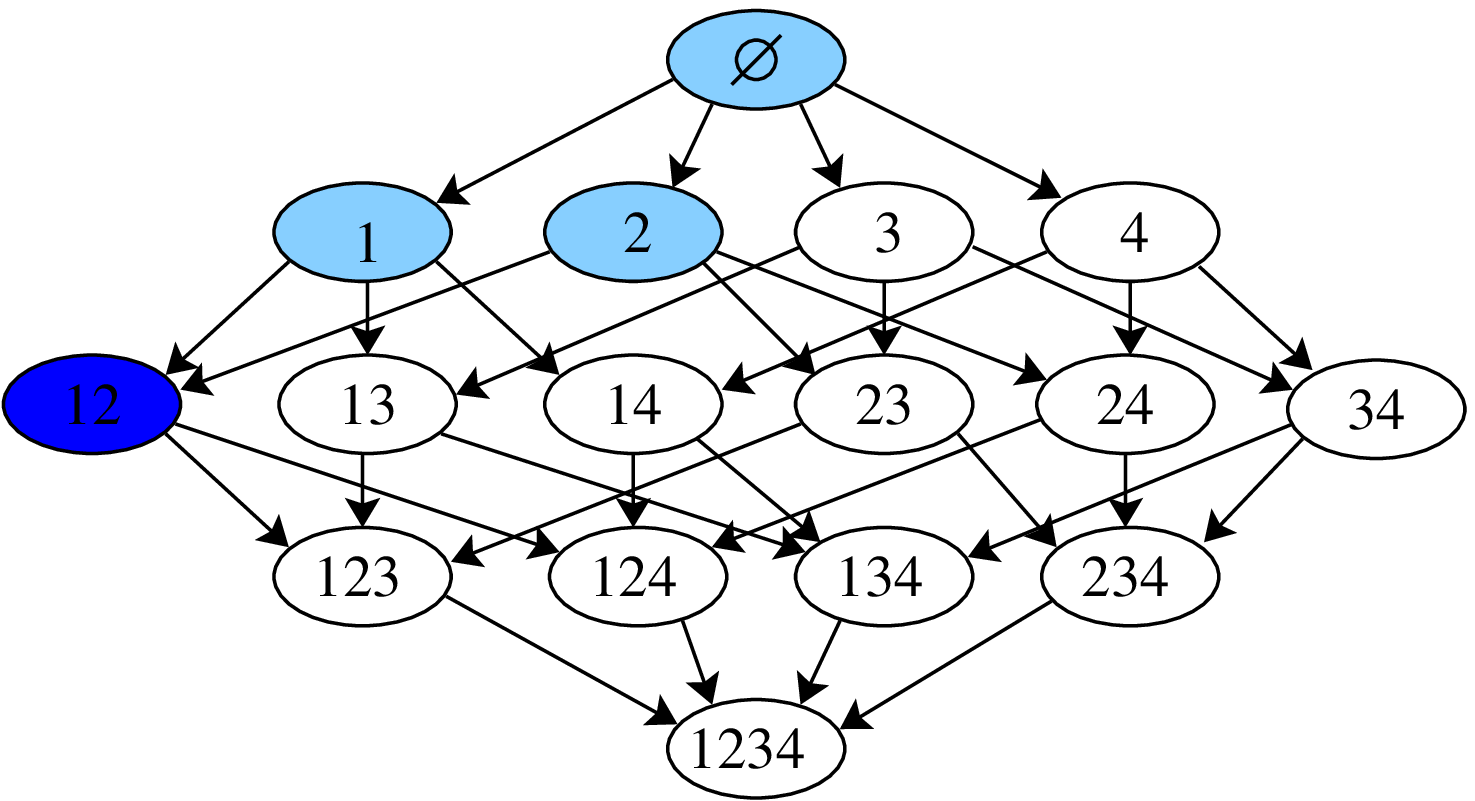} 
\hspace*{.25cm}
\includegraphics[scale=.5]{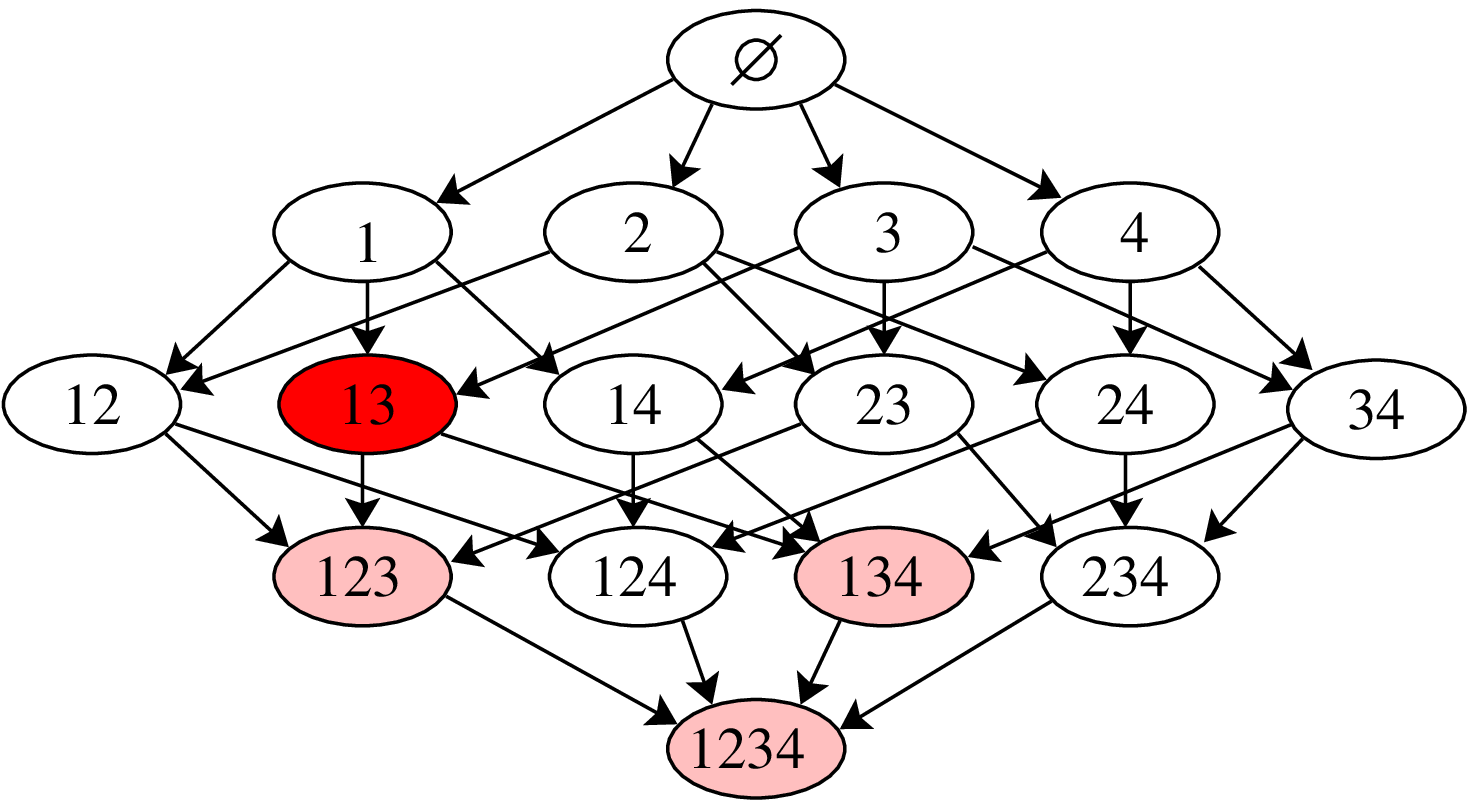} 
\hspace*{-.5cm}
 \end{center}

\vspace*{-.5cm}

\caption{Directed acyclic graph  of subsets of size 4: (left) a vertex (dark blue) with its ancestors (light blue), (right) a vertex (dark red) with its descendants (light red).  By zeroing out weight vectors associated with descendants of several nodes, we always obtained a set of non-zero weights which contains all of its own ancestors (i.e., the set of non-zero weights is equal to its hull).
}
\label{fig:norm}
\end{figure}

Penalizing by such a norm will indeed impose that some of the vectors
$f_{\des(v)} \in \prod_{w \in \des(v)} \mathcal{F}_w$ are exactly zero, and we show in \mysec{allowed} that these are the only patterns we might get. We thus consider the following minimization problem\footnote{Following~\citet{skm} and \mysec{mkl}, we consider the square of the norm, which does not change the regularization properties, but allow simple links with multiple kernel learning.}:
 \BEQ
\label{eq:milasso2}
\min_{   f \in \prod_{v\in V}\! \mathcal{F}_v , \ b \in \rb} \ \
\frac{1}{n} \sum_{i=1}^n \ell \bigg(y_i,  \sum_{v \in V} \langle f_v  , \Phi_v(x_i) \rangle + b \bigg)
+ \frac{\lambda }{2} \bigg(  \sum_{v \in V}  d_v \| f_{\des(v)}  \| \bigg)^2.
\EEQ
Our   norm is a Hilbert space instantiation of the hierarchical norms recently introduced by~\citet{cap}. If all Hilbert spaces are finite dimensional, our particular choice of norms corresponds to an  ``$\ell_1$-norm of $\ell_2$-norms''. While with uni-dimensional groups or kernels, the ``$\ell_1$-norm of $\ell_\infty$-norms'' allows an efficient path algorithm for the square loss and when the DAG is a tree~\citep{cap}, this is not possible anymore with groups of size larger than one, or when the DAG is not a tree~\citep[see][for examples on two-layer hierarchies]{marie}. In \mysec{optimization}, we propose a novel algorithm  to solve the  associated optimization problem in polynomial time in the number of selected groups or kernels, for all group sizes, DAGs and losses. Moreover, in \mysec{consistency}, we show under which conditions a solution to the problem in \eq{milasso2} consistently estimates the hull of the sparsity pattern.

\section{Optimization}
\label{sec:optimization}

In this section, we give optimality conditions  for the
problems in  \eq{milasso2}, as well as optimization algorithms with polynomial time complexity in the number of selected kernels. In simulations, we consider total numbers of kernels up to $4^{256}$, and thus such efficient algorithms that can take advantage of the sparsity of solutions are essential to the success of hierarchical multiple kernel learning (HKL).

\subsection{Reformulation in terms of Multiple Kernel Learning}

Following~\citet{simpleMKL}, we can simply derive an equivalent formulation of \eq{milasso2}. Using  Cauchy-Schwarz inequality, we have that for all $\eta \in \rb_+^V$
such that   $ \sum_{v \in V}  d_v^2 \eta_v \leqslant 1$, a variational formulation of $\Omega(f)^2$ defined in \eq{omega}:
\BEAS
\Omega(f) ^2 &  = &  \bigg( \sum_{v \in V} d_v \| f_{\des(v)} \| \bigg)^2 
= \bigg( \sum_{v \in V} ( d_v \eta_v^{1/2})  \frac{ \| f_{\des(v)}\|}{ \eta_v^{1/2}} \bigg)^2 \\
&
\leqslant  &  \sum_{v \in V} d_v^2 \eta_v \times  \sum_{v \in V}
\frac{ \| f_{\des(v)}\|^2}{ \eta_v}
\leqslant \sum_{w \in V}\! \bigg( \sum_{v \in \anc(w) } \eta_v^{-1} \bigg) \| f_w\|^2,
\EEAS
 with equality if and only if for all $v \in V$ $\eta_v  = d_v^{-1} \| f_{\des(v)}\|  \left(\sum_{w \in V} d_w
 \| f_{\des(w) } \| \right)^{-1}= \frac{ d_v^{-1} \| f_{\des(v)} \|}{\Omega(f)}$.

 We associate to the vector $\eta \in \rb^V_+$, the vector $\zeta \in \rb^V_+$ such that
\BEQ
\label{eq:zeta}
\forall w \in V, \ \zeta_w(\eta)^{-1}  =   \sum_{v \in \anc(w)} \eta_v^{-1}.
\EEQ
We use the natural convention that if $\eta_v$ is equal to zero, then $\zeta_w(\eta)$ is equal to zero for all  descendants  $w$ of $v$. We let denote $H = \{ \eta \in \rb_+^V, \ \sum_{v \in V} d_v^2 \eta_v \leqslant 1 \}$ the set of allowed $\eta$ and $Z = \{ \zeta(\eta), \ \eta \in H \}$ the set of all associated $\zeta(\eta) $ for $\eta \in H$. The set $H$ and $Z$ are in bijection, and we can interchangeably use $\eta
\in H$ or the corresponding $\zeta(\eta) \in Z$.
Note that $Z$ is in general not convex (unless the DAG is a tree, see Proposition~\ref{prop:tree} in Appendix~\ref{app:tree}), and if $\zeta \in Z$, then $\zeta_w \leqslant \zeta_{v}$ for all $w \in \des(v)$, i.e., weights of descendant kernels are always smaller, which is consistent with the known fact that \emph{kernels should always be selected after all their ancestors} (see \mysec{allowed} for a precise statement).

 The problem in \eq{milasso2} is thus equivalent to 
\BEQ
\label{eq:milasso2-eq}
 {
\min_{ \eta \in H}
\min_{   f \in \prod_{v\in V}\! \mathcal{F}_v , \ b \in \rb} \ \frac{1}{n} \sum_{i=1}^n \ell \bigg(y_i, \sum_{v \in V} \langle f_v , \Phi_v(x_i) \rangle  +b \bigg)
+ \frac{\lambda }{2}
\sum_{w \in V}  \zeta_w(\eta)^{-1} \|f_w\|^2}.
\EEQ
From \mysec{mkl}, we know that at the optimum, $f_w = \zeta_w(\eta) \sum_{i=1}^n \alpha_i \Phi_w(x_i) \in \F_w$, where $\alpha \in \rb^n$ are the dual parameters associated with the single kernel learning problem in Proposition~\ref{prop:single}, with kernel matrix $\sum_{w \in V} \zeta_w(\eta) K_w$.

Thus, the solution is entirely determined by $\alpha \in \rb^n$ and $\eta \in H \subset \rb^V$ (and its corresponding $\zeta(\eta) \in Z$). We also associate to $\alpha$ and $\eta$ the corresponding functions $f_w$, $w \in V$, and optimal constant $b$, for which we can check optimality conditions.
More precisely, we have (see proof in Appendix~\ref{app:gap}):

\begin{proposition}[Dual problem for HKL]
\label{prop:dual-hkl}
The convex optimization problem in \eq{milasso2} has the following dual problem:
\BEQ
\label{eq:dual-hkl}
\max_{\alpha \in \rb^n , \ 1_n^\top \alpha = 0}- \frac{1}{n} \sum_{i=1}^n \psi_i(-n\lambda \alpha_i)  
 - \frac{\lambda}{2} \max_{\eta \in H}\sum_{w \in V} \zeta_w(\eta)  \alpha^\top  \Kc_w \alpha. \EEQ
 Moreover, at optimality, $\forall w \in V, f_w\! =\! \zeta_w(\eta) \sum_{i=1}^n \alpha_i \Phi_w(x_i)$ and $b = b^\ast\left( \sum_{w \in V} \zeta_w(\eta) K_w \alpha\right)$, with $\eta$ attaining, given 
 $\alpha$, the maximum of $\sum_{w \in V} \zeta_w(\eta)  \alpha^\top  \Kc_w \alpha$.
\end{proposition}
\begin{proposition}[Optimality conditions for HKL]
\label{prop:gap}
Let 
$(\alpha,\eta ) \in \rb^{n} \times H$, such that $1_n^\top \alpha=0$. Define functions $f \in \F$ through $\forall w \in V, f_w\! =\! \zeta_w (\eta) \sum_{i=1}^n \alpha_i \Phi_w(x_i)$ and $b = b^\ast\left( \sum_{w \in V} \zeta_w(\eta) K_w \alpha\right)$ the corresponding constant term.  The vector of functions $f$ is optimal for \eq{milasso2}, if and only if  :\BIT
\item[(a)]  given $\eta \in H$,   the vector $\alpha$ is optimal for the single kernel learning problem with kernel matrix 
 $K = \sum_{w\in V} \zeta_w(\eta)  K_w$,  
 \item[(b)]
  given $\alpha$, $\eta \in H $ maximizes 
  \BEQ
  \label{eq:etat}
  \sum_{w \in V}  \left( \textstyle \sum_{v \in \anc(w) } \eta_v^{-1}  \right)^{-1} \!\!\!\alpha^\top \Kc_w \alpha
  = \displaystyle \sum_{w \in V}  \zeta_w(\eta)\alpha^\top \Kc_w \alpha
  .\EEQ
  \EIT
 \end{proposition}
 
  Moreover, as shown in Appendix~\ref{app:gap}, the total duality gap can be upperbounded as the sum of the two separate duality gaps for the two optimization problems, which will be useful in \mysec{reduced} for deriving sufficient conditions of optimality (see Appendix~\ref{app:gap} for more details):
  \BEQ
  \label{eq:gap}
   { \rm gap }_{\rm kernel} \bigg( \sum_{w \in V}  \zeta_w(\eta)  \Kc_w , \alpha \bigg)
+\frac{\lambda}{2} { \rm gap }_{\rm weights} \big(  (\alpha^\top \Kc_w \alpha)_{w \in V}, \eta \big),
  \EEQ
  where $ { \rm gap }_{\rm weights}$ corresponds to the duality gap of \eq{etat}.
Note that in the case of ``flat'' regular  multiple kernel learning, where the DAG has no edges, we obtain back usual optimality conditions~\citep{simpleMKL,pontil-jmlr}.

 Following a common practice for convex sparse problems~\citep{ng-sparsecoding,roth2}, we will try to solve a small problem where we assume we know the set of $v$ such that $\| f_{\des(v)}\|$ is equal to zero (\mysec{small-dual}). We then need   (a) to check that variables in that set may indeed be left out of the solution, and (b) to propose variables to be added if the current set is not optimal. In the next section, we show that this can be done in polynomial time although the number of kernels to consider leaving out is exponential (\mysec{reduced}).
 
 Note that an alternative approach would be to consider the regular multiple kernel learning problem with additional linear constraints  $\zeta_{\pi(v)} \geqslant  \zeta_v$ for all non-sources $v \in V$. However, it would not lead to the analysis through sparsity-inducing norms outlined in \mysec{consistency} and might not lead to polynomial-time algorithms.

\subsection{Conditions for Global Optimality of Reduced Problem}

\label{sec:reduced}
We consider a subset $W$ of $V$ which is equal to its hull---as shown in \mysec{allowed}, those are the only possible active sets. We  consider the optimal solution $f$ of the reduced problem (on $W$), namely,
\BEQ
\label{eq:milasso-reduced}
\min_{ f_{W} \in \prod_{v\in W }\! \mathcal{F}_v, \ b \in \rb}  \
\frac{1}{n} \sum_{i=1}^n \ell \bigg(y_i,  \sum_{v \in W } \langle f_v , \Phi_v(x_i)  \rangle + b \bigg)
+ \frac{\lambda }{2} \bigg(  \sum_{v \in W}  d_v \| f_{\des(v) \cap W}  \| \bigg)^2,
\EEQ
with optimal primal variables $f_W$, dual variables $\alpha \in \rb^n$ and optimal pair $(\eta_W,\zeta_W)$. From these, we can construct a full solution $f$ to the problem, as $f_{W^c}=0$, with $\eta_{W^c}= 0$. That is, we keep $\alpha$ unchanged and add zeros to $\eta_W$.

We now consider necessary conditions and sufficient conditions for this augmented solution  to
be optimal with respect to the full problem in \eq{milasso2}. We denote by $ \Omega(f) = 
\sum_{v \in W} d_v \| f_{D(v) \cap W } \|$  the optimal value of the norm for the reduced problem.
\begin{proposition}[Necessary optimality condition] 
\label{prop:CNopt}
If the reduced solution is optimal for the full problem in \eq{milasso2} and all kernels indexed by $W$ are active, then we have:
\BEQ
\label{eq:CNopt}
  \max_{t \in \sou(W^c) }    \frac{\alpha^\top \Kc_t \alpha}
{ d_t^2 } \leqslant
\Omega(f)^2 .
\EEQ
\end{proposition}

\begin{proposition}[Sufficient optimality condition]  
\label{prop:CSopt}
If 
\BEQ
\label{eq:CSopt}
\max_{ t \in \sou(W^c)}    
\sum_{w \in \des(t)}  \frac{\alpha^\top \Kc_w \alpha }{  
(\sum_{v \in \anc(w) \cap \des(t) } d_{v} )^2}
 \leqslant  \Omega(f)^2 + 2\varepsilon/\lambda
,\EEQ
 then the total duality gap in \eq{gap} is less than $  \varepsilon$.
\end{proposition}
The proof is fairly technical and can be found in Appendix~\ref{app:CSCN}; this result constitutes the main technical result of the paper: it essentially allows to design an algorithm for solving a  large optimization problem over exponentially many dimensions in polynomial time.  Note that when the DAG has no edges, we get back regular conditions for  unstructured MKL---for which \eq{CNopt} is equivalent to \eq{CSopt} for $\varepsilon=0$.
 
  The necessary condition in Eq.~(\ref{eq:CNopt}) does not cause any computational problems as the number of sources of $W^c$, i.e., the cardinal of $\sou(W^c)$, is upper-bounded by $|W|$ times the maximum out-degree of the DAG. 
  
  However, the sufficient condition in Eq.~(\ref{eq:CSopt}) requires to sum over all descendants of the active kernels, which is impossible without special structure (namely exactly being able to compute that sum or an upperbound thereof). Here, we need to bring to bear the specific structure of the full kernel $k$. In the context of directed grids we consider in this paper, if $d_v$ can also be decomposed as a product, then $\sum_{v \in \anc(w) \cap \des(t)} d_v $ can also be factorized, and we can compute the sum over all $v \in \des(t)$ in linear time in $p$. Moreover, we can cache the sums 
$$ \breve{K}_t  = 
\sum_{w \in \des(t)}   
\left( \textstyle \sum_{v \in \anc(w) \cap \des(t) } d_{v} \right)^{-2}  \displaystyle \Kc_w 
$$ in order to save running time in the active set algorithm presented in \mysec{algorithm}. Finally, in the context of directed grids, many of these kernels are either constant across iterations, or change slightly; that is, they are product of sums, where most of the sums are constant across iterations, and thus computing a new cached kernel can be considered of complexity $O(n^2)$, independent of the DAG and of $W$.

\subsection{Dual Optimization for Reduced or Small Problems}
\label{sec:small-dual}
In this section, we consider solving \eq{milasso2} for DAGs $V$ (or active set $W$) of small cardinality, i.e., for (very) small problems or for the reduced problems obtained from the algorithm presented in Figure~\ref{fig:kernelsearch} from \mysec{algorithm}.

When kernels $k_v$, $v \in V$, have low-dimensional feature spaces,  either by design
   (e.g., rank one if each node of the graph corresponds to a single feature), or after a low-rank decomposition such as a singular value decomposition or an incomplete Cholesky factorization~\citep{fine01efficient,csi}, we may use a ``primal representation'' and solve the problem in \eq{milasso2} using generic optimization toolboxes adapted to conic constraints~\citep[see, e.g.,][]{cvx}. With high-dimensional feature spaces, in order to reuse existing optimized supervised learning code and use  high-dimensional kernels, it is preferable to use a ``dual optimization''. Namely, we follow~\citet{simpleMKL}, and consider for $\zeta \in Z$, the function 
$$B(\zeta)= G(K(\zeta)) =  
\min_{   f\in \prod_{v\in V} \! \mathcal{F}_v, \ b \in \rb} \   
\frac{1}{n} \sum_{i=1}^n \ell \bigg(y_i, \sum_{v \in V} \langle f_v , \Phi_v(x_i) \rangle + b \bigg)
+ \frac{\lambda }{2}
\sum_{w \in V}  \zeta_w^{-1} \|f_w\|^2,$$ which is the optimal value of the single kernel learning problem with kernel matrix $\sum_{w \in V} \zeta_w K_w$. Solving \eq{milasso2-eq} is equivalent to minimizing $B(\zeta(\eta))$ with respect to $\eta \in H$.

 If the Fenchel conjugate of the loss is strictly convex (i.e., square loss, logistic loss, H\"uber loss, 2-norm support vector regression),  then the function $B$ is differentiable---because the dual problem in \eq{dual-single} has a unique solution $\alpha$ \citep{bonnans2000perturbation}. 
When the Fenchel conjugate is not strictly convex, a ridge (i.e., positive diagonal matrix) may be added to the kernel matrices, which has the exact effect of smoothing the loss---see, e.g.,~\cite{yosida} for more details on relationships between smoothing and adding strongly convex functions to the dual objective function.

 Moreover, the function $\eta \mapsto \zeta(\eta)$ is differentiable on $(\rb_+^\ast)^V$, but not at any points $\eta$ such that one $\eta_v$ is equal to zero. Thus,
the function $\eta \mapsto 
 B[ \zeta(  (1-\varepsilon)  \eta+ \frac{\varepsilon}{|V|}  d^{-2}  )]$ , where $d^{-2}$ is the vector with elements $d_v^{-2}$, is differentiable if $\varepsilon>0$, and its derivatives can simply be obtained from the chain rule. 
 In simulations, we use $\varepsilon = 10^{-3}$; note that adding this term is equivalent to smoothing the norm $\Omega(f)$ (i.e., make it differentiable), while retaining its sparsity-inducing properties (i.e., some of the optimal $\eta$ will still be exactly zero).

 We can then use the same projected gradient descent strategy as~\citet{simpleMKL} to minimize it. The overall complexity of the algorithm is then proportional to $O(|V| n^2)$---to form the kernel matrices---added to the complexity of solving a single kernel learning problem---typically between $O(n^2)$ and $O(n^3)$, using proper kernel classification/regression algorithms~\citep{simplesvm1,simplesvm2}.
 Note that we could follow the approach of~\citet{rakotochapelle} and consider second-order methods for optimizing with respect to $\eta$.

\subsection{Kernel Search Algorithm}

\label{sec:algorithm}
\label{sec:activeset}
We now  present the detailed algorithm which extends the search algorithm of~\citet{ng-sparsecoding} and~\citet{roth2}. Note that the kernel matrices are never all needed explicitly, i.e., we only need them (a) explicitly to solve the small problems (but we need only a few of those) and (b)   implicitly to compute the necessary condition in Eq.~(\ref{eq:CNopt}) and the sufficient condition
in Eq.~(\ref{eq:CSopt}), which requires to sum over all kernels which are not selected, as shown in \mysec{reduced}.

The algorithm works in two phases: first the (local) necessary condition is used to check optimality of the solution and add variables; when those are added, the augmented reduced problem must include the new variable into the active set. Once the necessary condition is fulfilled, we use the sufficient condition, which essentially sums over all non selected kernels and makes sure that if some information is present further away in the graph, it will indeed be selected. See \myfig{kernelsearch} for details\footnote{Matlab/C code for least-squares regression and logistic regression may be downloaded from the author's website.}.

\begin{figure}
\begin{center}
\begin{tabular}{l}
\hline
\\[-.3cm]
\parbox[t]{12cm}{\textbf{Input}: Kernel matrices $K_v  \in \rb^{n \times n}$, weights $d_v$, $v \in V$, maximal gap $\varepsilon$,  maximal number of kernels $Q$.}
\\[.2cm]
\textbf{Algorithm}: \\
\hspace{.5cm} 1. Initialization: active set $W = \varnothing$, cache kernel matrices $\breve{K}_w$, $w \in \sou(W^c)$   \\
\hspace{.5cm} 2. Compute $(\alpha,\eta)$ solutions of
\eq{milasso-reduced}, obtained using \mysec{small-dual} (with gap $\varepsilon$) \\
\hspace{.5cm} 3.  While necessary condition in Eq.~(\ref{eq:CNopt}) is not satisfied and $|W|\leqslant Q$\\
\hspace{1cm}  a.   Add violating kernel in $\sou(W^c)$ to $W$ \\
\hspace{1cm}  b.  Compute $(\alpha,\eta)$ solutions of \eq{milasso-reduced}, obtained using \mysec{small-dual}  (with gap $\varepsilon$) \\
\hspace{1cm}  c. Update cached kernel matrices $\breve{K}_w$, $w \in \sou(W^c)$   \\
\hspace{.5cm} 4.  While sufficient condition in Eq.~(\ref{eq:CSopt}) is not satisfied and $|W| \leqslant Q$\\
\hspace{1cm}  a.   Add violating kernel in $\sou(W^c)$ to $W$ \\
\hspace{1cm}  b.  Compute $(\alpha,\eta)$ solutions of \eq{milasso-reduced}, obtained using \mysec{small-dual}  (with gap $\varepsilon$)   \\
\hspace{1cm}  c. Update cached kernel matrices $\breve{K}_w$, $w \in \sou(W^c)$  
\\[.05cm]
\textbf{Output}:  $W$, $\alpha$, $\eta$, constant term $b$ \\
\hline
\end{tabular}

\caption{Kernel search algorithm for hierarchical kernel learning.  The algorithm stops either when the duality gap is provably less than $2\varepsilon$, either when the maximum number of active kernels has been achieved; in the latter case, the algorithm may or may not have reached a $2 \varepsilon$-optimal solution (i.e., a solution with duality gap less than $2 \varepsilon$). }
\label{fig:kernelsearch}
\end{center}
\end{figure}

\begin{figure}
\begin{center}

\includegraphics[scale=.73]{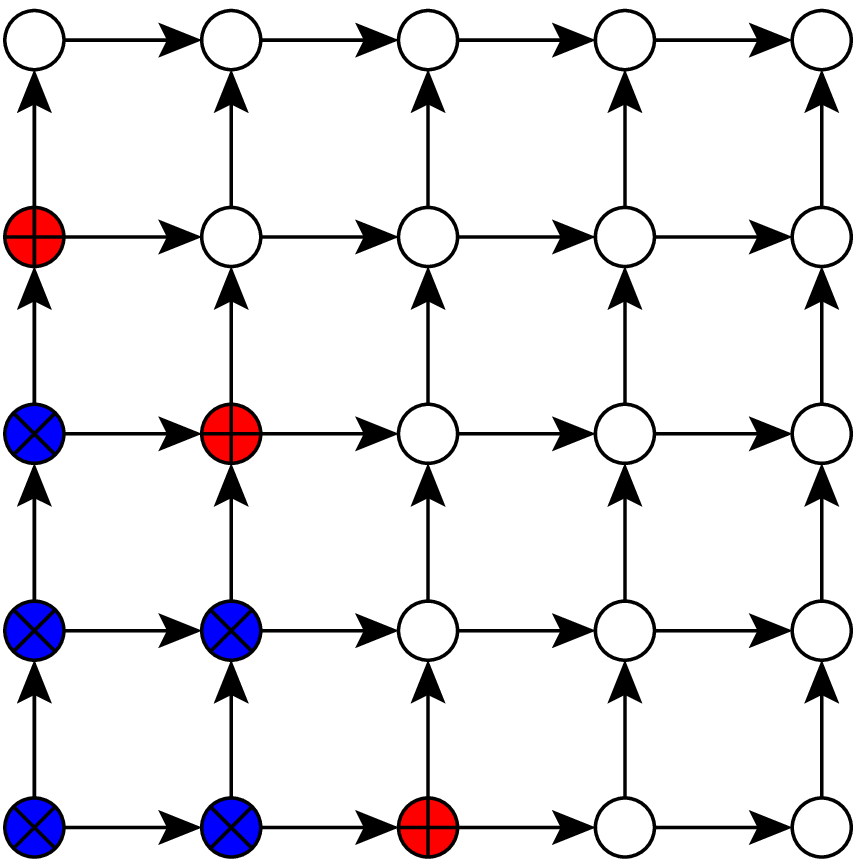} \hspace*{2cm}
\includegraphics[scale=.73]{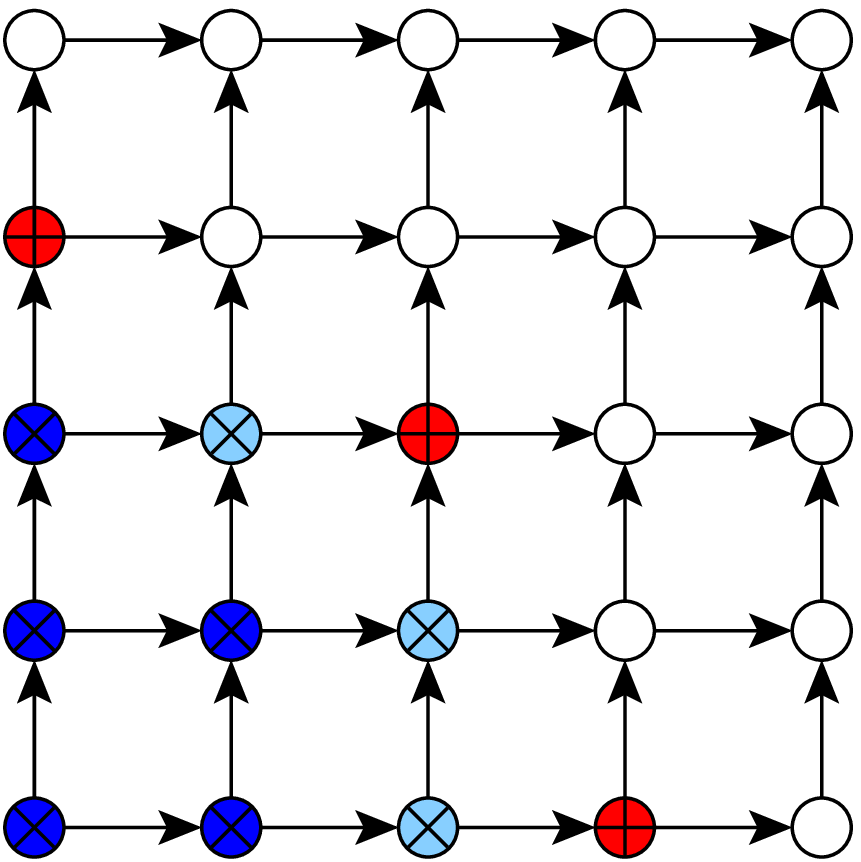}  
 \end{center}

\vspace*{-.6cm}

\caption{Example of active sets for the kernel search algorithms: (left) first phase, when checking necessary conditions, the dark blue nodes ($\times$) are the active kernels (non-zero $\eta$), and the red $+$ are the sources of the complement, which may be added at the next iteration; (right) second phase, when checking sufficient conditions,  the dark blue nodes ($\times$) are the active  kernels (non-zero $\eta$), 
 the light blue nodes ($\times$) are the  kernels with zero weights but are here just to check optimality conditions, and the red nodes ($+$) are the sources of the complement, which may be added at the next iteration.
}
\label{fig:kernelsearchpic}
\end{figure}

The   algorithm presented in \myfig{kernelsearch} will stop either when the duality gap is less than $2\varepsilon$ or when the maximal number of kernels $Q$ has been reached.  That is, our algorithm does not always yield a solution which is  {provably} approximately optimal.
In practice, when the weights $d_v$ increase with the depth of $v$ in the DAG (which we use in simulations), the provably small duality gap generally occurs before we reach a problem larger than $Q$ (however, we cannot make sharp statements). Note that some of the iterations only increase the size of the active sets to check the sufficient condition for optimality. Forgetting those would not change the solution as we add kernels with zero weights; however, in this case, we would not be able to actually certify that we have an $2\varepsilon$-optimal solution (see \myfig{kernelsearchpic} for an example of these two situations). Note that because of potential overfitting issues, settings of the regularization parameter $\lambda$ with solutions having more than $n$ active kernels are likely to have low predictive performance. Therefore, we may expect the algorithm to be useful in practice with moderate values of $Q$.

\paragraph{Running-time complexity.}

Let $D$ be the maximum out-degree (number of children) in the graph, $\kappa$ be the complexity of evaluating the sum in the sufficient condition in \eq{CSopt} (which  usually takes constant time), and $R=|W|$ the number of selected kernels (the number is the size of the active set $W$).
Assuming $O(n^3)$ for the single kernel learning problem, which is conservative~\citep[see, e.g.][for some approaches]{simplesvm1,simplesvm2}, solving all reduced problems has complexity $O(R n^3 )$. Computing all cached matrices has complexity $O(\kappa n^2 \times R D)$ and computing all necessary/sufficient conditions has complexity $O(n^2 \times R^2 D)$.
Thus, the total complexity is $O(R n^3 + \kappa n^2 R D + n^2 R^2 D)$.
Thus, in the case of the directed $p$-grid, we get
$O(R n^3  + n^2 R^2 p)$.
Note that the kernel search algorithm is also an efficient algorithm for 
unstructured MKL, for which we have complexity 
$O(R n^3 +  n^2 R^2 p)$.
Note that gains could be made in terms of scaling with respect to $n$ by using better kernel machine codes with complexity between $O(n^2)$ and $O(n^3)$~\citep{simplesvm1,simplesvm2}.
Note that while the algorithm has polynomial complexity, some work is still needed to make it scalable for more than a few hundreds variables, in particular because of the memory requirements of $O(Rpn^2)$. In order to save storing requirements for the cached kernel matrices, low-rank decompositions might be useful~\citep{fine01efficient,csi}.

\section{Theoretical Analysis in High-Dimensional Settings}

 \label{sec:consistency}

 In this section, we consider the consistency of kernel selection for the norm $\Omega(f)$ defined in \mysec{hkl}. In particular, we show formally in \mysec{allowed} that the active set is always equal to its hull, and provide in \mysec{hull} conditions under which the hull is consistently estimated in low and high-dimensional settings, where the cardinality of $V$ may be large compared to the number of observations.  Throughout this section, we  denote by $\hf$ any minimizer of \eq{milasso2} and  $\hW= \{ v \in V, \hf_v \neq 0\}$ the set of selected kernels.

\subsection{Allowed Patterns}
\label{sec:allowed}
We now show that under certain assumptions any solution of \eq{milasso2} will have a nonzero pattern which is equal to its hull, i.e., the set $\hW= \{ v \in V, \hf_v \neq 0\}$ must be such that $  \hW = \bigcup_{w  \in \hW } \anc(w)$---see~\citet{jenatton} for a more general result with overlapping groups without the DAG structure and potentially low-rank kernels:
\begin{theorem}[Allowed patterns]
Assume that all kernel matrices are invertible. Then the set of zeros $\hW$ of any  solution $\hat{f}$ of \eq{milasso2} is  
 equal to its hull.
\end{theorem}
\begin{proof}
Since the dual problem in \eq{dual-hkl} has a strictly convex objective function on the hyperplane $\alpha^\top 1_n = 0$, the minimum in $\alpha \in \rb^n$ is unique. Moreover, we must have $\alpha \neq 0$ as soon as the loss functions $\varphi_i$
 are not all identical.
Since $\| f_w\|^2 = \zeta_w^2 \alpha^\top \Kc_w \alpha$ for some $\zeta \in Z$, and all $\alpha^\top \Kc_w \alpha > 0$ (by invertibility of $K_w$ and $\alpha^\top 1_n = 0$), we get the desired result, from the sparsity pattern of the vector $\zeta \in \rb^V$, which is always equal to its hull.
\end{proof}
As shown above, the sparsity pattern of the solution of \eq{milasso2} will be equal to its hull, and thus we can only hope to obtain consistency of the hull of the pattern, which we consider in the next sections. In \mysec{hull}, we provide a sufficient condition for optimality, whose weak form tends to be also necessary for consistent estimation of the hull; these results extend the one for the Lasso and the group Lasso~~\citep{Zhaoyu,zou,yuanlin,martin,grouplasso}.

\subsection{Hull Consistency Condition}
\label{sec:hull}
\label{sec:nonasymptotic}
 \label{sec:hull-hilbert}

For simplicity, we consider the square loss for regression and leave out other losses presented in \mysec{loss} for future work. 
Following \citet{grouplasso}, we consider a random design setting where the pairs $(x_i,y_i) \in \X \times \Y$ are sampled from \emph{independent and identical distributions}. We make the following assumptions on the DAG, the weights of the norm and the underlying joint distribution of $(\Phi_v(X))_{v \in V}$ and~$Y $. These assumptions rely on \emph{covariance operators}, which are the tools of choice for analyzing supervised and unsupervised learning techniques with reproducing kernel Hilbert spaces~\citep[see][for a introduction to the main concepts which are used in this paper]{grouplasso,kenji,ktest}. We let denote $\bS$ the joint covariance operator for the kernel $k(x,y)$ defined by blocks corresponding to the decomposition indexed by $V$. We make the following assumptions:
\BIT

\item[\textbf{(A0)}] \emph{Weights of the DAG}: Each of the $\num(V)$ strongly connected components of $V$ has a unique source; the weights of the sources are equal to $d_r \in (0,1]$, while all other weights are equal to $d_v = \beta^{\depth(v)}$ with $\beta >1$. The maximum out-degree (number of children) of the DAG is less than $\deg(V)-1$.

\item[\textbf{(A1)}] \emph{Sparse non-linear model}: $\E(Y|X) = \sum_{w \in \W } \langle
\boldsymbol \f_w (X) + \mathbf{b}$ with $\W\subset V  $,
$\f_w \in \F_w$, $w \in \W$, and $\mathbf{b} \in \rb$; the conditional distribution of $Y|X$ is    Gaussian with variance  $\boldsymbol \sigma^2 >0$. The set $\W$ is equal to its hull, and for each $w \in \W$, $\f_{ \des(w) \cap \W } \neq 0$ (i.e., the hull of the non zero functions is actually $\W$).

\item[\textbf{(A2)}]  \emph{Uniformly bounded inputs}: for all $v \in V$,  $\| \Phi_v(X) \| \leqslant 1$ almost surely, i.e., $k_v(X,X) \leqslant 1$.

\item[\textbf{(A3)}] \emph{Compacity and invertibility of the correlation operator on the relevant variables}: The joint correlation operator $\bC$ of $(\Phi(x_v))_{v \in V}$ (defined with appropriate blocks $\bC_{vw}$) is such that $\bC_{\W \W}$ is compact and invertible (with smallest eigenvalue $\kappa =\lmin(\bC_{\W\W}) > 0$).

\item[\textbf{(A4)}] \emph{Smoothness of predictors}: For each $w \in \W$, there exists $\h_w \in \F_w$ such that $\f_w = \bS_{ww} \h_w$ and $\| \h_w\| \leqslant 1$.

\item[\textbf{(A5)}] \emph{Root-summability of eigenvalues of covariance operators}: For each $w \in \W$, the sum of the square roots of the eigenvalues of $\bS_{ww}$ is less than a constant $\C_1_2$.

\EIT

When the Hilbert spaces all have finite dimensions, covariance operators reduce to covariance matrices, and Assumption \textbf{(A3)} reduces to the invertibility of the correlation matrix $\bC_{\W\W}$ (as it is always compact) and thus of the covariance matrix $\bS_{\W\W}$, while  \textbf{(A4)}  and  \textbf{(A5)}  are always satisfied. These assumptions are discussed by~\citet{grouplasso} in the context of multiple kernel learning, which is essentially our framework with a trivial DAG with no edges (and as many connected components as kernels). Note however that Assumption \textbf{(A4)} is slightly stronger than the one used by~\citet{grouplasso} and that we derive here non asymptotic results, while \citet{grouplasso} was considering only asymptotic results. 

For $K $ a subset of  $ V$, we   denote by $\Omega_K(f_K) = \sum_{v \in K} d_v \| f_{\des(v) \cap K} \|$, the norm reduced to the functions in $K$ and by $\Omega^\ast_{K}$ its \emph{dual norm}~\citep{boyd,Rock70}, defined
as
$\Omega^\ast_{K}(g_K) = \max_{ \Omega_K(f_K) \leqslant 1 } \langle g_K, f_K \rangle$. We consider $ \mathbf{s}_\W\in (\F_v)_{v \in \W}$, defined through
$$
\forall w \in \W,
\ \mathbf{s}_w = \bigg( \sum_{v \in \anc(w) } d_v \| \f_{\des(v) } \|^{-1} \bigg) \h_w.
$$
When the DAG has no edges, i.e., for the regular group Lasso, we get back similar quantities than the ones obtained by~\citet{grouplasso}; if in addition, the feature spaces are all uni-dimensional, we get the vector of signs of the relevant variables, recovering the Lasso conditions~\citep{Zhaoyu,zou,yuanlin,martin}.
The following theorem shows that if the consistency condition in \eq{CScond} is satisfied, then we can upperbound the probability of incorrect hull selection  (see proof in Appendix~\ref{app:sufficient}):
\begin{theorem}[Sufficient condition for hull consistency]
\label{thm:sufficient}
Assume \emph{\textbf{(A0-5)}}  and
\BEQ
\label{eq:CScond}
\Omega^\ast_{\W^c} \left[ \Diag(\bS_{ww}^{1/2})_{ \W^c} \bC_{\W^c \W} \bC_{\W\W}^{-1} \mathbf{s}_\W \right] \leqslant 1- \eta,
\EEQ
with $\eta>0$;
let $\nu = \min_{ w \in \W} \|  \Diag(\bS_{vv})_{\des(w) } \f_{\des(w) }\|$ and
$\omega = \Omega(\f) d_r^{-2}$. Let
 $$\gamma(V) = \frac{4   \log(2 \num(V) ) }{(1-\beta^{-1})^2}  + \frac{4  
 \log \deg(V) }{(\log \beta)^{3}} .$$
Choose $\mu = \lambda \Omega(\f) d_r  \in \left[\frac{2 \boldsymbol\sigma \gamma(V)^{1/2} }{n^{1/2}},  \frac{ c_1 }{\omega^{11/2} |\W |^{7/2}} \right] $.
The probability of incorrect hull selection is upper-bounded by:
\BEQ
\label{eq:bound}
\exp\Big( \! - \frac{ \mu^2 n }{8 \boldsymbol \sigma^2  }
\Big) +
    \exp \Big( - c_2
 \frac{ \mu  n    }{\omega^{3} |\W|^{3}  }
\Big)+
 \exp \Big(  - c_3 \frac{ \mu^{3/2} n }{\boldsymbol\sigma^2 \omega^{7}  |\W|^4 }
\Big),   \EEQ
where $c_1$, $c_2$, $c_3$ are     positive monomials in $\kappa$, $\nu$, $\eta$ and $C_{1/2}^{-1}$.
\end{theorem}

The previous theorem is the main theoretical contribution of this paper. It is a non-asymptotic result which we comment on in the next paragraphs. The proof relies on novel concentration inequalities for empirical covariance operators and for structured norms, which may be useful in other settings (see results in Appendices~\ref{app:hoeff},~\ref{app:cova} and~\ref{app:ls}). Note that the last theorem is  not  a consequence of   similar results for flat multiple kernel learning or group Lasso~\citep{grouplasso,nardy,multi-task}, because the groups that we consider are overlapping.
 Moreover,  the last theorem shows that we can indeed estimate the correct hull of the sparsity pattern if the sufficient condition is satisfied. In particular, if we can make the groups such that the between-group correlation is as small as possible, we can ensure correct hull selection.

\paragraph{Low-dimensional settings.} When the DAG is assumed fixed (or in fact only the number of connected components $\num(V)$ and the maximum out-degree $\deg(V)$) and $n$ tends to $+\infty$, the probability of incorrect hull selection tends to zero as soon as $\lambda n^{1/2}$ tends to $+\infty$ and
$\lambda $ tends to zero, and the convergence is exponentially fast in  $\lambda n$.

\paragraph{High-dimensional settings.} When the DAG is large compared to $n$, then, the previous theorem leads to a consistent estimation of the hull, if the interval defining $\mu$ is not empty, i.e., 
$ n \geqslant 4 \boldsymbol \sigma^2 \gamma(V) \omega^{11} |\W|^7 c_1^{-2} $. Since $\gamma(V) 
  = O(  \log( \num(V) ) + \log( \deg(V) ) )$, this implies that we may have correct hull selection in situations where $n = O(  \log( \num(V) ) + \log( \deg(V)  )  )$. We may thus have an exponential number of connected components and an exponential out-degree, with no constraints on the maximum depth of the DAG (it could thus be infinite).
  
  Here, similar scalings could be obtained with a weighted $\ell_1$-norm (with the same weights $\beta^{\depth(v)}$;  however, such a weighted Lasso might select kernels which are far from the roor and would not be amenable to an efficient active set algorithm.
  
  \paragraph{Multiple kernel learning (group Lasso).} In this situation, we have a DAG with $p$ connected components (one for each kernel), and zero out-degree (i.e., $\deg(V) \!=\!  1$), leading to $
  \gamma(V) = O( (\log p)^{1/2})$, a classical non-asymptotic result in the unstructured settings for finite-dimensional groups~\citep{nardy,martin,multi-task}, but novel for the multiple kernel learning framework, where groups are infinite-dimensional Hilbert spaces. Note that the proof techniques would be much simpler and the result sharper in terms of power of $|\W|$ and~$\omega$ with finite-dimensional groups and with the assumption of invertibility of $\bS_{\W\W}$ and/or fixed design assumptions. Finally, Theorem~\ref{thm:sufficient} also applies
  for a modified version of the elastic net~\citep{zou2005regularization}, where the $\ell_2$-norm is added to the sum of block $\ell_1$ norm---by considering a single node with the null kernel connected to all other kernels.

\paragraph{Non linear variable selection.} For the power set and the directed grids that we consider for non-linear variable selection in \mysec{kernels}, we have $\num(V) =1$ and $\deg(V)=p$ where $p$ is the number of variables, and thus $\gamma(V) = O(
\log p   ) = O (  \log \log |V|   )$, i.e., we may have exponentially many variables to choose non-linearly from, or a \emph{doubly} exponential number of kernels to select from.

\paragraph{Trade-off for weight $\beta$.} Intuitively, since the weight on the norm $\| f_{\des(v)}\|$ is equal to $\beta^{\depth(v)}$, the greater the $\beta$ the stronger the prior towards selecting nodes close to the sources. However, if $\beta$ is too large, the prior might be too strong to allow selecting nodes away from the sources. 

This can be illustrated in the bound provided in Theorem~\ref{thm:sufficient}. The constant $\gamma(V)$ is a decreasing function of $\beta$, and thus having a large $\beta$, i.e., a large penalty on the deep vertices, we decrease the lower bound of allowed regularization parameters $\mu$ and thus increase the probability of correct hull selection (far away vertices are more likely to be left out). However, since $\Omega(\f)$ is a rapidly increasing function of $\beta$, the upper bound decreases, i.e., if we penalize too much, we would start losing some of the deeper relevant kernels.
 Finally, it is worth noting that if the constant $\beta$ tend to infinity slowly with $n$, then we could always consistently estimate the depth of the hull, i.e., the optimal interaction complexity.
Detailed results are the subject of ongoing work.

\paragraph{Results on estimation accuracy and predictive performance.} In this paper, we have focused on the simpler results of  hull selection consistency, which allow simple assumptions. It is however of clear interest of following the Lasso work on estimation accuracy and predictive performance~\citep{tsyb} and extend it to our structured setting. In particular, the rates of convergence should also depend on the cardinal of the active set $|\W|$ and not on the cardinality of the DAG~$|V|$.

\paragraph{Enhancing consistency condition.}
The sufficient condition in \eq{CScond} states that low correlation between relevant and irrelevant feature spaces leads to good model selection. As opposed to unstructured situations, such low correlation may be enhanced with proper hierarchical whitening of the data, i.e., for all $v \in V$, we may project $(\Phi_v(x_i))_{i=1,\dots,n}$ to the orthogonal of all ancestor vectors 
$(\Phi_w(x_i))_{i=1,\dots,n}$, $w \in \anc(v)$. This does not change the representation power of our method but simply enhances its statistical consistency. 

Moreover, Assumption (\textbf{A3}) is usually met for all the kernel decompositions presented in \mysec{kernels}, except the all-subset Gaussian kernel (because each feature space of each node contains the feature spaces associated with its parents). However, by the whitening procedure outlined above, similar results than Theorem~\ref{thm:sufficient} might be obtained.
Besides, if the \emph{original variables} used to define the kernel decompositions presented in \mysec{kernels} are independent, then the consistency condition in \eq{CScond} is always met  except  for the all-subset Gaussian kernel; again, a pre-whitening procedure might solve the problem in this case.

\paragraph{Necessary consistency condition.}

We also have a necessary condition which is a weak form of the sufficient condition in \eq{CScond}---the proof follows closely the one for the unstructured case from~\citet{grouplasso}:
\begin{proposition}[Necessary condition for hull consistency]
\label{prop:necessary}
Assume \emph{\textbf{(A1-3)}} and $V$ is fixed, with $n$ tending to $+\infty$. If there is a sequence of regularization parameters $\lambda$ such that both the prediction function  and the hull of the active kernels is consistently estimated, then we have  
\BEQ
\label{eq:CNcond}
\Omega^\ast_{\W^c}\big[ \Diag(\bS_{ww}^{1/2})_{ \W^c} \bC_{\W^c \W} \bC_{\W\W}^{-1} \mathbf{s}_\W \big]\leqslant 1.
\EEQ
 \end{proposition}
 
 The conditions in \eq{CScond} and \eq{CNcond} make use of the dual norm, but we can loosen them using lower and upper bounds on these dual norms: some are computable in polynomial time, like the ones used for the active set algorithm presented in \mysec{algorithm} and more detailed in Appendix~\ref{app:dualnorms}. However, we can obtain simpler bounds which require to look over the entire DAG; we obtain by lowerbounding $\|f_{\des(v)}\| $ by $\|f_v\|$ and upperbounding
 it by $\sum_{w \in \des(v)} \| f_w\|$ in the definition of $\Omega(f)$, for $g \in \F$:
$$
 \max_{w \in \W^c} \frac{ \| g_w \| } {\sum_{ v \in \anc(w) \cap \W^c} d_v}
 \leqslant  \Omega^\ast_{\W^c}( g_{\W^c} )   \leqslant   \max_{w \in \W^c} \frac{ \| g_w \| } {d_w} .
$$
 The lower and upper bounds are equal when the DAG is trivial (no edges), and we get back the usual weighted $\ell_\infty$-$\ell_2$ norm
 $ \max_{w \in \W^c} \frac{ \| g_w \| } {d_w}$.

\subsection{Universal Consistency}
\label{sec:universal}
In this section, we briefly discuss the universal consistency properties of our method when used for non-linear variable selection: do the kernel decompositions presented in \mysec{kernels} allow the estimation of arbitrary functions? The main rationale behind using all subsets of variables rather than only singletons is that most non-linear functions may not be expressed as a sum of functions which depend only on one variable---what regular MKL~\citep{skm} and SPAM~\citep{spam} would use. All subsets are thus required to allow universal consistency, i.e., to be able to approach any possible predictor function.

Our norm $\Omega(\f)$ is equivalent to a weighted Hilbertian norm, i.e.:
 $$
\sum_{v \in V} d_v \| f_v\|^2 \leqslant \Omega(f)^2    \leqslant    |V| \sum_{w \in V }  \bigg(\sum_{v \in \anc(w)} d_v \bigg) \| f_w \|^2.
$$
Therefore, the usual RKHS balls associated to the universal kernels we present in \mysec{kernels} are contained in the ball of our norms, hence we obtain universal consistency~\citep{universal2, universal1} in low-dimensional settings when $p$ is small. A more detailed and refined analysis that takes into account the sparsity of the decomposition and convergence rates is out of the scope of this paper, in particular for the different regimes for $p$, $q$ and $n$.

   \section{Simulations}
   \label{sec:simulations}
   \label{sec:experiments}
           
     In this section, we report simulation experiments on synthetic datasets and datasets from the UCI repository. Our goals here are (a) to compare various kernel-based approaches to least-squares regression from the \emph{same} kernel, (b) to compare the various kernel decompositions presented in \mysec{kernels} within our HKL framework, and (c) to compare predictive performance with non-kernel-based methods---more simulations may be found in earlier work~\citep{hkl}.

  \subsection{Compared Methods}
     
  In this section, we consider various nonparametric methods for non-linear predictions. Some are based on the kernel decompositions defined in \mysec{kernels}. Non-kernel based methods were chosen among methods with some form of variable selection capabilities. All these methods were used with two loops of 10-fold cross-validation to select regularization parameters and hyperparameters (in particular $\beta$). All results are averaged over 10 replications (medians, upper and lower quartiles are reported).
  
  \paragraph{Hierarchical kernel learning (HKL).} We use the algorithm presented in \mysec{algorithm} with the kernel decompositions presented in \mysec{kernels}, i.e., Hermite polynomials (``Hermite''), spline kernels (``spline'') and all-subset Gaussian kernels (``Gaussian'').
  
  \paragraph{Multiple kernel learning (MKL).} We use the algorithm presented in \mysec{algorithm} with the kernel decompositions presented in \mysec{kernels}, but limited to kernels of depth one, which corresponds to sparse generalized additive models.

  \paragraph{Constrained forward selection (greedy).} Given a kernel decomposition with rank one kernels, we consider a forward selection approach that satisfies the same constraint that we impose in our convex framework.
    
  \paragraph{Single kernel learning ($L_2$).} When using the full decomposition (which is equivalent to summing all kernels or penalizing by an $\ell_2$-norm) we can use regular single kernel learning.
  
  \paragraph{Generalized Lasso (Glasso).} Given the same kernel matrix as in the previous method, \citet{roth} considers predictors of the form $\sum_{i=1}^n \alpha_i k_i(x,x_i)$, with the regularization by the $\ell_1$-norm of $\alpha$ instead of $\alpha^\top K \alpha$ for the regular single kernel learning problem.
     
     \paragraph{Multivariate additive splines (MARS).} This method of~\citet{mars} is the closest in spirit to the one presented in this paper: it builds in a forward greedy way multivariate piecewise polynomial expansions. Note however, that in MARS, a node is added only after one of its parents (and not all, like in HKL).
      We use the R package with standard hyperparameter settings.
     
     \paragraph{Regression trees (CART).} We consider regular decision trees for regression using the standard R implementation~\citep{cart84} with standard hyperparameter settings.
     
     \paragraph{Boosted regression trees (boosting).} We use the R ``gbm'' package which implements the method of~\citet{friedman}.
     
     \paragraph{Gaussian processes with automatic relevance determinations (GP-ARD).} We use the code of~\citet{GP}, which learns widths for each variable within a Gaussian kernel, using a Bayesian model selection criterion (i.e., without using cross-validation). Note that HKL, with the all-subset Gaussian decomposition, does not search explictly for $A$ in the kernel $\exp( - ( x-x')^\top A ( x- x'))$, but instead considers a large set of particular values of $A$ and finds a linear combination of the corresponding kernel.

   \begin{figure}

   \vspace*{-.5cm}
   
\hspace*{-.45cm}   \includegraphics[scale=.475]{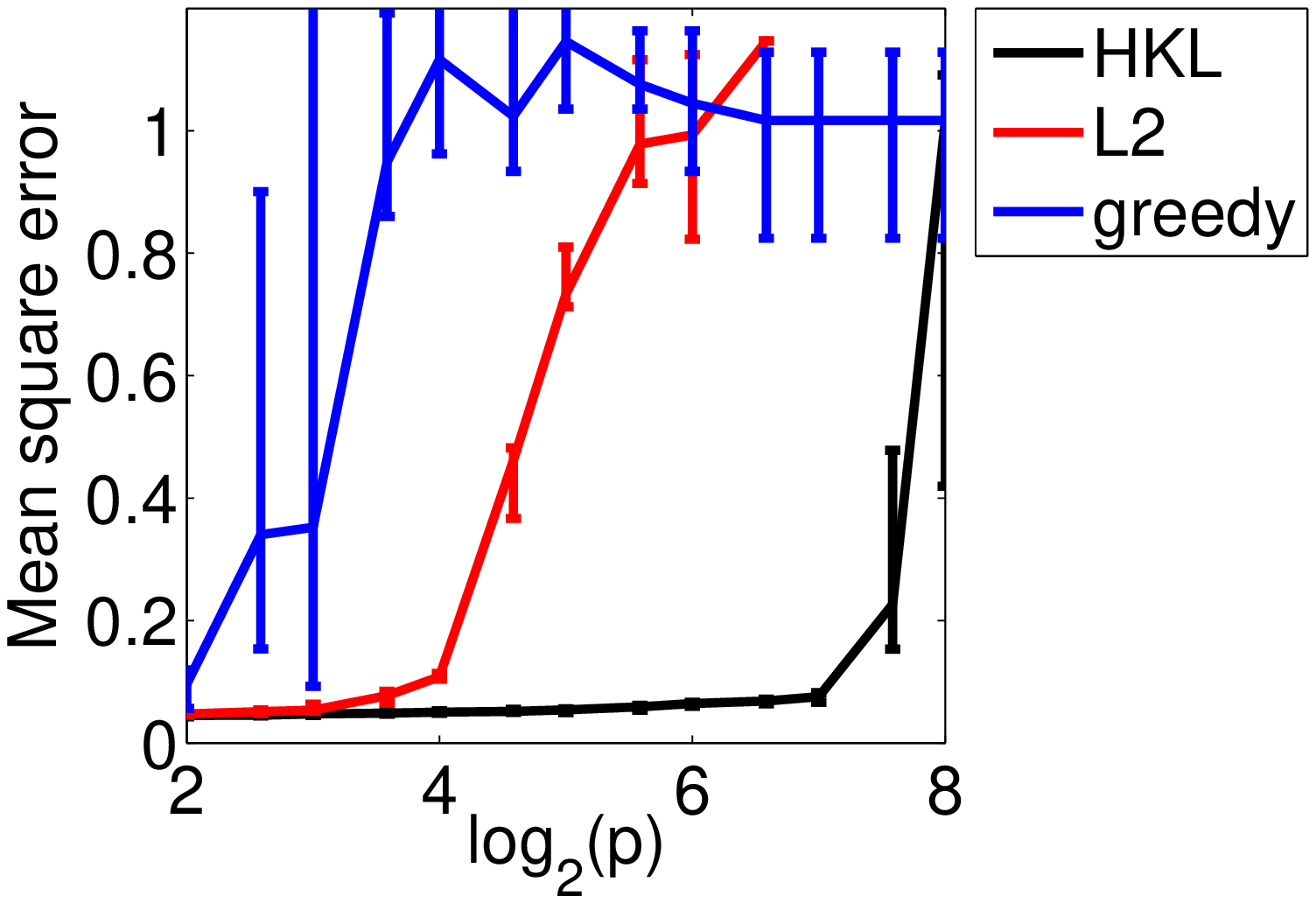}
\hspace*{.1cm}
   \includegraphics[scale=.475]{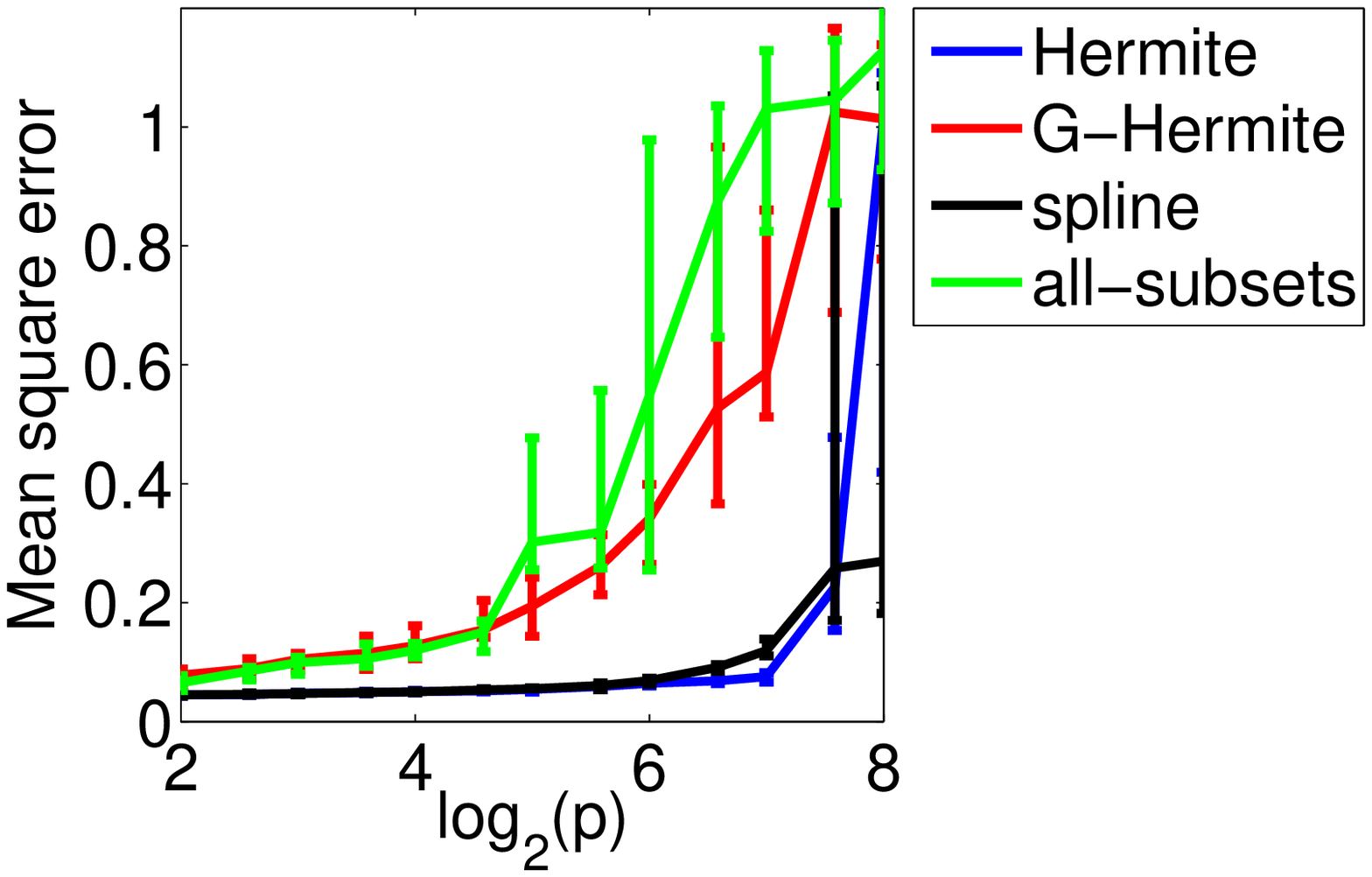}
\hspace*{-.25cm}   \includegraphics[scale=.475]{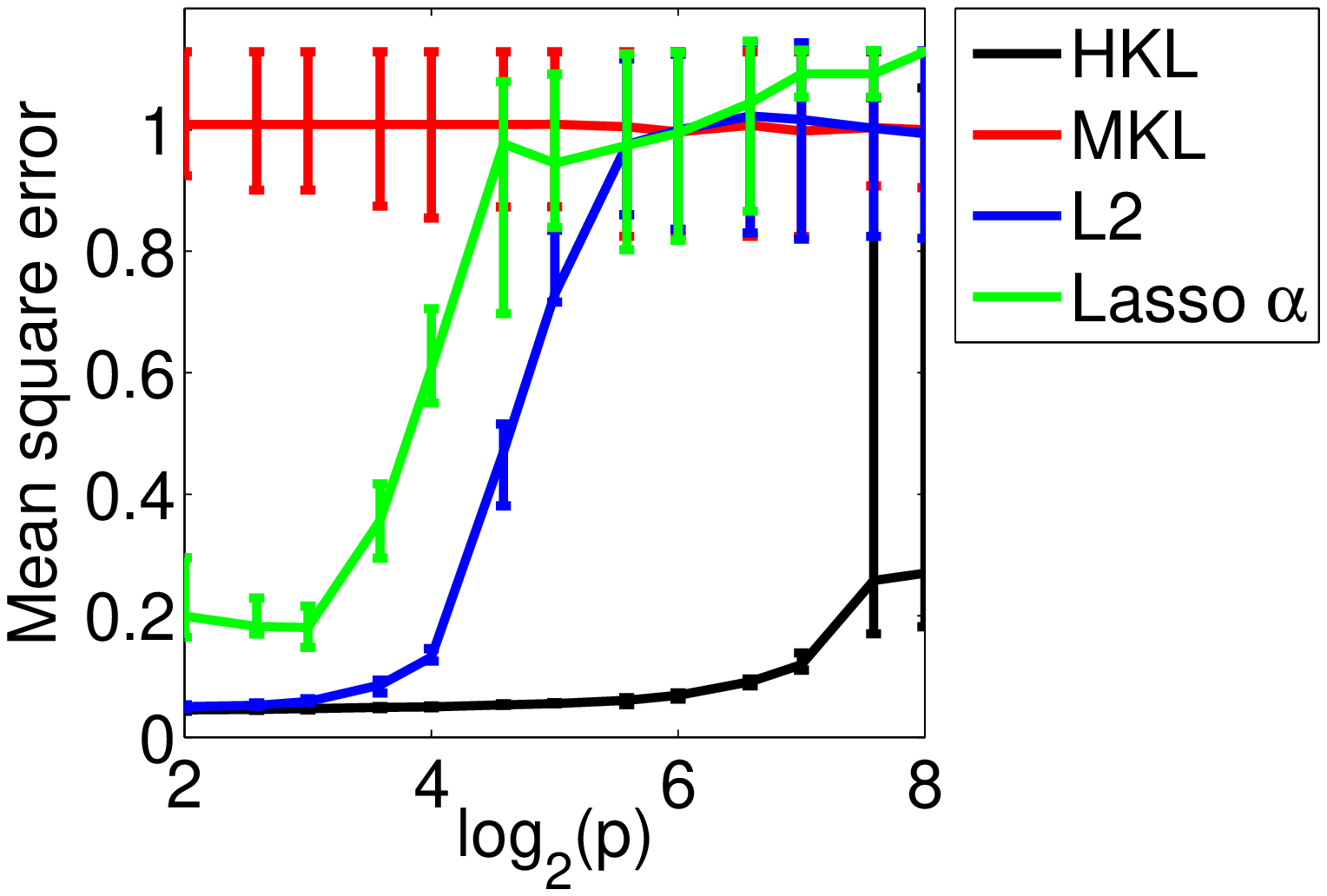}
\hspace*{.2cm}
   \includegraphics[scale=.475]{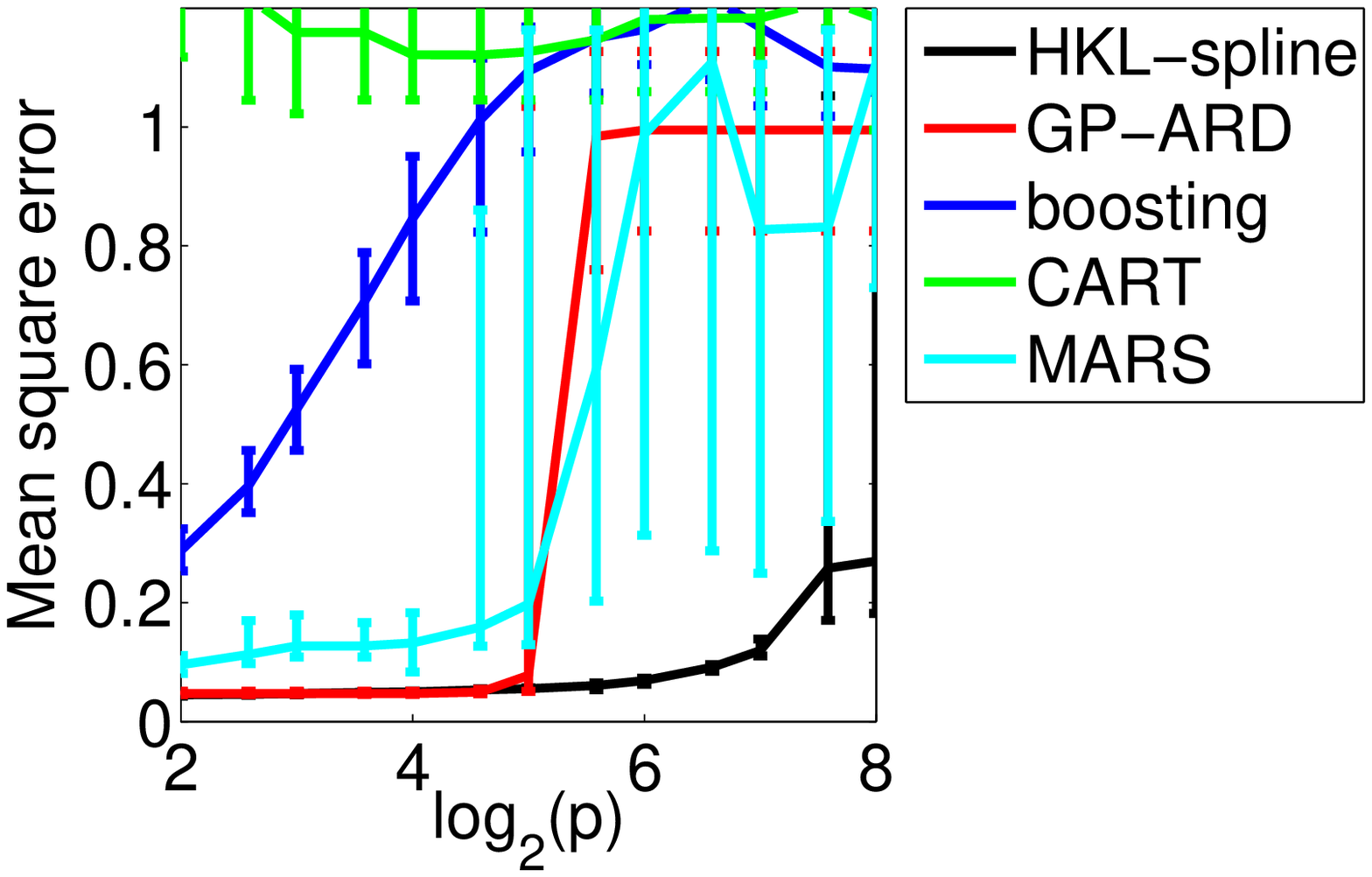}
   
   \vspace*{-.35cm}
   
   \caption{Comparison of non-linear regression methods (mean squared error vs. dimension of problem (in $\log$ scale). (Top left) comparison of greeedy, $\ell_2$ and $\ell_1$ (HKL) methods on the same Hermite kernel decomposition. (Top right) comparison of several kernel decompositions for HKL. (Bottom left) comparison with other kernel-based methods. 
   (Bottom right) comparison with   non-kernel-based methods.}  
   \label{fig:synthetic}
    \end{figure}

\subsection{Synthetic Examples}
 We generated synthetic data as follows: we generate a covariance matrix from a Wishart distribution of dimension $p$ and with $2p$ degrees of freedom. It is then normalized to unit diagonal and $n$ datapoints are then sampled i.i.d.~from a Gaussian distribution with zero  mean  and this covariance matrix. We then consider the non-linear function $f(X) = \sum_{i = 1}^ r \sum_{j=i+1}^r X_j X_i$, which takes all cross products of the first $r$ variables. The output $Y$ is then equal to $f(X)$ plus some Gaussian noise with known signal-to-noise ratio.

 Results are reported in \myfig{synthetic}. On the top left plot, we compare different strategies for linear regression, showing that in this constrained scenario where the generating model is sparse, $\ell_1$-regularization based methods outperform other methods (forward selection and ridge regression). On the top right plot, we compare different kernel decompositions: as should be expected, the Hermite and spline decompositions (which contains exactly the generating polynomial) performs best. On the bottom left plot, we compare several kernel-based methods on the same spline kernel, showing that when sparsity is expected, using sparse methods is indeed advantageous. Finally, on the bottom right plot, we compare to non-kernel based methods, showing that ours is more robust to increasing input dimensions $p$. It is also worth noting the instabilities of the greedy methods such as MARS or ``greedy'', which sometimes makes wrong choices at the start of the procedure, leading to low performance.

       \begin{figure}
\hspace*{-.45cm}   \includegraphics[scale=.45]{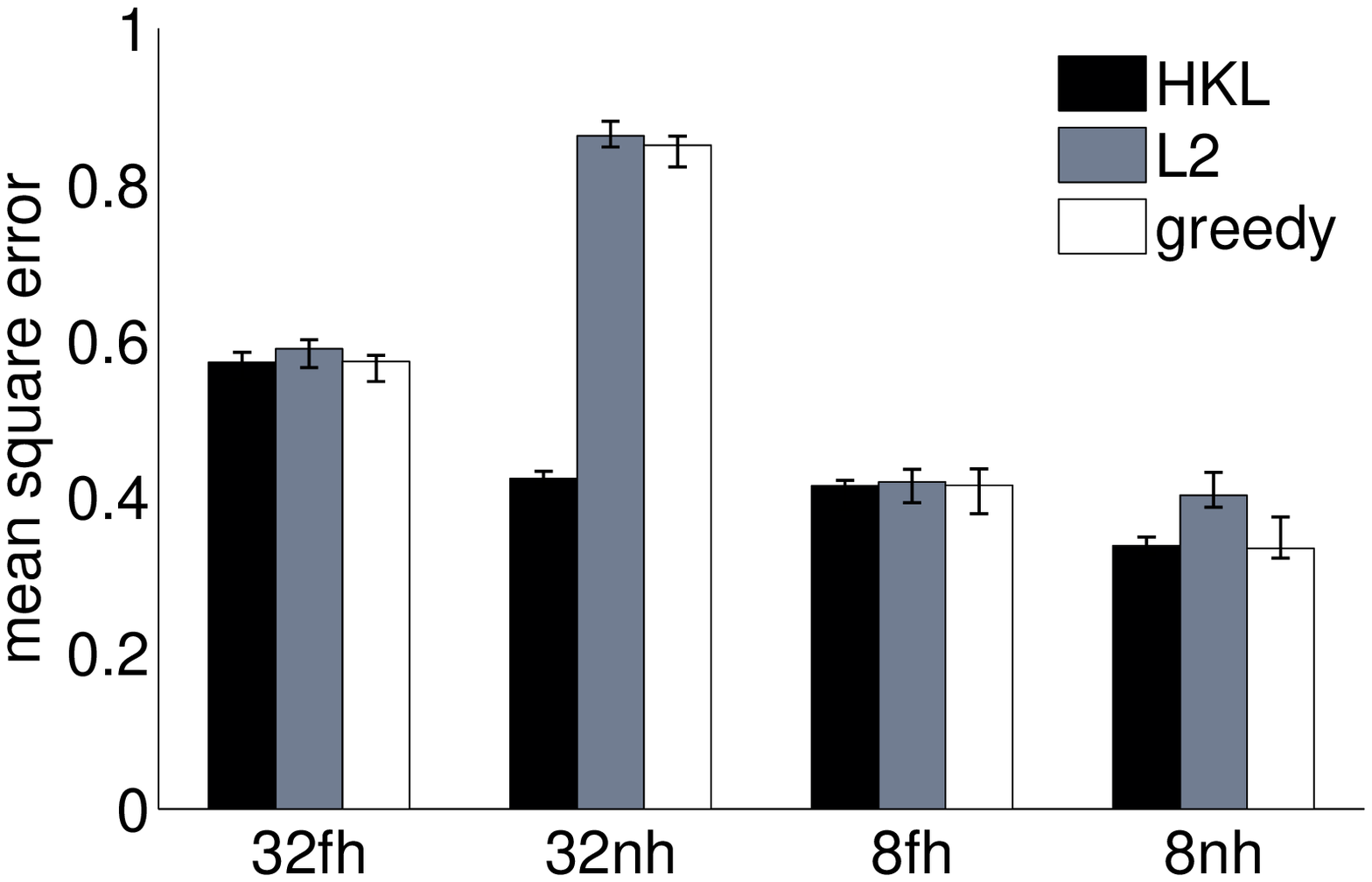}
\hspace*{.3cm}
   \includegraphics[scale=.45]{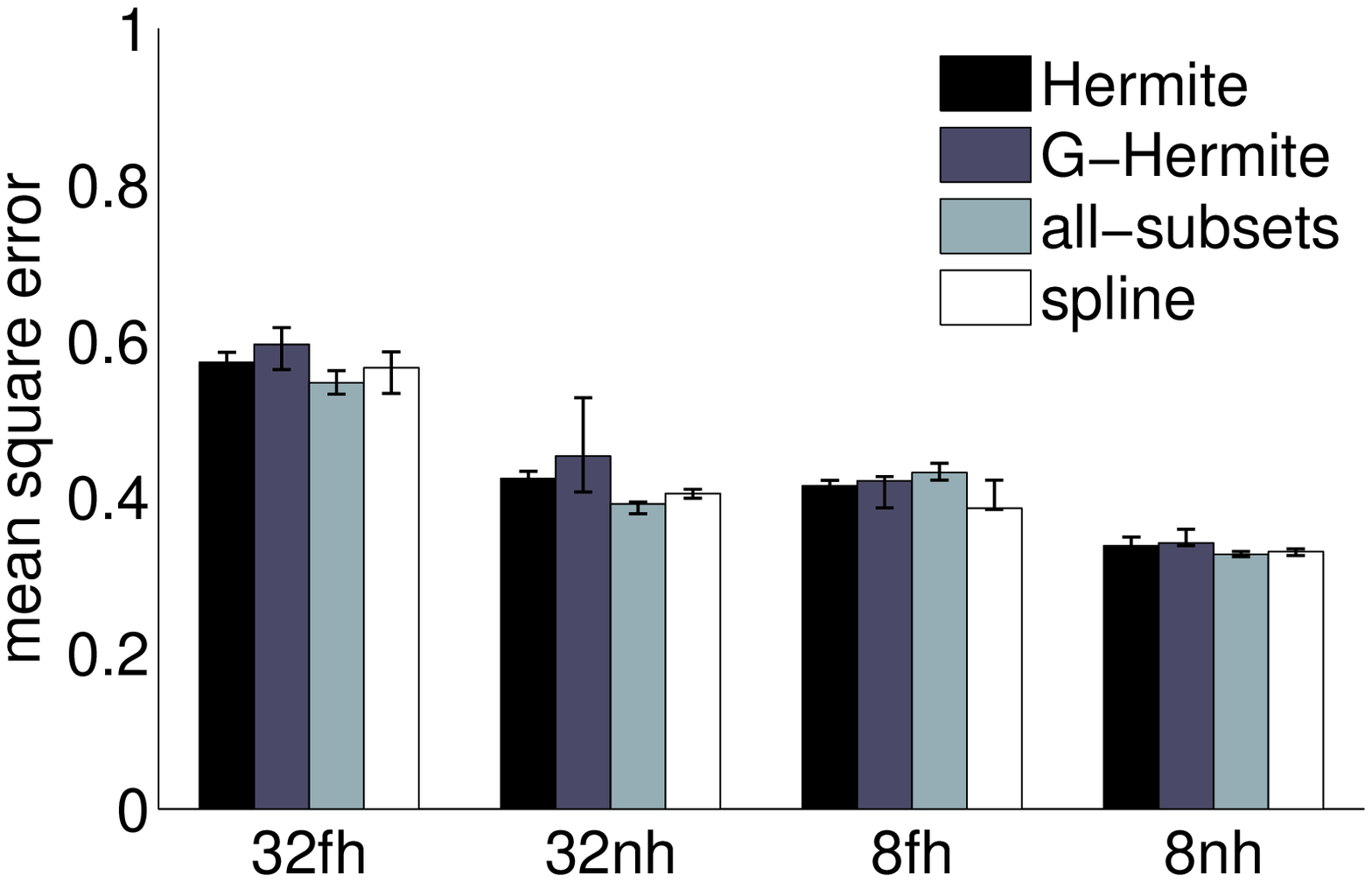}
\hspace*{-.25cm}   \includegraphics[scale=.45]{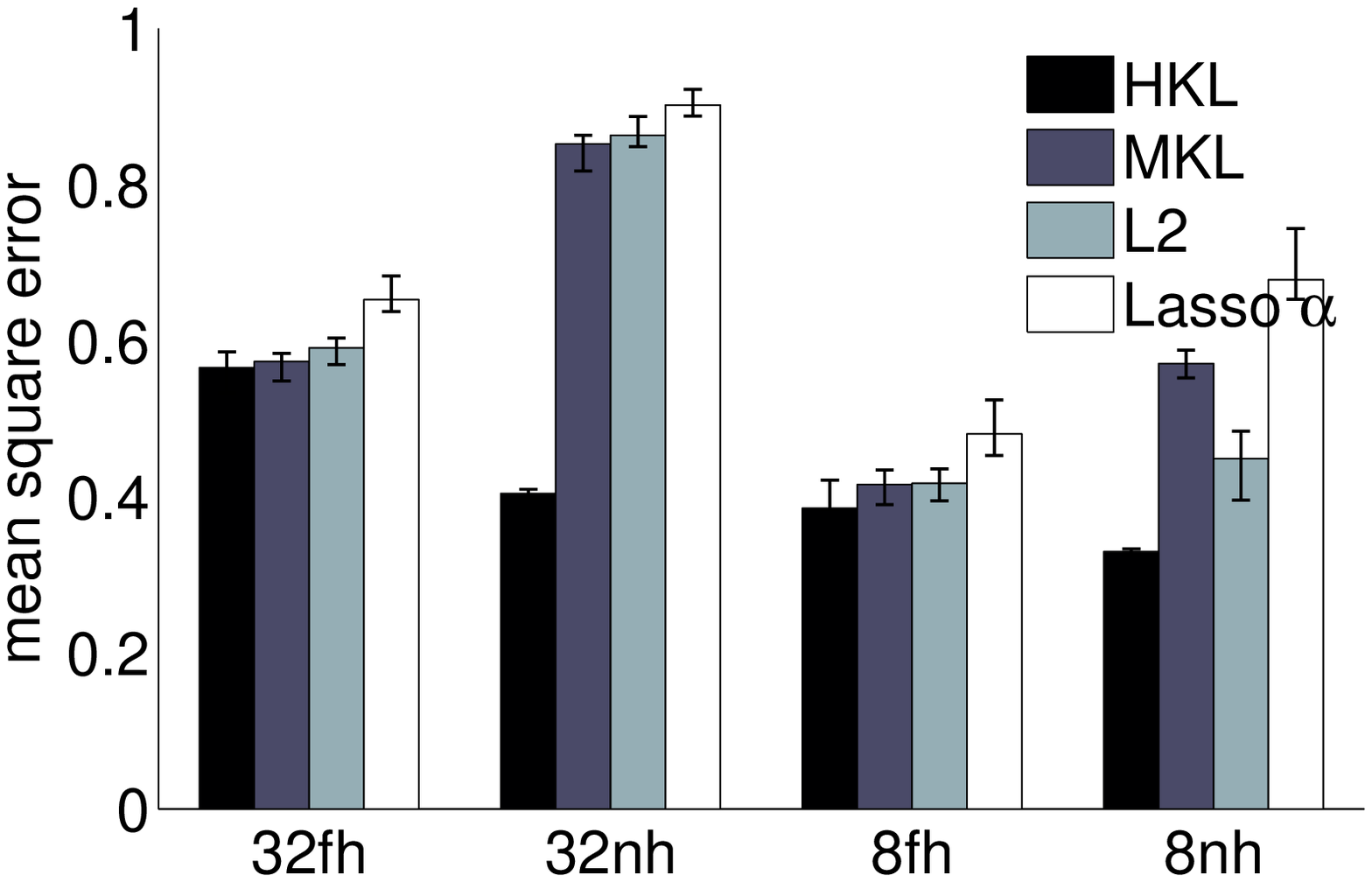}
\hspace*{.55cm}
   \includegraphics[scale=.45]{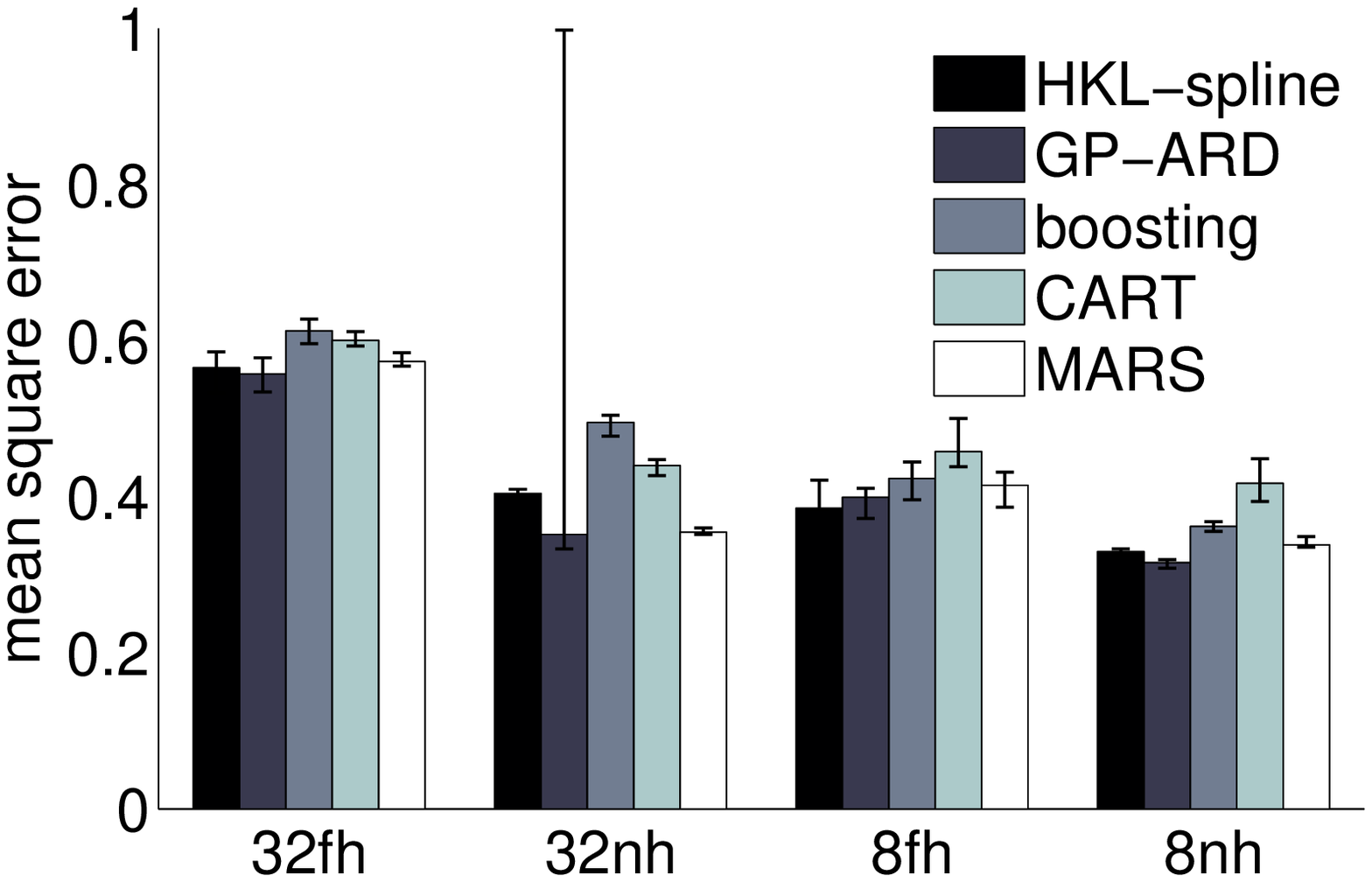}
   
   \vspace*{-.25cm}
   
   \caption{Comparison of non-linear regression methods (mean squared error vs. dimension of problem (in $\log$ scale). (Top left) comparison of greeedy, $\ell_2$ and $\ell_1$ (HKL) methods on the same Hermite kernel decomposition. (Top right) comparison of several kernel decompositions for HKL. (Bottom left) comparison with other kernel-based methods. 
   (Bottom right) comparison with other non-kernel-based methods.}  
   \label{fig:pumadyn}
    \end{figure}

  \subsection{UCI Datasets} 
 We perform simulations on the ``pumadyn'' datasets from the UCI repository~\citep{uci}. These datasets are obtained from realistic simulations of the dynamics of a robot arm, and have different strengths of non-linearities  (fh: fairly linear, high noise; nh: non-linear, high noise) and two numbers (8 and 32) of input variables. 
 
 Results are reported in \myfig{pumadyn}.
On the top left plot, we compare different strategies for linear regression with $n=1024$ observations: with moderately non-linear problems (32fh, 8fh), all performances are similar, while for non-linear problems (32nh, 8nh), HKL outperforms  other methods (forward selection and ridge regression). On the top right plot, we compare different kernel decompositions: here, no decomposition includes the generating model, and therefore, none clearly outperforms the other ones.
 On the bottom left plot, we compare several kernel-based methods on the same spline kernel: it is interesting to note that for moderately linear problems, MKL performs well as expected, but not anymore for highly non-linear problems. 
 
 Finally, on the bottom right plot, we compare to non-kernel based methods: while boosting methods and CART are clearly performing worse, HKL, MARS and Gaussian processes perform better, with a significant advantage to MARS and Gaussian processes for the dataset ``32nh''. There are several explanations regarding the worse performance of HKL that could lead to interesting developments for improved performance: first, HKL relies on estimating a regularization parameter by cross-validation, while both MARS and GP-ARD rely on automatic model selection through frequentist or Bayesian procedures, and it is thus of clear interest to consider methods to automatically tune the regularization parameter for sparse methods such as HKL. Moreover, the problem is not really high-dimensional as $n$ is much larger than $p$, and our regularized method has a certain amount of bias that the other methods don't have; this is a classical problem of $\ell_1$-regularized problems, and this could be fixed by non-regularized estimation on the selected variables.

\section{Conclusion}

  We have shown how to perform hierarchical multiple kernel learning (HKL) in polynomial time in the number of selected kernels. This framework may be applied to many positive definite kernels and  we have focused on kernel decompositions
  for non-linear variable selection: in this setting, we can both select which variables should enter and the corresponding degrees of interaction complexity. We have proposed an active set algorithm as well a theoretical analysis that suggests that we can still perform \emph{non-linear} variable selection from a number of variables which is exponential in the number of observations.
  
 Our framework can be extended in multiple ways:
  first,  this paper shows that trying to use $\ell_1$-type penalties may be advantageous inside the feature space. That is, one may take the opposite directions than usual kernel-based methods and look inside the feature spaces with sparsity-inducing norms instead of building feature spaces of ever increasing dimensions.
  We are currently investigating applications to other kernels, such as the pyramid match kernels~\citep{grauman,cuturi}, string kernels, and graph kernels~\citep[see, e.g.,][]{Cristianini2004}.  Moreover, theoretical and algorithmic connections with the recent work of~\citet{structuredsparsity} on general structured sparsity and greedy methods
   could be made.

Moreover, we have considered in this paper a specific instance of block $\ell_1$-norms with overlapping groups, i.e., groups organized in a hierarchy, but some of the techniques and frameworks presented here can be extended to more general overlapping structures~\citep{jenatton}, for DAGs or more general graphs; it would also be interesting to consider non discrete hierarchical structures with a partial order, such as positive definite matrices.

Finally, we hope to make connections with other uses of sparsity-inducing norms, in particular in signal processing, for compressed sensing~\citep{cs1,cs2}, dictionary learning~\citep{field} and sparse principal component analysis~\citep{dspca}.

\appendix

\section{Proofs of Optimization Results}
\label{app:optimization}
In this first appendix, we give proofs of all results related to the optimization problems. 

\subsection{Set of Weights for Trees}
\label{app:tree}
We   prove that the set of weights $\zeta$, i.e., $Z$, is itself convex when the DAG is a tree. We conjecture that the converse is true as well.

\begin{proposition}
\label{prop:tree}
If  $V$ is a tree, the
 set $Z = \{ \zeta(\eta) \in \rb^V,  \eta \in \rb_+^V,  \sum_{v \in V} d_v^2 \eta_v \leqslant 1  \}$ is convex.
\end{proposition}
\begin{proof}
When the DAG is a tree (i.e., when each vertex has at most one parent and there is a single source $r$), then, we have for all~$v$ which is not the source of the DAG (i.e., for which there is exactly one parent),
$\zeta_{\pi(v)}^{-1} - \zeta_v^{-1} = - \eta_{v}^{-1}$. This implies that the constraint $\eta\geqslant 0$ is equivalent to
$\zeta_v \geqslant 0$ for all leaves $v$,  and for all $v$ which is not a source, $\zeta_{\pi(v)} \geqslant  \zeta_v$, with equality possible only when they are both equal to zero.

Moreover, for the source  $r$, $\zeta_r = \eta_r$. 
The final constraint $\sum_{v \in V} \eta_v d_v^2 \leqslant 1$, may then be written as
$
\sum_{v   \neq r} d_v^2 \frac{1}{ \zeta_v^{-1} - \zeta_{\pi(v)}^{-1}}  +  \zeta_r d_r^2 \leqslant 1$, 
that is,
$
\sum_{ v   \neq r} d_v^2 \left( \zeta_v + \frac{ \zeta_v^2 }{ \zeta_{\pi(v)} -\zeta_v  }  \right) +  \zeta_r d_r^2 \leqslant 1,$
which is   a convex constraint~\citep{boyd}.
 \end{proof}
 
 \subsection{Proof of Proposition~\ref{prop:single}}
 \label{app:single}
We introduce auxiliary variables $u_i =   \langle f  , \Phi(x_i) \rangle + b$ and consider the Lagrangian:
$$\mathcal{L}(u,f,b,\alpha) = \frac{1}{n} \sum_{i=1}^n \varphi_i(u_i)
+\frac{\lambda}{2}  \|f \|^2  + \lambda \sum_{i=1}^n
 \alpha_i  ( u_i -  \langle f  , \Phi(x_i) \rangle - b).$$
Minimizing with respect to the primal variable $u$ leads to the term 
$- \frac{1}{n} \sum_{i=1}^n \psi_i(-n\lambda \alpha_i) $; minimizing with respect to $f$ leads to the term  $- \frac{\lambda}{2}  \alpha^\top K \alpha$ and to the expression of $f$ as a function of $\alpha$, and minimizing with respect to $b$ leads to the constraint $1_n^\top \alpha = \sum_{i=1}^n \alpha_i =0$.

\subsection{Preliminary Propositions}

We will   use the following simple result, which implies that each component $\zeta_w(\eta)$ is a concave function of   $\eta$ (as the minimum of linear functions of $\eta$):
\begin{lemma}
\label{lemma:1}
Let $a \in (\rb_+^\ast)^m$.
The minimum of $\sum_{j=1}^m a_j x_j^2$ subject to $x \geqslant 0$ and $\sum_{j=1}^m x_j=1$ is equal to
$\left( \sum_{j=1}^m a_i^{-1} \right)^{-1}$ and is attained at
$x_i = a_i^{-1}  \left( \sum_{j=1}^m a_i^{-1} \right)^{-1}$.
\end{lemma}
\begin{proof}
The result is a consequence of applying Cauchy-Schwartz inequality, applied to vectors with components $x_j a_j^{1/2}$ and $a_j^{-1/2}$.
 Note that when some of the $a_j$ are equal to zero, then the minimum is zero, with optimal $x_j$ being zero whenever $a_j\neq 0$.
\end{proof}
 
 The following proposition derives the dual of the problem in $\eta$, i.e., the dual of \eq{etat}:

 \begin{proposition} 
 \label{prop:kappa}
 Let $L = \{ \kappa \in \rb_+^{V \times V}, \ \kappa_{ \anc(w)^c w} = 0
 \mbox{ and } \forall w \in V, \ \sum_{v \in \anc(w)} \kappa_{vw}=1\}$. The following convex optimization problems are dual to each other, and there is no duality gap : 
\BEQ 
\label{eq:kappaPB}\displaystyle
\min_{\kappa \in L} \bigg\{
\max_{v \in V} d_v^{-2} \sum_{w \in \des(v)}\kappa_{vw}^2 \alpha^\top \Kc_w
\alpha \bigg\},
\EEQ
\BEQ
\max_{\eta \in H} \sum_{w \in V}
\zeta_w(\eta) \alpha^\top  \Kc_w \alpha
 .\EEQ
 \end{proposition}
 \begin{proof}
We have the Lagrangian
 $\mathcal{L}(A, \kappa,\eta)
 = A + \sum_{v \in V} \eta_v \left( 
 \sum_{w \in \des(v)} \kappa_{vw}^2 \alpha^\top \Kc_w \alpha  - A d_v^2\right)$, with $\eta\geqslant 0$,
 which,  using Lemma~\ref{lemma:1}, can be minimized in closed form with respect to $A$, to obtain the constraints
 $\sum_{v \in V} \eta_v d_v^2 = 1$  and with respect to  $\kappa \in L$. We thus get
 \BEAS
 \min_{\kappa \in L}
 \max_{v \in V}
d_v^{-2}  \sum_{w \in \des(v) } \kappa_{vw}^2 \alpha^\top \Kc_w \alpha  & =  &\max_{\eta} \alpha^\top  \bigg(
\sum_{w \in V} \textstyle  \left( \sum_{v \in \anc(w)} \eta_v^{-1} \right)^{-1} \Kc_w \bigg) \alpha, \\
& = &  \max_{\eta} \alpha^\top  \bigg(
\sum_{w \in V}  \zeta_w(\eta) \Kc_w \bigg) \alpha.
\EEAS
Given $\eta$, the optimal value for $\kappa$ has a specific structure (using Lemma~\ref{lemma:1}, for all $w \in V$):
(a)
if for all $v \in \anc(w)$, $\eta_v > 0 $, then $\kappa_{vw} = \zeta_w \eta_v^{-1}$ for all $v \in \anc(w)$,
(b) if there exists $v \in \anc(w)$ such that  $\eta_v = 0 $, then for all $v  \in \anc(w)$ such that $\eta_v >0$, we must have $\kappa_{vw} = 0$.
 \end{proof}

\subsection{Proof of Proposition~\ref{prop:gap}}
\label{app:gap}
We consider the following function of $\eta \in H$ and $\alpha \in \rb^n$ (such that $1_n^\top \alpha=0$):
$$F(\eta,\alpha) = 
- \frac{1}{n} \sum_{i=1}^n \psi_i(-n\lambda \alpha_i)  
 - \frac{\lambda}{2}  \alpha^\top  \bigg( \sum_{w \in V} \zeta_w(\eta)\Kc_w \bigg) \alpha.
$$
This function is convex in $\eta$ (because of Lemma~\ref{lemma:1}) and concave in $\alpha$; standard arguments (e.g., primal and dual strict feasibilities) show that there is no duality gap to the variational problems:
$$
\inf_{\eta \in H } \sup_{\alpha \in \rb^n, \ 1_n^\top \alpha=0} F(\eta,\alpha)
 =   \sup_{\alpha \in \rb^n, \ 1_n^\top \alpha=0} \inf_{\eta \in H } F(\eta,\alpha).
$$
We can decompose the duality gap, given a pair $(\eta,\alpha)$ (with associated $\zeta$, $f$ and $b$) as:
\BEAS
& & \sup_{\alpha' \in \rb^n , \ 1_n^\top \alpha'=0} F(\eta,\alpha')  -  \inf_{\eta' \in H } F(\eta',\alpha) \\
&\!\!  = \!\! &   \min_{f, b}  \bigg\{
 \frac{1}{n} \sum_{i=1}^n \varphi_i \bigg( \sum_{v \in V} \langle f_v , \Phi_v(x_i) \rangle  
+  b \bigg)
+ \frac{\lambda }{2}
\sum_{w \in V}  \zeta_w(\eta)^{-1} \|f_w\|^2 \bigg\}   -  \inf_{\eta' \in H } F(\eta',\alpha)  ,
\\
&\!\! \leqslant  \!\!&  
 \frac{1}{n} \sum_{i=1}^n \varphi_i \bigg( \sum_{w \in V}\zeta_w(\eta) (K_w \alpha)_{i} + b \bigg)
+ \frac{\lambda }{2}
\sum_{w \in V}  \zeta_w \alpha^\top \Kc_w \alpha + \frac{1}{n} \sum_{i=1}^n \psi_i(-n\lambda \alpha_i) \\
& & +  \sup_{\eta' \in H }  \frac{\lambda}{2}  \alpha^\top  \sum_{w \in V} \zeta_w(\eta') \alpha, \\
&\!\! =  \!\!&  
 \frac{1}{n} \sum_{i=1}^n \varphi_i \bigg( \sum_{w \in V}\zeta_w(\eta) (K_w \alpha)_{i} + b \bigg)
+ \frac{1}{n} \sum_{i=1}^n \psi_i(-n\lambda \alpha_i)  +  {\lambda }
\sum_{w \in V}  \zeta_w(\eta) \alpha^\top \Kc_w \alpha  \\
& & +  \frac{\lambda}{2} \bigg[   \sup_{\eta' \in H }   \sum_{w \in V} \zeta_w(\eta')  \alpha^\top K_w \alpha
-  
\sum_{w \in V}  \zeta_w(\eta)  \alpha^\top \Kc_w \alpha  \bigg], \\
& = & { \rm gap }_{\rm kernel} \bigg( \sum_{w \in V}  \zeta_w(\eta)  \Kc_w , \alpha \bigg)
+\frac{\lambda}{2} { \rm gap }_{\rm weights}\left(  (\alpha^\top \Kc_w \alpha)_{w \in V}, \eta \right).
\EEAS
We thus get the desired upper bound from which Proposition~\ref{prop:gap} follows, as well as the upper bound on the duality gap in \eq{gap}.

\subsection{Proof of Propositions~\ref{prop:CNopt} and~\ref{prop:CSopt}}
\label{app:CSCN}
We assume that we know the optimal solution of a truncated problem
where the entire set of decendants of some nodes have been removed. We let denote $W$ the hull of the set of active variables.
We now consider necessary conditions and sufficient conditions for this solution to
be optimal with respect to the full problem.  
This will lead to Propositions~\ref{prop:CNopt} and~\ref{prop:CSopt}.

We first use Proposition~\ref{prop:kappa}, to get a set of $\kappa_{vw}$ for $(v,w)\in W$ for the reduced problem; the goal here is to get necessary conditions by relaxing the dual problem in \eq{kappaPB}, defining $\kappa \in L$ and find an approximate solution, while for the sufficient condition, any candidate leads to a sufficient condition. It turns out that we will use the solution of the relaxed solution required for the necessary condition for the sufficient condition.

\paragraph{Necessary condition.}
If we assume that all variables in $W$ are   active and the reduced set is optimal for the full problem, then any optimal $\kappa \in L$ must be such that $\kappa_{vw} = 0$ if $v \in W$ and $w \in W^c$, and we must have $\kappa_{vw} = \zeta_w \eta_v^{-1}$ for $v \in W$ and $w \in \des(v) \cap W$ (otherwise, $\eta_W$ cannot be optimal for the reduced problem, as detailed in the proof of Proposition~\ref{prop:kappa}).
 We then let free 
$\kappa_{vw}$ for $v,w$ in $W^c$. Our goal is to find good candidates for those free dual parameters.

We can  lowerbound the sums by maxima:
$$\max_{v \in V \cap W^c}
d_v^{-2}  \sum_{w \in \des(v) } \kappa_{vw}^2 \alpha^\top \Kc_w \alpha
\geqslant 
\max_{v \in V \cap W^c}
d_v^{-2}  \max_{w \in \des(v) } \kappa_{vw}^2 \alpha^\top \Kc_w \alpha,
$$
which can be minimized in closed form with respect to $\kappa$ leading to
$\kappa_{vw} = d_v \left( \sum_{v' \in A(w) \cap W^c} d_{v'} \right)^{-1}$ and, owing to Proposition~\ref{prop:kappa} to the following lower bound
for $\max_{\eta \in H}\sum_{w \in V} \zeta_w(\eta)  \alpha^\top  \Kc_w \alpha$:
\BEQ
\max \bigg\{ \delta^2 ,
\max_{w \in W^c } \frac{\alpha^\top \Kc_w \alpha}
{ (\sum_{v \in A(w) \cap W^c} d_{v})^2} \bigg\}
\geqslant  \!
\max \bigg\{ \delta^2 ,
\max_{w \in \sou(W^c) }
 \frac{\alpha^\top \Kc_w \alpha}
{ (\sum_{v \in A(w) \cap W^c} d_{v})^2} \bigg\},
\EEQ
where $\delta^2 = \sum_{w \in W} \zeta_w(\eta_W) \alpha^\top \Kc_w \alpha =
\Omega(f)^2$. If the reduced solution is optimal we must have this lower bound smaller than $\delta^2$, which leads to \eq{CNopt}.
Note that this necessary condition may also be obtained by considering the addition (alone) of any of the sources $w \in \sou(W^c)$ and checking that they would not enter the active set.

\paragraph{Sufficient condition.}

For sufficient conditions, we simply take the previous value obtained before for $\kappa$, which leads to the following upperbound for $\max_{\eta \in H}\sum_{w \in V} \zeta_w(\eta)  \alpha^\top  \Kc_w \alpha$:
$$ 
\max \bigg\{ \delta^2 \!,  \max_{t \in   W^c}
\sum_{w \in \des(t) } \! \frac{\alpha^\top \Kc_w \alpha}
{ (\sum_{v \in A(w) \cap W^c} d_{v})^2} \bigg\} = \!
\max \bigg\{ \delta^2\! , \!\!
\max_{t \in  \sou(W^c) }\!
\sum_{w \in \des(t) }\! \!\frac{\alpha^\top \Kc_w \alpha}
{ (\sum_{v \in A(w) \cap W^c} d_{v})^2} \bigg\} ,
$$
because for all $v \in W^c$, there exists $t \in \sou(W^c)$ such that $v \in \des(t)$.
 We have moreover for all $t \in W^c$,
 $$
   \sum_{v \in A(w) \cap W^c} d_{v} \geqslant
   \sum_{v \in A(w) \cap D(t) } d_{v},
 $$
 leading to the   upper bound:
$
A = \max \left\{ \delta^2 \! ,  \!\max_{ t \in \sou(W^c)}
\sum_{w \in D(t)} \!\frac{ \alpha^\top \Kc_w \alpha}{
(\sum_{v \in A(w) \cap D(t) } d_{v} )^2
}
 \right\}.
$
The gap in \eq{gap} is thus less than $\lambda/2 ( A - \delta^2)$, which leads to the desired result.

\subsection{Optimality Conditions for the Primal Formulation}
\label{sec:primal}

We now derive optimality conditions for the primal problem in \eq{milasso2}, when the loss functions $\varphi_i$ are differentiable, which we will need in Appendix~\ref{sec:consistency-proof}, that is:
$$
\min_{ f \in  {\F}, \ b \in \rb}  
L(f,b) 
+ \frac{\lambda }{2}  \Omega(f) ^2,
$$
where $L(f,b)$ is the differentiable loss function.  
Following~\citet{grouplasso} and Proposition~\ref{prop:dual-hkl}, the solution may be found by solving a finite-dimensional problem, and thus usual notions of calculus may be used.

Let $f\in \F = \prod_{v\in V} \F_v $ and $b\in \rb$, where $f \neq 0$, with $W$ being the   hull of the active functions (or groups). The directional derivative in the direction $(\Delta,\tau) \in \F^V \times \rb$ is equal to
$$
\langle \nabla_f L(f,b), \Delta \rangle + \nabla_b L(f,b) \tau+ \lambda 
\Omega(f)
 \bigg(  \sum_{v \in W} d_v   \Big\langle \frac{ f_{\des(v) }}{\| f_{\des(v) }\| }  , \Delta_v  \Big\rangle  + \sum_{v \in W^c} d_v \| \Delta_{\des(v)} \| \bigg),
$$
and thus $(f,b)$ if optimal if and ony if $\nabla_b L(f,b)=0$ (i.e., $b$ is an optimal constant term) and if, with $\delta = \Omega(f)  $:
\BEQ
\label{eq:optimalA-primal}
\forall w \in W, \
\nabla_{f_w} L(f,b) + \lambda  \delta
 \bigg(
 \sum_{v \in \anc(w)}   \frac{  d_v }{\| f_{\des(v) }\| }  \bigg) f_w = 0,
\EEQ
 \BEQ
 \label{eq:optimalB-primal}
 \mbox{ and }
 \forall \Delta_{W^c} \in \rb^{W^c}, \ 
     \sum_{w\in W^c}  \langle \nabla_{f_w} L(f,b) ,\Delta_w \rangle + \lambda 
 \delta
 \bigg(  \sum_{v \in W^c} d_v \| \Delta_{\des(v)} \| \bigg) \geqslant 0.
 \EEQ
 We can now define
for $K \subset V$,  $\Omega_K(f_K) = \sum_{v \in K} d_v \| f_{\des(v) \cap K} \|$, the norm reduced to the functions in $K$ and $\Omega^\ast_{K}$ its dual norm~\citep{boyd,Rock70}. The last equation may be rewritten: $\Omega_{W^c}^\ast(   \nabla_{f_W} L(f,b)) \leqslant \lambda \delta$.
Note that when regularizing by  $ \lambda \Omega(f) =  {\lambda } \sum_{v \in V}  d_v \| f_{\des(v)}  \| $ instead of $\frac{\lambda }{2} \bigg(  \sum_{v \in V}  d_v \| f_{\des(v)}  \| \bigg)^2$, we have the same optimality condition with $\delta=1$.

\section{Proof of Theorem~\ref{thm:sufficient}}

 \label{app:sufficient}

\label{sec:consistency-proof}

In this appendix, we provide the proof of Theorem~\ref{thm:sufficient} with several intermediate results. Following usual proof techniques from the Lasso literature, we will consider the optimization reduced to kernels/variables in~$\W$, and (a) show that the hull of the selected variables is indeed the hull of $\W$ (i.e., itself because we have assumed in (\textbf{A0}) that $\W$ is equal to its hull) with high probability, and (b) show that when the reduced solution is extended to $\W^c$ with zeros, we have the optimal global solution of the problem with high probability. The main difficulties are to use bounds on the dual norms of our structured norms, and to deal with the infinite-dimensional group structure within a non-asymptotic analysis, which we deal with new concentration inequalities (Appendices~\ref{app:hoeff},~\ref{app:cova} and~\ref{app:ls}).

\subsection{Notations}
Let $\hat{\mu}_v = \frac{1}{n} \sum_{i=1}^n \Phi_v(x_i) \in \F_v$ be the empirical mean and $\bmu_v = \E \Phi_v(X) \in \F_v$ the population mean of $\Phi_v(X)$ and $\hS_{vw} = \frac{1}{n} \sum_{i=1}^n (\Phi_v(x_i) - \hat{\mu}_v)\otimes (\Phi_w(x_i) -\hat{\mu}_w) $ be the empirical cross-covariance operator from $\F_{w}$ to $\F_v$ and
$q_v = \frac{1}{n} \sum_{i=1}^n  \varepsilon_i ( \Phi_v(x_i) - \hat{\mu}_v) \in \F_v$ for $v,w \in V$, where
$\varepsilon_i = y_i - \sum_{w \in \W} \f_w(x_i) - \mathbf{b}$ is the i.i.d.~Gaussian noise with mean zero and variance $\boldsymbol \sigma^2$.
By assumption (\textbf{A2}), we have  $\tr \bS_{vv} \leqslant  1$  and  $\tr \hS_{vv} \leqslant  1$  for all $v \in V$, which implies that $\lmax(\bS_{\W\W}) \leqslant |\W|$ and
 $\lmax(\hS_{\W\W}) \leqslant |\W|$.
 
 All norms on vectors in Euclidean or Hilbertian spaces are always the Euclidean or Hilbertian norms of the space the vector belongs to (which can always be inferred from context). However, we consider several norms on self-adjoint operators between Hilbert spaces. All our covariance operators are \emph{compact} and can thus be diagonalized in an Hilbertian basis, with a sequence of eigenvalues that tends to zero~\citep[see, e.g.,][]{brezis80analyse,rkhs,conway}. The usual operator norm of a self-adjoint operator $A$ is the eigenvalue of largest magnitude of $A$ and is denoted by $\| A\|_{\rm op}$; the Hilbert-Schmidt norm is the $\ell_2$-norm of eigenvalues, and is denoted by
 $\| A\|_{\rm HS}$, and is equal to the Frobenius norm in finite dimensions. Finally, the trace norm is equal to the $\ell_1$-norm of eigenvalues, and is denoted by $\| A\|_{\rm tr}$. In \mysec{operators}, we provide novel non asymptotic results on the convergence of empirical covariance operators to the population covariance operators.

 \subsection{Hoeffding's Inequality in Hilbert Spaces}
 \label{app:hoeff}
 
 In this section, we prove the following proposition, which will be useful throughout this appendix:
 \begin{proposition}
 \label{prop:hoeffding}
 Let $X_1,\dots,X_n$ be i.i.d.~zero-mean random observations in the Hilbert space $\mathcal{H}$, such that for all $i$, $\| X_i \| \leqslant 1$ almost surely. Then, we have:
 \BEQ
 \label{eq:hoeff}
 \P \bigg(  \bigg\| \frac{1}{n}\sum_{i=1}^n X_i  \bigg\| \geqslant t \bigg)
 \leqslant 2 \exp \Big( - \frac{ nt^2}{8} \Big).
 \EEQ
 \end{proposition}
 \begin{proof}
 We denote $Z= \left\| \frac{1}{n} \sum_{i=1}^n X_i \right\|$. If all $X_i$ are held fixed but one, then $Z$ may only change by $\frac{2}{n}$. Thus, from Mc Diarmid's inequality~\citep[see, e.g.,][Theorem 5.1, page  148]{massart-concentration}, we have, for all $t\geqslant 0$:
 $$
 \P( Z - \E Z \geqslant t ) \leqslant \exp( - n t^2/2 ).
 $$
 Moreover,  using the Hilbertian structure of $\mathcal{H}$:
 $$ \E Z \leqslant ( \E Z^2 )^{1/2}
 = \bigg(
 \frac{1}{n^2} \sum_{i,j=1}^n \E \langle X_i, X_j \rangle
 \bigg)^{1/2} = n^{-1/2} ( \E \| X_ 1\|^2 )^{1/2}  \leqslant  n^{-1/2}. $$
 This leads to 
$ \P( Z  \geqslant n^{-1/2}t + n^{-1/2} ) \leqslant \exp( -   t^2/2 )$ for all $t \geqslant 0$, i.e., for all
$ t \geqslant 1$, $ \P( Z  \geqslant t n^{-1/2} ) \leqslant \exp( -  (t-1)^2/2 )$. If $t\geqslant 2$, then
$(t-1)^2 \geqslant t^2 /4$, and thus 
 $ \P( Z  \geqslant t n^{-1/2} ) \leqslant \exp( -  t^2/8 ) \leqslant 2 \exp( - nt^2/8 )$. For $t \leqslant 2$, then the right hand side is greater than $ 2 \exp(-1/2) > 1$, and the bound in \eq{hoeff} is trivial.
 \end{proof}

\subsection{Concentration Inequalities for Covariance Operators}
\label{sec:operators}
\label{app:cova}
We prove the following general proposition of concentration of empirical covariance operators for the Hilbert-Schmidt norm:
\begin{proposition}
Let $X_1$,\dots,$X_n$ be i.i.d.~random observations in a measurable space $\X$, equipped with a reproducing kernel Hilbert space $\mathcal{F}$ with kernel $k$, such that $k(X_i,X_i) \leqslant 1$ almost surely. Let $\bS$ and $\hS$ be the population and empirical covariance operators. We have, for all $x\geqslant 0$:
$$
\P(  \| \bS  - \hS  \|_{\rm HS} \geqslant x n^{-1/2} )
  \leqslant    4 \exp \Big(
- \frac{ x^2}{ 32  }
\Big).
$$ 
\end{proposition}
\begin{proof}
We first concentrate the mean, using Proposition~\ref{prop:hoeffding}, since the
data is universally bounded by $ 1$:
$$
\P( \| \hat{\mu}  - \bmu  \| \geqslant t ) \leqslant 2 \exp\Big( - \frac{ n t^2}{8 } \Big).
$$

 The random variables $ (\Phi (X_i) -  {\bmu} )\otimes  (\Phi (X_i) -  {\bmu} )$ are uniformly bounded by $1$ in the Hilbert space of self-adjoint operators, equipped with the Hilbert-Schmidt norm. Thus,
 using Proposition~\ref{prop:hoeffding}, we get
 $$
\P \bigg( \bigg\| \bS -  \frac{1}{n} \sum_{i=1}^n (\Phi (X_i) -  {\bmu} )\otimes (\Phi(X_i) - {\bmu})  \bigg\|_{\rm HS} \geqslant x \bigg)
\leqslant 2 \exp \Big(
- \frac{ n x^2}{ 8  }
\Big).
$$
Thus, since $
\hS =  \frac{1}{n} \sum_{i=1}^n (\Phi (X_i) -  {\bmu} )\otimes (\Phi(X_i) - {\bmu})  + ( \bmu - \hat{\mu} ) \otimes   ( \bmu - \hat{\mu} ) $, and
$ \| ( \bmu - \hat{\mu} ) \otimes   ( \bmu - \hat{\mu} )  \|_{\rm HS} = \|  \bmu - \hat{\mu} \|^2$, 
 we get:
$$
\P(  \| \bS - \hS  \|_{\rm HS} \geqslant x )
  \leqslant     2\exp \Big(
- \frac{ n x^2}{ 32  }
\Big) + 2 \exp \Big( - \frac{ n x }{16 } \Big)
\leqslant 4 \exp \Big(
- \frac{ n x^2}{ 32   }
\Big),
$$ 
as long as $x \leqslant 2$. When $x > 2$, the bound is trivial because $\| \bS  - \hS  \|_{\rm HS} \geqslant x$ occurs with probability zero.
\end{proof}

We now prove the following general proposition of concentration of empirical covariance operators for the  \emph{trace norm}:
\begin{proposition}
\label{prop:tr}
Let $X_1$,\dots,$X_n$ be i.i.d.~random observations in a measurable space $\X$, equipped with a reproducing kernel Hilbert space $\mathcal{F}$ with kernel $k$, such that $k(X_i,X_i) \leqslant 1$ almost surely.  Let $\bS$ and $\hS$ the population and empirical covariance operators. Assume that the eigenvalues of $\bS$ are root-summable with sum of square roots of eigenvalues equal to $\C_1_2$. We have, if $ x \geqslant 4 \C_1_2$:
$$
\P(  \| \bS  - \hS  \|_{\rm tr} \geqslant x n^{-1/2} )
  \leqslant    3 \exp \Big(
- \frac{   x^2}{ 32  }
\Big).
$$ 
\end{proposition}
\begin{proof} It is shown by~\citet{ktest} that 
$$
\E \| \bS  - \hS  \|_{\rm tr}  \leqslant  \C_1_2 n^{-1/2}.
$$
Thus, following the same reasoning as in the proof of Proposition~\ref{prop:hoeffding}, we get
$$
\P \bigg( \bigg\| \bS -  \frac{1}{n} \sum_{i=1}^n (\Phi (X_i) -  {\bmu} )\otimes (\Phi(X_i) - {\bmu})   \bigg\|_{\rm tr}  \geqslant (\C_1_2 + t) n^{-1/2} \bigg) \leqslant \exp( - t^2 / 2 ),
$$
and thus if $ t\geqslant 2 \C_1_2$, we have: 
$$
\P \bigg( \bigg\| \bS -  \frac{1}{n} \sum_{i=1}^n (\Phi (X_i) -  {\bmu} )\otimes (\Phi(X_i) - {\bmu})   \bigg\|_{\rm tr}  \geqslant t n^{-1/2} \bigg) \leqslant \exp( - t^2 / 8 ).
$$ 
We thus get, for $ x \geqslant 4 \C_1_2$,
$$ \P(  \| \bS  - \hS  \|_{\rm tr} \geqslant xn^{-1/2} ) \leqslant 
\exp( - x^2 / 32 )
+ 2 \exp \bigg( - \frac{x  n^{+1/2} }{16 } \bigg) \leqslant  3 \exp( - x^2 / 32 ),
$$
as long as $ x n^{-1/2} \leqslant 2$. If this is not true, the bound to be proved is trivial.
\end{proof}

\subsection{Concentration Inequality for Least-squares Problems}
\label{app:ls}
In this section, we prove a concentration result that can be applied to several problems involving least-squares and covariance operators~\citep{ktest,kenji,grouplasso}:

\begin{proposition}
\label{prop:df}
Let $X_1$,\dots,$X_n$ be i.i.d.~random observations in a measurable space $\X$, equipped with a reproducing kernel Hilbert space $\mathcal{F}$ with kernel $k$, such that $k(X_i,X_i) \leqslant 1$ almost surely.  Let $\bS$ and $\hS$ the population and empirical covariance operators. Assume that the eigenvalues of $\bS$ are root-summable with sum of square roots of eigenvalues equal to $\C_1_2$. Let $\varepsilon$ be an independent Gaussian vector with zero mean and covariance matrix $\boldsymbol\sigma^2 \idm$. Define $q = \frac{1}{n}\sum_{i=1}^n \varepsilon_i ( \Phi(X_i) - \hat{\mu})$. We have, for all  $t  \geqslant \left( 4 \boldsymbol\sigma^2 n^{-1}
\left[
 \lambda^{-1/2}    \C_1_2 
 + \| \hS  -  \bS \|_{\rm tr}
    \lambda^{-1 }
\right]\right)^{1/2}$:
$$
\P ( \|  ( \hS  + \lambda \idm )^{-1/2}  q\| \geqslant t | X )
\leqslant  \exp( - nt^2 /2 \boldsymbol \sigma^2 )
$$
\end{proposition}
\begin{proof}
Given the input variables,  $\|  ( \hS + \lambda \idm)^{-1/2}  q \|$ is a Lipschitz-continuous function of the i.i.d.~noise vector $\varepsilon$, with Lipschitz constant $n^{-1/2}$.
Moreover, we have
\BEAS
  \E \left( \|  ( \hS  + \lambda \idm )^{-1/2}  q \| | X \right) 
& \leqslant & \E \left( \|  ( \hS  + \lambda    \idm)^{-1/2}  q \|^2 | X \right)^{1/2} 
= \boldsymbol \sigma n^{-1/2} \left( \tr \hS  ( \hS + \lambda   \idm )^{-1}   \right)^{1/2}.
\EEAS
We now follow~\citet{ktest} for bounding the empirical degrees of freedom:
\BEAS
& & \tr \hS  ( \hS + \lambda   \idm )^{-1} - \tr \bS ( \bS + \lambda  \idm )^{-1}\\
& = &  
\lambda  \tr ( \bS  + \lambda  \idm )^{-1} ( \hS  -  \bS   )( \hS + \lambda    \idm )^{-1}
\\
& \leqslant & 
  \lambda  \| \hS  -  \bS \|_{\rm tr} 
  \|( \hS + \lambda    \idm )^{-1}  \|_{\rm op}
   \|( \bS  + \lambda    \idm )^{-1}  \|_{\rm op}  \leqslant \lambda^{-1}  \| \hS  -  \bS  \|_{\rm tr} .
  \EEAS
Moreover,  we have:
$
 \tr \bS  ( \bS + \lambda  \idm )^{-1}  \leqslant 
    \lambda^{-1/2}  C_{1/2} 
 $. 
This leads to:
$$
\E \left( \|  ( \hS + \lambda \idm)^{-1/2}  q \| | X \right)^2 \leqslant 
\boldsymbol \sigma^2 n^{-1}
\left[
 \lambda^{-1/2}    \C_1_2 
 + \| \hS  -  \bS \|_{\rm tr}
    \lambda^{-1 }
\right].
$$
The final bound is obtained from concentration of Lipschitz-continuous functions of Gaussian variables~\citep{massart-concentration}:
$$
\P ( \|  ( \hS  + \lambda \idm )^{-1/2}  q\| \geqslant t | X )
\leqslant  \exp( - nt^2 /2 \boldsymbol \sigma^2 )
$$
as soon as $t^2 \geqslant 4  \boldsymbol \sigma^2 n^{-1}
\left[
 \lambda^{-1/2}    \C_1_2 
 + \| \hS  -  \bS \|_{\rm tr}
    \lambda^{-1 }
\right]$.
 \end{proof}
 
  \subsection{Concentration Inequality for Irrelevant Variables}
 In this section,  we upperbound, using Gaussian concentration inequalities~\citep{massart-concentration}, the tail-probability 
 $$\P (  \Omega_{\W^c}^\ast[ z] \geqslant t ), $$
 where $z =  - q_{\W^c } + \hS_{\W^c \W} ( \hS_{\W\W} +  D  )^{-1}  q_\W$, for a given deterministic nonnegative diagonal matrix $D$. The vector $z$ may be expressed as weighted sum of the components of the Gaussian vector $\varepsilon$. In addition,
 $ \Omega_{\W^c}^\ast[g_{\W^c}] $ is upperbounded by $\max_{w \in \W^c} \| g_w\| d_w^{-1} \leqslant d_r ^{-1} \max_{w \in \W^c} \| g_w\|$.
 Thus by concentration of Lipschitz-continuous functions of multivariate standard random variables (we have a $d_r^{-1} n^{-1/2}$-Lipschitz function of $\varepsilon$), we have~\citep{massart-concentration}:
 $$
\P \!\left[  \Omega_{\W^c}^\ast[z ] \!\geqslant\!  t \! + \! 
\E ( \Omega_{\W^c}^\ast[z ]|x) |x  \right] \leqslant  
\exp\left( \! - \frac{n t^2 d_r^2}{2  \boldsymbol\sigma^2}
\right).
$$
For all $w \in \W^c$, given $(x_1,\dots,x_n)$,  $n^{1/2} \boldsymbol\sigma^{-1}z_w \in \F_w $ is normally distributed with covariance operator which has  largest eigenvalue less than  one. 
We now decompose $\W^c$ by values of $d_w$: by assumption, $d_w$ may take value $d_r$ or a power of $\beta$ (we let denote $\mathcal{D}$ the set of values of $d_w$, $w \in V$). We get (where $x$ denotes all input observations):
\BEAS
  n^{1/2} \boldsymbol \sigma^{-1} \E  \left( \max_{w \in \W^c} \frac{ \|z_w \|}{d_w} \big|x \right)&
\leqslant & n^{1/2} \boldsymbol \sigma^{-1} \sum_{ d  \in \mathcal{D} }  \E \left( \max_{  w \in \W^c, \ d_w = d }  \frac{ \|z_w \|}{d} \Big|x \right) \\
& \leqslant& \sum_{ d  \in \mathcal{D} } \frac{2 }{d}\log( 2 | \{ w \in \W^c, d_w = d \} | )^{1/2}  \\
& \leqslant& \sum_{ d  \in \mathcal{D} } \frac{2  }{d}  \log( 2 | \{ w \in V  , d_w = d \} | )^{1/2} \\
& \leqslant& d_r^{-1} \sum_{  k \geqslant 0 } \frac{2  }{\beta ^k}  \log( 2 | \{ w \in V, \ 
\depth(w) = k \} | )^{1/2}  \\
& \leqslant& d_r^{-1} \sum_{ k \geqslant 0 } \frac{2  }{\beta ^k}  \log( 2 | \depth^{-1}(k)|)
^{1/2} =d_r^{-1}  A.
\EEAS
We thus get
 $
\P \!\left[  \Omega_{\W^c}^\ast[z ] \!\geqslant\!  \frac{\boldsymbol\sigma ( t \! + \!A  )
}{d_r n^{1/2}}  \Big| x \right] \leqslant  
\exp\left( \! - \frac{  t^2  }{2  }
\right)
$, and if we use $ t \geqslant 2 A $, we get 
\BEQ
\label{eq:irrelevant}
\P \Big( \Omega_{\W^c}^\ast[Q ] \geqslant  \frac{\boldsymbol\sigma t }{d_r n^{1/2} } \Big| x \Big)
\leqslant  \exp\left( \! - \frac{ t^2 }{8  }
\right).
\EEQ
Note that we have used the expectation of the maximum of q norms of Gaussian vectors is less than $2 (\log (2 q ) )^{1/2}$ times the maximum of the expectation of the norms.

\paragraph{Upper bound on A.}
The cardinal of $ \depth^{-1}(k)$ is less than $\num(V) \deg(V)^k$, thus, since $\beta>1$,
\BEAS
 A & =  &  \sum_{ k \geqslant 0 } \frac{2  }{\beta ^k}  \log( 2 | \depth^{-1}(k)|)
^{1/2}  \\
& \leqslant &  \sum_{ k \geqslant 0 } \frac{2  }{\beta ^k}  [ (\log( 2  \num(V) )^{1/2}  + ( k \log \deg(V) )^{1/2} ]  \\
& \leqslant &  \frac{2  }{ 1 - \beta^{-1} } (\log( 2  \num(V) )^{1/2} +
(\log \deg(V) )^{1/2} 2\sum_{ k \geqslant 0}  \beta ^{-k}   k^{1/2}  .
\EEAS
Moreover, we have, by splitting the sum at $(2 \log \beta)^{-1}$, and using the fact that after the split, the function $x \mapsto \beta ^{-x}   x^{1/2} $ is decreasing:
\BEAS 
2\sum_{ k \geqslant 0}  \beta ^{-k}   k^{1/2}   & \leqslant & 2\sum_{ k \geqslant 1}  \beta ^{-k}   k^{1/2}   \leqslant    2   \sum_{ k=1}^{(2 \log \beta)^{-1}} \beta ^{-k}   k^{1/2}
+ 2    \sum_{ k=(2 \log \beta)^{-1}}^\infty \beta ^{-k}   k^{1/2} , \\
& \leqslant & \frac{ 2   } { ( 2 \log \beta)^{3/2} }
+ 2   \int_{0}^{+\infty} \beta ^{-x}   x^{1/2} dx  ,\\
& \leqslant &  \frac{ 2  } { ( 2 \log \beta)^{3/2} }
+  2   ( \log \beta )^{-3/2}  \int_{0}^{+\infty} e^{-x}   x^{1/2} dx,  \\
& \leqslant &  \frac{  1 } { (\log \beta)^{3/2} }
\left(
1 + \Gamma(3/2)
\right)  \leqslant \frac{2  } { (\log \beta)^{3/2} }, \mbox{ where } \Gamma(\cdot) \mbox{ is the Gamma function}.
\EEAS
This leads to $A \leqslant 
 \frac{2  }{ 1 - \beta^{-1} } (\log( 2  \num(V) )^{1/2} +
(\log \deg(V) )^{1/2} \frac{2  } { (\log \beta)^{3/2} }$ and the expression for $\gamma(V)$ in Theorem~\ref{thm:sufficient}.

\subsection{Error of the Reduced Solution}
\label{sec:reduced22}
 
We   have the following loss function  (optimized with respect to the constant term $b \in \rb$)
$$L(f) = \frac{1}{2} \langle f-\f, \hS (f-\f)\rangle - \langle q, f-\f \rangle. $$
Following~\citet{grouplasso} and~\citet{nardy}, we consider the reduced problem on $\W$,  $\min_{f \in \F, \ f_{\W^c}=0}  L(f) + \lambda\Omega_{\W}(f_\W)$, with non unique solution $\hat{f}$ (since $\hS_{\W\W}$ is not invertible in general). The goal here is to show that $\hat{f}$ and $\f$ are close enough so that for all $w \in \W$, $\hat{f}_{\des(w)} \neq 0 $; this will implies that the hull of the active set of $\hat{f}$ is indeed $\W$.

As opposed to the Lasso case, we also need to consider $\tilde{f}_\W$ the minimum of
$f_\W \mapsto L(f_\W) + \frac{\lambda}{2}\sum_{v \in \W} \frac{ \| f _w\|^2}{\bzeta_w}$, which corresponds to the local quadratic approximation of the norm around $\f_\W$, where 
$$\bzeta_w^{-1}= \zeta_w(\f_\W)^{-1}
=\Omega(\f) \sum_{ v \in \anc(w) } \frac{d_v}{\| \f_{\des(v) } \| }
.$$ 
Moreover, we consider the corresponding noiseless version $\tilde{\f}_\W$ of $\tilde{f}_\W$ (the solution for $\varepsilon=0$). We will compute error bounds $\| \tilde{\f}_\W - \f_\W \|$,
 $\| \tilde{f}_\W - \tilde{\f}_\W \|$ and  $\| \tilde{f}_\W - \hat{f}_\W \|$, which will provide an upper bound on $\| \hat{f}_\W - \f_\W \|$ (see Proposition~\ref{prop:reducedbound}). In particular, once we have $\| \hat{f}_\W - \f_\W \| \leqslant \nu/2$, then we must have $\| \hat{f}_{\des(w)}\|>0$ for all $w \in \W$ and thus the hull of selected kernels is indeed $\W$.

\begin{lemma}
We have:
\BEQ
\| \tilde{\f}_\W - \f_\W \| \leqslant \left( \lambda + 
 \|  \hS_{\W\W} - \bS_{\W\W} \|_{\rm op} d_r^{-2} \right) \frac{   \Omega(\f)^2 |\W|^{1/2} }{\kappa \nu}.
\EEQ
\end{lemma}
\begin{proof}
 The function $ \tilde{\f}$ is defined
as, with $D = \Diag ( \bzeta_w^{-1} \idm ) $,
$$
\tilde{\f}_\W = ( \hS_{\W\W} + \lambda D )^{-1}  \hS_{\W\W} \f_\W \\
= \f_\W  - \lambda    ( \hS_{\W\W} + \lambda D )^{-1}  D   \f_\W .
$$
 Thus, we have 
 \begin{multline*}
 \| \tilde{\f}_\W - \f_\W \| \leqslant  \lambda \left\|
   ( \hS_{\W\W} + \lambda D )^{-1}  ( \hS_{\W\W} - \bS_{\W\W} )
  ( \bS_{\W\W} + \lambda D )^{-1} D \f_\W
 \right\| \\
 + \lambda \left\|
    ( \bS_{\W\W} + \lambda D )^{-1} D \f_\W
 \right\|.
 \end{multline*}
 We can now upper bound
 $ \left\|
    ( \bS_{\W\W} + \lambda D )^{-1} D \f_\W
 \right\| \leqslant \| \h_\W \| \kappa^{-1} \| D\|_{\rm op} \leqslant |\W|^{1/2} \kappa^{-1}   \Omega(\f)^2  \nu^{-2}$.
  \BEAS
 \| \tilde{f}_\W - \f_\W \| & \leqslant & 
 \left(
 \lambda + \|  \hS_{\W\W} - \bS_{\W\W} \|_{\rm op}  \| D^{-1} \|_{\rm op}
 \right)
 \left\|
    ( \bS_{\W\W} + \lambda D )^{-1} D \f_\W
 \right\|
  \\
& \leqslant & 
\left(
 \lambda + \|  \hS_{\W\W} - \bS_{\W\W} \|_{\rm op} d_r^{-2}
 \right)\frac{   \Omega(\f)^2 }{\nu^2} |\W|^{1/2} \kappa^{-1}  .
   \EEAS
We have used moreover the following identities:  
$$ \bzeta_w^{-1}   \geqslant  d_r^2 \ \ \mbox{ and } \ \ 
\bzeta_w^{-1} =  \Omega(\f) \sum_{v \in \anc(w)} \frac{ d_v}{ \| \f_{\des(v)} \|}  \leqslant \frac{ \Omega(\f)^2 }{\nu^2},
$$
which leads to $\|D^{-1} \|_{\rm op} \leqslant d_r^{-2}$ and $\| D\|_{\rm op} \leqslant   \Omega(\f)^2 \nu^{-2}$.
\end{proof}
\begin{lemma}
We have:
\BEQ
\| \tilde{f}_\W - \tilde{\f}_\W \| \leqslant    \lambda^{-1/2} d_r^{-1} \|  ( \hS_{\W\W} + \lambda D)^{-1/2}  q_\W\|  
 .
\EEQ
\end{lemma}
\begin{proof}
The difference $\tilde{f} - \tilde{\f}$ is equal to, with $D = \Diag ( \bzeta_w^{-1} \idm ) $,
$
\tilde{f}_\W - \tilde{\f}_\W =  ( \hS_{\W\W} + \lambda D)^{-1}  q_\W 
$.
 Thus, 
$
 \| \tilde{f}_\W - \tilde{\f}_\W  \| \leqslant \lambda^{-1/2} \| D^{-1/2} \|_{\rm op}  \times \|  ( \hS_{\W\W} + \lambda D)^{-1/2}  q_\W\|  
$, which leads to the desired result. \end{proof}

\begin{lemma}
Assume $ \| \tilde{f}_\W  - \f_\W \| \leqslant 
\nu /4 $, $\lambda \leqslant |\W| d_r^{-2}$ and  $ \| \bS_{\W\W} - \hS_{\W\W} \|_{\rm op}\leqslant \frac{\nu^2 \kappa}{16 |\W|}$.
 We have:
$$\| \tilde{f}_\W - \hat{f}_\W  \| \leqslant  \min \bigg\{ \frac{96 |\W|^{3/2}  \| \f_\W - \tilde{f}_\W \| \Omega(\f)^2}{\nu^5 \kappa d_r^2 } , \frac{ \nu^2}{8 |\W|^{3/2}} , \frac{\nu}{4} \bigg\}. $$
\end{lemma}
\begin{proof}
We consider the ball of radius $\delta \leqslant \min \{  \frac{ \nu^2}{8 |\W|^{3/2}}, \frac{\nu}{4} \}$ around $\tilde{f}_\W$, i.e.,  $B_\delta(\tilde{f}_\W) = \{ f_\W
\in \F_\W, \ \| f_\W - \tilde{f}_\W \| \leqslant \delta \}$. Since $\delta \leqslant \nu /4$
and $ \| \tilde{f}_\W  - \f_\W \| \leqslant 
\nu /4 $, then in the ball $B_\delta(\tilde{f}_\W)$, we have for all $w \in \W$,
$ \| f_{\des(w) \cap \W }\| \geqslant \nu/2$. On the ball 
$B_\delta(\tilde{f}_\W)$, the function $L_\W: f_\W \mapsto L(f_\W)$ is twice differentiable with Hessian $\hS_{\W\W}$, while the function
$H_\W: f_\W \mapsto \frac{1}{2}\Omega_\W(f_\W)^2$ is also twice differentiable.
The function $H_\W$ is the square of a sum of differentiable convex terms; a short calculation shows that the Hessian is greater than the sum of the functions times the sums of the Hessians. Keeping only the Hessians corresponding to the (assumed unique) sources of each of the connected components of $\W$, we obtain the lower bound (which still depends on $f$):
$$
 \frac{ \partial^2 H_\W} {\partial f_\W \partial f_\W} (f_\W)
 \succcurlyeq  d_r \Omega_\W(f_\W)  \Diag \left[
 \frac{1}{\| f_C\|} ( \idm - \| f_C\|^{-2} f_C f_C^\top)
 \right]_{C \in \mathcal{C}(\W)},
 $$
where $ \mathcal{C}(\W)$ are the connected components of $\W$. We can now use Lemma~\ref{lemma:lemma1} to find a lower bound on the Hessian of the objective function $L_\W + \lambda H_\W$ on the ball $B_\delta(\f_\W) $:
with $A =  \lmin[ (\langle f_C, \hS_{CD} f_D \rangle )_{C,D \in \mathcal{C}(\W) }]$, we obtain the lower bound
$$
 B = \frac{A}{3}  \min\left\{ 1 , \frac{ \lambda d_r^2 }{     |\W|  } \right\} = \frac{A \lambda d_r^2}{3|\W|},
 $$
 because $\Omega_\W(f_\W) \| f_C\|^{-1} \geqslant d_r$, $\lmax(\hS_{\W\W}) \leqslant |\W|$, and $\lambda \leqslant |\W| d_r^{-2}$.

 We have moreover on the ball $B_\delta(\tilde{f}_\W)$ (on which $\| f_\W \| \leqslant 2 \| \f_\W\| \leqslant 2 |\W|^{1/2}$),
\BEAS A & \geqslant &  \lmin[ (\langle f_C, \bS_{CD} f_D \rangle )_{C,D \in \mathcal{C}(\W) }] - \max_{C \in \mathcal{C}(\W) } \| f_C\|^2 \| \bS_{\W\W} - \hS_{\W\W} \|_{\rm op} \\
&  \geqslant  &  \kappa \min_{C \in \mathcal{C}(\W) } \sum_{w \in C} \|\bS_{ww}^{1/2} f_w\|^2   - 4   |\W|\| \bS_{\W\W} - \hS_{\W\W} \|_{\rm op} \\
 &  \geqslant  & 
 \kappa \min_{C \in \mathcal{C}(\W) } \sum_{w \in C} \|\bS_{ww}^{1/2} \f_w\|^2
 - 2 \kappa |\W|^{1/2} \delta |\W|
  - 4 |\W| \| \bS_{\W\W} - \hS_{\W\W} \|_{\rm op} \\
  & \geqslant    & \kappa \nu^2 -  \kappa \nu^2/4 -  \kappa \nu^2/4 \geqslant  \kappa \nu^2 / 2,
 \EEAS
because we have assumed that that $  2 \kappa |\W|^{1/2}\delta |\W| \leqslant \nu^2 \kappa / 4$
 and $4 |\W|\| \bS_{\W\W} - \hS_{\W\W} \|_{\rm op}   \leqslant \nu^2 \kappa / 4$.

We can now show that $\hat{f}_\W$ and $\tilde{f}_\W$ are close, which is a simple consequence  of the lower bound~$B$ on the Hessian. Indeed,
  the gradient of the objective at $\tilde{f}_W$ (applied to $z$) is equal to
 \BEAS
 \langle  \nabla_{f_\W}  L_\W(\tilde{f}_\W) + \lambda \nabla_{f_\W}  H_\W(\tilde{f}_\W)  , z \rangle
& =  & 
 + \lambda   \sum_{v \in \W} \langle  (\bzeta_w^{-1} -  {\zeta}_w(\tilde{f} _w)^{-1} )  \tilde{f} _w , z_w \rangle  \\
 & \leqslant  & 
 2 \lambda  \| z\| \ |\W|^{1/2}    \max_{v \in \W} |\bzeta_w^{-1} - {\zeta}_w(\tilde{f} _w)^{-1}  |
 \\
  & \leqslant  &     \lambda  \| z \| \ |\W|^{1/2}
 \frac{ 8  \| \f_\W - \tilde{f}_\W \|}{\nu^3   } \Omega(\f)^2,
 \EEAS
 because
 $ 
 |\bzeta_w^{-1} - {\zeta}_w(\tilde{f} _w)^{-1}| \leqslant \frac{ 2 \| \f_\W - \tilde{f}_\W \|}{\nu \zeta_w }
 \leqslant  \| \f_\W - \tilde{f}_\W \|
    \frac{ 4  \Omega(\f)^2 }{\nu^3   }.
 $
 If we choose $$\delta \geqslant 2 \frac{  \lambda    |\W|^{1/2}
 \frac{ 8  \| \f_\W - \tilde{f}_\W \|}{\nu^3   } \Omega(\f)^2 }{
 \frac{\kappa \nu^2}{2} \frac{\lambda d_r^2 }{ 3 |\W|}
 }=
 \frac{96 |\W|^{3/2}  \| \f_\W - \tilde{f}_\W \| \Omega(\f)^2}{\nu^5 \kappa d_r^2 }
 ,$$ then the minimum of the reduced cost function must occur within the ball $B_\delta(\tilde{f}_\W)$.
\end{proof}
 
 We can now combine the four previous lemma into the following proposition:
 \begin{proposition}
 \label{prop:reducedbound}
 We have:
 \BEQ
 \! \| \tilde{f}_\W  - \f_\W \|
\!   \leqslant  \! \bigg( \! \lambda + \frac{
 \|  \hS_{\W\W} - \bS_{\W\W} \|_{\rm op}}{ d_r^{2}} \! \bigg) \frac{  \Omega(\f)^2 |\W|^{1/2} }{\kappa \nu}
 +    \frac{\lambda^{-1/2}}{ d_r} \|  ( \hS_{\W\W} + \lambda D)^{-1/2}  q_\W\| .
 \EEQ
 Assume moreover $ \| \tilde{f}_\W  - \f_\W \| \leqslant 
\nu /4 $, $\lambda \leqslant |\W| d_r^{-2}$ and  $ \| \bS_{\W\W} - \hS_{\W\W} \|_{\rm op} \leqslant \frac{\nu^2 \kappa}{16 |\W| }$; then:
\BEQ
\| \f_\W - \hat{f}_\W  \| \leqslant   \| \tilde{f}_\W  - \f_\W \| + \min \bigg\{ \frac{96 |\W|^{3/2}   \| \tilde{f}_\W  - \f_\W \| \Omega(\f)^2}{\nu^5 \kappa d_r^2 } , \frac{ \nu^2}{8 |\W|^{3/2}} , \frac{\nu}{4} \bigg\}.
\EEQ
 \end{proposition}

\subsection{Global Optimality of the Reduced Solution}
\label{app:dualnorms}

We now prove, that the padded solution of the reduced problem $\hat{f}$ is indeed optimal for the full problem if we have the following inequalities (with $\mu = \lambda \Omega(\f) d_r$ and $\omega = \Omega(\f) d_r^{-1}$):
\BEA
\label{eq:A} &&\| \bS_{\W\W} - \hS_{\W\W} \|  \leqslant    
\frac{ d_r \eta \kappa \nu^2   }
{  10 \Omega(\f) |\W|^{1/2}   } = O\left( \omega^{-1} |\W|^{-1/2}
\right)
\\
\label{eq:B} && \| \bS_{\W\W} - \hS_{\W\W} \|  \leqslant    
 \frac{ \lambda^{1/2} d_r ^2 \eta \kappa \nu^2   }
{  10 \Omega(\f)|\W|^{1/2}  } = \mu^{1/2} 
 O\left( \omega^{-3/2} |\W|^{-1/2}
\right)
\\
\label{eq:C}  &&\| \f_\W - \hat{f}_\W \|   \leqslant     
\frac{ \lambda^{-1/2}  d_r ^2 \eta \kappa \nu^5   }
{  40 \Omega(\f)^3 |\W|^{1/2}   }
=
\mu^{-1/2} 
  O\left( \omega^{-5/2} |\W|^{-1/2}
\right)
\\
\label{eq:D}  && \| \f_\W - \hat{f}_\W \|   \leqslant     \min \left\{ \nu\eta/5,
\frac{ d_r  \eta   \nu^3   }
{  20 \Omega(\f)    } \right\} =  O\left( \omega^{-1}
\right)
\\
\label{eq:E}  &&\lambda^{1/2}    \leqslant    
\frac{ d_r \eta \kappa^{3/2} \nu^3   }
{  20 \Omega(\f)^2 |\W|^{1/2}   }
\mbox{ i.e., }
\mu^{1/2}  =    O\left( \omega^{-3/2} |\W|^{-1/2}
\right)
\EEA
\BEA
\label{eq:F}  &&\Omega_{\W^c}^\ast[ - q_{\W^c } + \hS_{\W^c \W} ( \hS_{\W\W} + \lambda    {D} )^{-1}  q_\W  ]   \leqslant   \lambda \Omega(\f) \eta/5 = O( \mu d_r^{-1}) \\
\label{eq:G}  &&
 \| \hat{f}_\W - \f_\W \|
 \|  ( \hS_{\W\W} + \lambda    {D} )^{-1/2}  q_\W \|
  \leqslant   \frac{ \lambda d_r^{3} \nu^3 \eta }{ 20 \Omega(\f) }
= \mu   O\left( \omega^{-2}
\right) .
\EEA

Following Appendix~\ref{sec:primal}, since $\| \hat{f}_\W - \f_\W \| \leqslant \nu/2$, the hull is indeed selected, and $\hat{f}_\W$ satisfies the local optimality condition
\BEAS
\hS_{\W\W} (\hat{f}_\W - \f_\W)- q_\W+ \lambda \Omega_\W(\hat{f}_\W)\hat{s}_\W= 0,
\\
\hS_{\W\W} (\hat{f}_\W - \f_\W)- q_\W+ \lambda  \Diag( \hat{\zeta}_w^{-1} ) \hat{f}_\W= 0,
\EEAS
where $\hat{s}_\W$ is defined  as (following the definition of $\mathbf{s}$) and
$\hat{\zeta} = \zeta( \hat{f}_\W)$:
$$\hat{s}_w = \bigg( \sum_{v \in \anc(w) } d_v \| \hf_{\des(v) } \|^{-1} \bigg) \hf_w  = \hat{\zeta}_w ^{-1} \Omega(\hat{f})^{-1}
 \hat{f}_w , \ 
\forall w \in \W. 
$$
This allows us to give a ``closed form'' solution (not really closed form because it depends on $\hat{\zeta}$, which itself depends on $\hat{f}$):
\BEAS
\hat{f}_\W - \f_\W & =  & 
( \hS_{\W\W} + \lambda   \Diag( \hat{\zeta}_w^{-1} ))^{-1} ( q_\W 
- \lambda   \Diag( \hat{\zeta}_w^{-1} ) \f_\W )  .
\EEAS
 We essentially replace $\hat{\zeta}$ by $\bzeta$ and check the optimality conditions from Appendix~\ref{sec:primal}.
 That is, we consider the event $ \Omega_{\W^c}^\ast[ \nabla L(\hat{f})_{\W^c} ] \leqslant \lambda \Omega(\hat{f}) $. We use the following inequality, with the notations
 $\g_{\W^c} = 
\Diag(\bS_{vv})_{\W^c}  C_{\W^c \W} C_{\W\W}^{-1} \Diag( \bS_{ww}^{1/2} \Omega(\f)^{-1}\bzeta_w^{-1})_\W \h_\W$ and
  $\hat{D} =   \Diag( \hat{\zeta}_w^{-1} )_\W$,   $D = \Diag({\bzeta}_w^{-1} )_\W$:
\BEAS
\Omega_{\W^c}^\ast[ \nabla L(\hat{f})_{\W^c} ]
& =  &   \Omega_{\W^c}^\ast[ - q_{\W^c } + \hS_{\W^c \W}  (\hat{f}_\W - \f_\W) ]  \\
& =  &   \Omega_{\W^c}^\ast[ - q_{\W^c } + \hS_{\W^c \W} ( \hS_{\W\W} + \lambda   \hat{D} )^{-1} ( q_\W 
- \lambda   \hat{D} \f_\W )  ]  \\
& \leqslant  &   \Omega_{\W^c}^\ast[ - q_{\W^c } + \hS_{\W^c \W} ( \hS_{\W\W} + \lambda    {D} )^{-1}  q_\W  ]   + \lambda \Omega_{\W^c}^\ast[ \mathbf{g}_{\W^c}   ]   \\
& & 
 + \lambda \Omega_{\W^c}^\ast[  \mathbf{g}_{\W^c}    -  \bS_{\W^c \W} ( \bS_{\W\W} + \lambda   {D})^{-1}     {D}\f_\W )   ]  
\\
& & + \lambda \Omega_{\W^c}^\ast[
\bS_{w \W} ( \bS_{\W\W} + \lambda  D)^{-1}    D \f_\W  
- \hS_{w \W} ( \hS_{\W\W} + \lambda  \hat{D} )^{-1}    \hat{D}\f_\W    ]\\
& & + \Omega_{\W^c}^\ast[  \hS_{\W^c \W}
 ( \hS_{\W\W} + \lambda    \hat{D} )^{-1} q_\W
 -    ( \hS_{\W\W} + \lambda    {D} )^{-1}  q_\W] \\
 & \leqslant  &   \Omega_{\W^c}^\ast[ - q_{\W^c } + \hS_{\W^c \W} ( \hS_{\W\W} + \lambda    {D} )^{-1}  q_\W  ]   + \lambda \Omega_{\W^c}^\ast[ \mathbf{g}_{\W^c}   ]  \\
 & &   + \lambda( A + B + C ).
\EEAS
We will bound the last  three terms $A$, $B$ and $C$ by $  \Omega(\f) \eta/5$, bound the difference
$| \Omega(\f)  - \Omega(\hf)| \leqslant \eta \Omega(\f)/5$ (which is implied by \eq{D}) and use the assumption
$  \Omega_{\W^c}^\ast[\mathbf{g}_{\W^c}  ] \leqslant 1 - \eta$, 
 and use the bound in \eq{F} to bound $ \Omega_{\W^c}^\ast[ - q_{\W^c } + \hS_{\W^c \W} ( \hS_{\W\W} + \lambda    {D} )^{-1}  q_\W  ] \leqslant \lambda \Omega(\f) \eta/5$. 
 Note that we have the bound $\Omega_{\W^c}^\ast [ g_{\W^c} ] \leqslant \max_{ v\in \W^c} \frac{ \| g_v \| }{d_v}$, obtained by lower bounding $\|f_{\des(v)}\|$ by $\| f_v\|$ in the definition of $\Omega_{\W^c}$.
 
\paragraph{Bounding $B$.}
We have:
\BEAS
R & = & \hS_{w \W} ( \hS_{\W\W} + \lambda  \hat{D} )^{-1}    \hat{D}\f_\W )   
-
\bS_{w \W} ( \bS_{\W\W} + \lambda  D)^{-1}    D \f_\W )  \\
& = & ( \bS_{w \W} -  \hS_{w \W} ) ( \bS_{\W\W} + \lambda D)^{-1}    D\f_\W ) \\
& & + 
 \hS_{w \W} ( \hS_{\W\W} + \lambda   \hat{D} )^{-1} 
 ( \bS_{\W\W} + \lambda   D
 -  \hS_{\W\W} + \lambda   \hat{D})
 ) (  ( \bS_{\W\W} + \lambda   D)^{-1}    D\f_\W ) ) \\
 & & + 
  \hS_{w \W} ( \hS_{\W\W} + \lambda   \hat{D})^{-1} 
    \Diag(  \hat{\zeta}_w^{-1}-  {\bzeta}_w^{-1} )   \f_\W
\\
\| R\| 
& \leqslant &   \| \bS_{w \W} -  \hS_{w \W} \|_{\rm op} \|D \|_{\rm op} |\W|^{1/2} \kappa^{-1} \\
& & 
+ \lambda^{-1/2} \| \hat{D}^{-1/2} \|_{\rm op}  \|D \|_{\rm op} |\W|^{1/2} \kappa^{-1}
\left(  \| \bS_{\W \W} -  \hS_{\W \W} \|_{\rm op} + \lambda \| D - \hat{D} \|_{\rm op}
\right)  \\
& & +   \| D - \hat{D} \|_{\rm op} |\W|^{1/2} 
\\
& \leqslant &   \| \bS -  \hS \| 2 \Omega(\f)^2 \nu^{-2} |\W|^{1/2} \kappa^{-1} \\
& & 
+ \lambda^{-1/2} d_r^{-1} \| 2 \Omega(\f)^2 \nu^{-2}  |\W|^{1/2}  \kappa^{-1}
\left(  \| \bS_{\W \W} -  \hS_{\W \W} \|_{\rm op} + \lambda 4 \Omega(\f)^2 \nu^{-3} \| \hat{f} - \f \| 
\right)  \\
& & +   |\W|^{1/2}   4 \Omega(\f)^2 \nu^{-3} \| \hat{f} - \f \| ,
\EEAS
which leads to an upper bound $B \leqslant d_r^{-1} \| R\|$. The constraints imposed by \eq{A}, \eq{B}, \eq{C} and \eq{D} imply that $B \leqslant \Omega(\f) \eta / 5$.

\paragraph{Bounding $A$.}
We consider the term $\bS_{\W^c \W} ( \bS_{\W\W} + \lambda   {D})^{-1}     {D}\f_\W )$. Because of the operator range conditions used by~\citet{grouplasso} and~\citet{kenji}, we can write
$$ \Diag( \bS_{vv}^{1/2} )  C_{\W\W}  \Diag( \bS_{vv}^{1/2} )  \gamma =  \bS_{\W\W} \gamma = {D}  \Diag( \bS_{vv} )  \h_\W,$$
where $\| \gamma \| \leqslant  \| D\| \kappa^{-1} \| \h \|$. We thus have
\BEAS
\bS_{w \W} ( \bS_{\W\W} + \lambda   {D})^{-1}     {D}\f_\W 
& = & \bS_{ww}^{1/2} C_{w \W} \Diag( \bS_{vv}^{1/2} )_{\W} 
 ( \bS_{\W\W} + \lambda   {D})^{-1}     {D}  \Diag( \bS_{vv} )_{\W}  \h_\W \\
 & = & \bS_{ww}^{1/2} C_{w \W} \Diag( \bS_{vv}^{1/2} )_{\W} 
 ( \bS_{\W\W} + \lambda   {D})^{-1}   \bS_{\W\W}   \gamma \\
 & = & \bS_{ww}^{1/2} C_{w \W} \Diag( \bS_{vv}^{1/2} )_{\W}  \gamma  \\
 & & 
 - \bS_{ww}^{1/2} C_{w \W} \Diag( \bS_{vv}^{1/2} )_{\W} 
 ( \bS_{\W\W} + \lambda   {D})^{-1}   \lambda D   \gamma .
\EEAS
We have moreover
\BEAS
 \bS_{ww}^{1/2} C_{w \W}  C_{\W\W}^{-1}    {D}  \Diag( \bS_{vv}^{1/2} )_{\W}  \h_\W
 & = &  \bS_{ww}^{1/2} C_{w \W}  C_{\W\W}^{-1} C_{\W\W}  \Diag( \bS_{vv}^{1/2} )  \gamma,
\EEAS
which leads to an upper bound for $A$:
$$
A \leqslant  \kappa^{-1/2} \lambda^{1/2} \|D\|_{\rm op}^{1/2} \| \gamma \|
\leqslant \kappa^{-3/2} \lambda^{1/2} \|D\|_{\rm op}^{3/2} |\W|^{1/2}  
\leqslant 4 \kappa^{-3/2} \lambda^{1/2} \Omega(\f)^3 \nu^{-3}  |\W|^{1/2}.
$$
The constraint imposed on \eq{E} implies that $A \leqslant \Omega(\f) \eta / 5$.

\paragraph{Bounding $C$.}
We consider, for $w \in \W^c$:
\BEAS
T & = & \hS_{w \W}
 ( \hS_{\W\W} + \lambda    \hat{D} )^{-1} q_\W
 -  \hS_{w \W}  ( \hS_{\W\W} + \lambda    {D} )^{-1}  q_\W \\
 & = &  \lambda
 \hS_{w \W}  ( \hS_{\W\W} + \lambda    {D} )^{-1} ( {D} - \hat{D} )
 ( \hS_{\W\W} + \lambda    \hat{D} )^{-1} q_\W \\ 
 \lambda^{-1} \| T \|
 & \leqslant & \lambda^{-1} \| D^{-1} \|_{\rm op}  \| {D} - \hat{D} \|_{\rm op}
 \|  ( \hS_{\W\W} + \lambda    {D} )^{-1/2}  q_\W \| \\
 & \leqslant &  4 \lambda^{-1} d_r^{-2}  \Omega(\f)^2 \nu^{-3} \| \hat{f}_\W - \f_\W \|
 \|  ( \hS_{\W\W} + \lambda    {D} )^{-1/2}  q_\W \|,
 \EEAS
 leading to the bound
 $C \leqslant d_r^{-1} \lambda^{-1} \| T \|$. 
The constraint imposed on \eq{G} implies that $C \leqslant \Omega(\f) \eta / 5$.

\subsection{Probability of Incorrect Hull Selection}
We now need to lower bound the probability of all events from \eq{A}, \eq{B}, \eq{C}, \eq{D}, \eq{E}, \eq{F} and \eq{G}.
They can first be summed up as:
\BEAS
\| \f_\W - \hat{f}_\W \| & \leqslant &  O\left( \mu^{1/4} \omega^{-1} |\W|^{-1/2}
\right) \\
\mu & \leqslant &  O\left( \omega^{-3} |\W|^{-1}
\right) \\
\| \bS_{\W\W} - \hS_{\W\W} \|_{\rm tr} & \leqslant &  O\left( \omega^{-3/2} |\W|^{-1/2}
 mu^{1/2} 
\right) \\
\Omega_{\W^c}^\ast[ - q_{\W^c } + \hS_{\W^c \W} ( \hS_{\W\W} + \lambda    {D} )^{-1}  q_\W  ] & \leqslant & \lambda \Omega(\f) \eta/5 = O( \mu d_r^{-1})   \\
 \|  ( \hS_{\W\W} + \lambda    {D} )^{-1/2}  q_\W \| & \leqslant & O\left( \mu^{3/4} \omega^{-1} |\W|^{1/2}
\right) .
\EEAS
From Proposition~\ref{prop:reducedbound}, in order to have $\| \f_\W - \hat{f}_\W \|  \leqslant O\left( \mu^{1/4} \omega^{-1} |\W|^{-1/2}
\right) $, we need to have
$
\| \f_\W - \tilde{f}_\W \|   \leqslant    O\left( \mu^{1/4} \omega^{-3} |\W|^{-2}
\right)$, i.e., 
$\|  ( \hS_{\W\W} + \lambda    {D} )^{-1/2}  q_\W \| \leqslant O( \mu^{3/4}  \omega^{-7/2}  |\W|^{-2})$, $\mu = O( \mu^{1/4} \omega^{-4}  |\W|^{-5/2})$ and 
$\| \bS_{\W\W} - \hS_{\W\W} \|_{\rm tr} = O ( \mu^{1/4}\omega^{-5}  |\W|^{-5/2})$.
 
From Proposition~\ref{prop:df}, in order to bound $ \|  ( \hS_{\W\W} + \lambda    {D} )^{-1/2}  q_\W \|$, we require
$\| \bS_{\W\W} - \hS_{\W\W} \|_{ \rm tr} = O(  \mu^{1/2} \omega^{-1/2} |\W|^{-3/2})$.
We finally require the following bounds:
\BEAS
\mu & \leqslant &  O\left( \omega^{-11/2} | \W |^{-7/2}
\right) \\
\| \bS_{\W\W} - \hS_{\W\W} \|_{\rm tr} & \leqslant &  O\left(  \mu^{1/2}  \omega^{-3/2} |\W|^{-1/2} 
\right) \\
\Omega_{\W^c}^\ast[ - q_{\W^c } + \hS_{\W^c \W} ( \hS_{\W\W} + \lambda    {D} )^{-1}  q_\W  ] & \leqslant &  O( \mu d_r^{-1})   \\
 \|  ( \hS_{\W\W} + \lambda    {D} )^{-1/2}  q_\W \| & \leqslant & O\left( \mu^{3/4} \omega^{-7/2}
 |\W|^{-2}  \right) .
\EEAS
We can now use Propositions~\ref{prop:tr} and \ref{prop:df} as well as \eq{irrelevant} to obtain the desired upper bounds on probabilities.

\subsection{Lower Bound on Minimal Eigenvalues}
We provide a lemma used earlier in Section~\ref{sec:reduced22}.
\begin{lemma}
\label{lemma:lemma1}
Let $Q$ be a symmetric matrix defined by blocks and $(u_i)$ a sequence of unit norm vectors adapted to the blocks defining $Q$. We have:
\BEAS \lmin\left(Q + \Diag\left[ \mu_i( \idm - u_i u_i ^\top ) \right] \right)
 & 
 \geqslant &
\frac{ \lmin[ ( u_i^\top Q_{ij} u_j)_{i,j}]}{3} \min \left\{ 1 , \frac{ \min_{i} \mu_i}{ \lmax(Q) } \right\}.
\EEAS
\end{lemma}
\begin{proof}
We consider the  orthogonal complements $V_i$ of $u_i$, we then have 
\BEAS
&  [ u_1,\dots, u_p ]^\top 
\left(Q + \Diag\left[ \mu_i( \idm - u_i u_i ^\top ) \right] \right) 
[ u_1,\dots, u_p ] = ( u_i^\top Q_{ij} u_j)_{i,j}   
\\ 
&{  [ V_1,\dots, V_p ]^\top }
\left(Q + \Diag\left[ \mu_i( \idm - u_i u_i ^\top ) \right] \right) 
[ V_1,\dots ,V_p ] = ( V_i^\top Q_{ij} V_j + \delta_{i=j} \mu_i \idm )_{i,j}
  \\ 
& {[ V_1,\dots, V_p ]^\top }
\left(Q + \Diag\left[ \mu_i( \idm - u_i u_i ^\top ) \right] \right) 
[ u_1,\dots ,u_p ] = ( V_i^\top Q_{ij} u_j   )_{i,j}.
\EEAS
We can now consider Schur complements: the eigenvalue we want to lower-bound is greater than $\nu$ if
$\nu \leqslant \lmin[ ( u_i^\top Q_{ij} u_j)_{i,j}
 ]$ and
 $$
  ( V_i^\top Q_{ij} V_j + \delta_{i=j} \mu_i \idm )_{i,j}
 -  ( V_i^\top Q_{ij} u_j   )_{i,j}
 (  ( u_i^\top Q_{ij} u_j)_{i,j} - \nu \idm)^{-1}
 ( u_i^\top Q_{ij} V_j   )_{i,j} \succcurlyeq \nu \idm
 $$
 which is equivalent to 
 \begin{multline}
  ( V_i^\top Q_{ij} V_j )_{i,j}  + \Diag(   \mu_i \idm )
   -  ( V_i^\top Q_{ij} u_j   )_{i,j}
   ( u_i^\top Q_{ij} u_j)_{i,j} ^{-1}
 ( u_i^\top Q_{ij} V_j   )_{i,j}  
 \\
 +
  ( V_i^\top Q_{ij} u_j   )_{i,j}
  \left[  ( u_i^\top Q_{ij} u_j)_{i,j} ^{-1} -
 (  ( u_i^\top Q_{ij} u_j)_{i,j} - \nu \idm)^{-1} \right]
 ( u_i^\top Q_{ij} V_j   )_{i,j} \succcurlyeq \nu \idm.
  \end{multline}
If we assume that $\nu \leqslant \lmin[ ( u_i^\top Q_{ij} u_j)_{i,j}
 ]/2$, then the second term has spectral norm less than
 $\frac{2 \nu \lmax(Q) }{ \lmin[ ( u_i^\top Q_{ij} u_j)_{i,j})}$. The result follows.
\end{proof}

\acks{I would like to thank Rodolphe Jenatton, Guillaume Obozinski, Jean-Yves Audibert and Sylvain Arlot for fruitful discussions related to this work. This work was supported by a French grant from the Agence Nationale de la Recherche (MGA Project ANR-07-BLAN-0311).
}

\bibliography{hkl_hal}

\end{document}